%% file: sample.tex
\documentclass[twoside,11pt]{article}

\usepackage{blindtext}

\usepackage[abbrvbib, preprint]{jmlr2e}
\usepackage{jmlr2e}
\usepackage{lmodern}
\usepackage{pdflscape}

\usepackage[utf8]{inputenc}
\usepackage{amsmath}
\usepackage{amssymb}
\usepackage{bm} 
\usepackage{mathrsfs} 
\usepackage{algorithm}
\usepackage{multicol}
\usepackage{multirow}
\usepackage{tikz}
\usepackage{graphicx} 
\usepackage{url}
\usetikzlibrary{arrows}
\usepackage{changepage}
\usepackage{booktabs}
\usetikzlibrary{arrows.meta}
\usepackage{algorithmic}
\newcommand{\kronecker}{\otimes}     
\newcommand{\khatri}{\odot}          
\newcommand{\hadamard}{\circledast}      

\newcommand{\sd}[1]{$\,\pm\,$#1}
\newcommand{\dpp}{\overline{\mathrm{\mathcal{CE}}}}

\usepackage{tikz}
\usetikzlibrary{calc}
\usepackage{bm}        
\usepackage{amsmath}   

\newcommand{\tensor}[1]{\boldsymbol{\mathcal{#1}}}
\newcommand{\W}[1]{\tensor{W}^{(#1)}}

\usepackage{lastpage}
\jmlrheading{23}{2026}{1-\pageref{LastPage}}{1/21; Revised 5/22}{9/22}{21-0000}{Afra Kilic and Kim Batselier. \textit{A shorter version of this paper has been submitted to ESANN 2027}}

\ShortHeadings{ Bayesian Tensor Train Kernel Machines}{ Bayesian Tensor Train Kernel Machines}
\firstpageno{1}

\begin{document}

\title{Bayesian Tensor Train Kernel Machines with Automatic Model Complexity Inference}

\author{\name Afra Kilic \email h.a.kilic@tudelft.nl \\
           \addr Delft Center for Systems and Control\\
       Delft University of Technology\\
       Delft, 2628 CN, The Netherlands
       \AND
       \name Kim Batselier \email k.batselier@tudelft.nl \\
       \addr Delft Center for Systems and Control\\
       Delft University of Technology\\
       Delft, 2628 CN, The Netherlands}

\maketitle
\begin{abstract}
Tensor network (TN) kernel machines speed up supervised learning by using low-rank TNs to reduce computation and memory. Most of these methods, however, are deterministic: they ignore parameter uncertainty, and their model complexity has to be tuned manually. Bayesian TN kernel machines remove both limitations, but only for the CANDECOMP/PARAFAC decomposition (CPD), which has a single rank. For the more general tensor train (TT), manual tuning is even more costly, since every one of its ranks must be chosen. In this paper we propose Bayesian Tensor Train Kernel Machines (BTT-KM), which place hierarchical ARD-type priors on the TT cores and infer all TT ranks and feature dimensions from data. Mean-field variational inference leads to a Bayesian alternating least squares algorithm with the same computational complexity as its deterministic counterpart, and gives a closed-form predictive distribution. On synthetic data, BTT-KM shrinks an over-parameterized initialization to a low-rank model close to the ground truth, and on UCI benchmarks it is competitive with state-of-the-art methods, with predictive coverage closest to the nominal value.
\end{abstract}

\vspace{3mm}

\input{sections/1.introduction}

\input{sections/2.background}
\input{sections/3.model_learning}

\input{sections/4.experiments}

\input{sections/5.conclusion}

\section*{Code Availability}
The implementation of BTT-KM and the code to reproduce the experiments are available at
\url{https://github.com/afrakilic/BTTKM}.
\acks{This publication is part of the project Sustainable learning for Artificial Intelligence from noisy large-scale data (with project number VI.Vidi.213.017) which is financed by the Dutch Research Council (NWO).}

\bibliography{references}

\clearpage
\appendix

\begin{adjustwidth}{-1.2cm}{-1.2cm}
\input{sections/6.appendix}
\end{adjustwidth}

\end{document}

%% file: sections/1.introduction.tex
\section{Introduction}
\label{sec:introduction}

Tensor network (TN) kernel machines are a class of kernel methods designed for high-dimensional feature spaces and large-scale datasets. By learning directly in feature space and imposing a low-rank tensor structure on the model weights, they avoid the memory and computational cost of standard kernel methods. For a set of data samples $\{ (\mathbf{x}_n, y_n) \mid \mathbf{x}_n \in \mathbb{R}^D,\; y_n \in \mathbb{R} \}_{n=1}^N$, TN kernel machines consider the model
\begin{equation}
    y_n = \bm{\varphi}(\mathbf{x}_n)^\top \mathbf{w} + e_n,
    \label{eq:1}
\end{equation}
where $\mathbf{w}$ denotes the model weight vector and $e_n$ represents measurement noise. The feature map $\bm{\varphi}(\cdot) : \mathbb{R}^D \rightarrow \mathbb{R}^{M_1 M_2 \cdots M_D}$ is constrained to a Kronecker-product structure
\begin{equation}
    \bm{\varphi}(\mathbf{x}_n)
    = \bm{\varphi}^{(1)}(x^{(1)}_n) \otimes \bm{\varphi}^{(2)}(x^{(2)}_n) \otimes \cdots \otimes \bm{\varphi}^{(D)}(x^{(D)}_n),
    \label{eq:2}
\end{equation}
where each factor $\bm{\varphi}^{(d)} \in \mathbb{R}^{M_d}$ is a feature vector for dimension $d$. This feature map induces the finite-dimensional kernel $k_M(\mathbf{x}_n, \mathbf{x}_m) = \langle \bm{\varphi}(\mathbf{x}_n), \bm{\varphi}(\mathbf{x}_m) \rangle$, which, depending on the choice of basis functions, can represent the target kernel exactly, as for polynomial features \citep{batselier_tensor_2017, Novikov_A._Exponential_2018}, or approximate it, as for Gaussian kernels \citep{wesel_large-scale_2021, wesel2024quantizedfourierpolynomialfeatures}. Model~\eqref{eq:1} is therefore a primal formulation of kernel machines \citep{suykens2002, wesel_large-scale_2021, pmlr-v258-wesel25a}: $\mathbf{w}$ is learned directly in feature space, avoiding the $\mathcal{O}(N^2)$ memory and $\mathcal{O}(N^3)$ cost of the dual formulation.

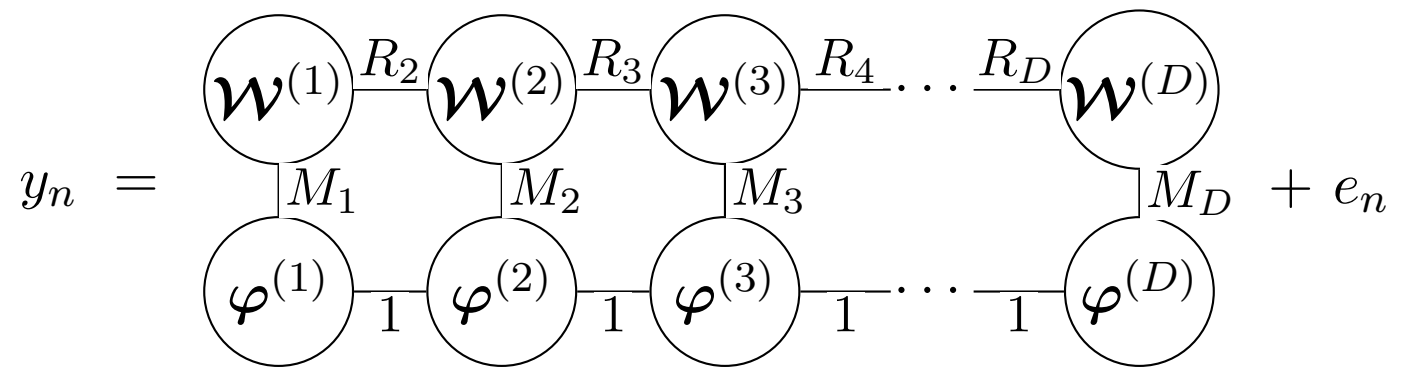
\begin{figure}[!tb]
\centering
\input{TN_diagram}
\caption{\footnotesize{Tensor network diagram of model~\eqref{eq:1}: nodes are tensors, edges are their indices labeled by dimension, and connected edges are summed over. The top row is $\mathbf{w}$ in the TT format of Definition~\ref{def:tt}: each node is a TT core $\bm{\mathcal{W}}^{(d)} \in \mathbb{R}^{R_d \times M_d \times R_{d+1}}$, and each horizontal edge $R_d$ corresponds to the sum over $r_d$ in~\eqref{eq:3}; the boundary edges $R_1=R_{D+1}=1$ are omitted. The bottom row is $\bm{\varphi}(\mathbf{x}_n)$ with $\bm{\varphi}^{(d)}=\bm{\varphi}^{(d)}(x_n^{(d)})$. Each vertical edge $M_d$ connects $\W{d}$ to $\bm{\varphi}^{(d)}$ and corresponds to the sum over $m_d$ in $\bm{\varphi}(\mathbf{x}_n)^\top\mathbf{w}$.}}
\vspace{-0.5cm}
\label{fig:tt}
\end{figure}

\vspace{0.35cm}
The price of the primal formulation is the size of $\mathbf{w}$: assuming $M_1=\cdots=M_D=M$ for simplicity, it has $M^D$ entries, a number that grows exponentially with $D$. By exploiting the Kronecker-product structure of~\eqref{eq:2}, TN kernel machines impose a low-rank TN structure on $\mathbf{w}$, so that it is never formed explicitly but represented and learned directly through its tensor components. Two common choices are the CANDECOMP/PARAFAC decomposition (CPD)~\citep{harshman1970parafac, carroll_analysis_1970} and the tensor train (TT) decomposition~\citep{oseledets2011tensor}. Fig.~\ref{fig:tt} shows model~\eqref{eq:1} with $\mathbf{w}$ in the TT format, of which the CPD is the special case with a single rank $R_2=\cdots=R_D=R$ and diagonal inner TT cores. Assuming also equal TT ranks $R_d = R$ for simplicity, the storage of $\mathbf{w}$ drops from $\mathcal{O}(M^D)$ to $\mathcal{O}(DMR^2)$ for TT and $\mathcal{O}(DMR)$ for CPD, which is linear in $D$. Both are typically learned with the alternating least squares (ALS) algorithm~\citep{batselier_tensor_2017, wesel_large-scale_2021, holtz_alternating_2012}, which updates one component at a time by solving a linear least-squares problem while keeping the others fixed. Its cost per iteration, $\mathcal{O}(ND(MR)^2+D(MR)^3)$ for CPD and $\mathcal{O}(ND(MR^2)^2+D(MR^2)^3)$ for TT, is linear in $N$ and $D$, which makes both formats suitable for large-scale, high-dimensional data. Finally, the low rank acts as an implicit regularizer by limiting the degrees of freedom of the model.

\vspace{0.35cm}
Despite their efficiency, most TN-based methods are deterministic. They ignore parameter uncertainty and therefore cannot quantify the uncertainty of their predictions, which matters when data are noisy or limited. They also require the model complexity parameters, the tensor rank $R$ and the feature dimensions $M_d$, to be tuned in advance. Typically, this is done via trial and error, which can be time-consuming and imprecise. Bayesian inference addresses both issues. It quantifies the uncertainty of the predictions and, with suitable priors, can infer the model complexity from data, promote sparsity and incorporate prior knowledge. Existing Bayesian TN models either infer the ranks only for tensor factorization and completion~\citep{Zhao_2015_rank_det,Xu2023}, or fix them a priori, as in TT-GP~\citep{Izmailov2018} and SP-BTN~\citep{Konstantinidis2022}. Bayesian TN kernel machines (BTN-KM) were introduced in~\citet{kilic_interpretable_2025}, where hierarchical automatic relevance determination (ARD)-type priors on the factor matrices infer the tensor rank and feature dimensions from data. Mean-field variational inference then gives a Bayesian ALS algorithm with the same computational complexity as its deterministic counterpart. 
BTN-KM, however, is restricted to the CPD, which has a single rank, and without an additional stabilization step its training can become numerically unstable on some high-dimensional data sets.

The TT format, which is more general than the CPD, has been used for image classification~\citep{stoudenmire2016supervised}, polynomial classification~\citep{chen2018parallelized}, recommender systems~\citep{Novikov_A._Exponential_2018} and nonlinear system identification~\citep{batselier_tensor_2017}. Since each rank $R_d$ is shared only by two neighboring TT cores (Fig.~\ref{fig:tt}), the model capacity can vary along the train, with each $R_d$ controlling the interaction between the inputs on its left and on its right, whereas the CPD uses one rank for all inputs. Moreover, for large $D$, a TT model converges faster to a GP than a CPD model with the same number of parameters~\citep{pmlr-v258-wesel25a}. Intuitively, a CPD output is a sum of products of $D$ terms, whereas a TT averages over its ranks at every step of the train. With random initial cores, the TT output may therefore be better behaved, which suggests that a Bayesian TT model is less prone to the numerical instability of BTN-KM on high-dimensional data. Existing TT kernel machines, however, are deterministic and provide no uncertainty estimates. In addition, their $D-1$ ranks make manual tuning far more costly than for CPD, since a grid search over $K$ values per rank already requires $K^{D-1}$ trainings. Thus, a Bayesian treatment of TT kernel machines that quantifies uncertainty and infers the ranks from data is needed.

\textbf{In this paper}, we therefore extend BTN-KM to the TT format and propose Bayesian Tensor Train Kernel Machines (BTT-KM). This extension is not a direct translation of the CPD case. Since each TT core is connected to the rest of the train only through its neighbors, its closed-form posterior update requires propagating the contributions of all other cores along the train.
We adapt the hierarchical prior to the TT structure through two sets of precisions, which act on the edges of Fig.~\ref{fig:tt}. The rank precisions $\bm{\lambda}_d$ act on the edge $R_d$, so they are shared by the two cores it connects and regularize each TT rank separately. The feature precisions $\bm{\delta}_d$ act on the edge $M_d$, and therefore only within the core $\W{d}$. We show that mean-field variational inference again leads to a Bayesian ALS algorithm with the same computational complexity as deterministic TT-ALS, and derive a closed-form Student's t predictive distribution. The resulting method infers the TT ranks $\{R_d\}_{d=2}^D$ and the feature dimensions $\{M_d\}_{d=1}^D$ directly from data. A synthetic experiment illustrates how the regularization acts on the TT cores and shrinks an over-parameterized initialization to a low-rank model close to the ground truth. Experiments on UCI benchmarks show that BTT-KM is competitive with state-of-the-art methods in accuracy and uncertainty quantification, with predictive coverage closest to the nominal value, and that, unlike BTN-KM, it trains stably without such a stabilization step.

The remainder of the paper is organized as follows. Section~\ref{sec:preliminaries} introduces the notation and TT preliminaries. Section~\ref{sec:model} presents BTT-KM: the model and priors (Section~\ref{sec:priors}), the variational updates and lower bound (Section~\ref{sec:learning}), the predictive distribution (Section~\ref{sec:predictive}), and implementation details and complexity (Section~\ref{sec:implementation}). Section~\ref{sec:experiments} illustrates the regularization on synthetic data (Section~\ref{sec:synthetic}) and compares BTT-KM with BTN-KM and a GP on UCI benchmarks (Section~\ref{sec:uci}), and Section~\ref{sec:conclusion} concludes. Proofs are in the Appendix, and code is available at \url{https://github.com/afrakilic/BTTKM}.

%% file: TN_diagram.tex
\vspace{-1cm}
\usetikzlibrary{calc}
\begin{tikzpicture}[
  x=1pt, y=1pt,
  every node/.style={font=\normalsize},
  circ/.style={circle, draw, minimum size=28pt, inner sep=0pt},
  lbl/.style={font=\small, fill=white, inner sep=1pt}
]
  \def\dx{42}    
  \def\dxd{78}   
  \def\dy{38}    

  \node[circ] (w1) at (0,0)          {$\W{1}$};
  \node[circ] (w2) at (\dx,0)        {$\W{2}$};
  \node[circ] (w3) at (2*\dx,0)      {$\W{3}$};
  \node[circ] (wD) at (2*\dx+\dxd,0) {$\W{D}$};

  \node[circ] (p1) at (0,-\dy)          {$\bm{\varphi}^{(1)}$};
  \node[circ] (p2) at (\dx,-\dy)        {$\bm{\varphi}^{(2)}$};
  \node[circ] (p3) at (2*\dx,-\dy)      {$\bm{\varphi}^{(3)}$};
  \node[circ] (pD) at (2*\dx+\dxd,-\dy) {$\bm{\varphi}^{(D)}$};

  \draw (p1) -- (p2) node[lbl,midway,below] {$1$};
  \draw (p2) -- (p3) node[lbl,midway,below] {$1$};
  \draw (p3) -- ($(p3)!0.4!(pD)$) node[lbl,midway,below] {$1$};
  \draw ($(pD)!0.4!(p3)$) -- (pD) node[lbl,midway,below] {$1$};
  \node at ($(p3)!0.5!(pD)$) {$\cdots$};

  \node at (-\dx*0.85,-\dy/2) {$y_n\ =$};
  \node at (2*\dx+\dxd+\dx*0.85,-\dy/2) {$+\ e_n$};

  \draw (w1) -- (w2) node[lbl,midway,above] {$R_2$};
  \draw (w2) -- (w3) node[lbl,midway,above] {$R_3$};
  \draw (w3) -- ($(w3)!0.4!(wD)$) node[lbl,midway,above] {$R_4$};
  \draw ($(wD)!0.4!(w3)$) -- (wD) node[lbl,midway,above] {$R_D$};
  \node at ($(w3)!0.5!(wD)$) {$\cdots$};

  \draw (w1) -- (p1) node[lbl,midway,right] {$M_1$};
  \draw (w2) -- (p2) node[lbl,midway,right] {$M_2$};
  \draw (w3) -- (p3) node[lbl,midway,right] {$M_3$};
  \draw (wD) -- (pD) node[lbl,midway,right] {$M_D$};
\end{tikzpicture}

%% file: sections/2.background.tex
\section{Preliminaries and notation}
\label{sec:preliminaries}
The order of a tensor is the number of dimensions, also called modes. Scalars are denoted by lowercase letters (e.g., $a$), vectors (first-order tensors) by bold lowercase letters (e.g., $\mathbf{a}$), matrices (second-order tensors) by bold uppercase letters (e.g., $\mathbf{A}$), and higher-order tensors (of order $D \geq 3$) by bold calligraphic symbols (e.g., $\bm{\mathcal{A}}$). For a $D$th-order tensor $\bm{\mathcal{A}} \in \mathbb{R}^{I_1 \times \cdots \times I_D}$, the entry indexed by $(i_1,\ldots,i_D)$ is denoted by $a_{i_1\cdots i_D}$, where $i_d = 1,\ldots,I_d$ for $d = 1,\ldots,D$.
The vectorization $\operatorname{vec}(\bm{\mathcal{A}}) \in \mathbb{R}^{I_1 I_2 \cdots I_D}$ of a tensor $\bm{\mathcal{A}} \in \mathbb{R}^{I_1 \times I_2 \times \cdots \times I_D}$ is the vector with $\operatorname{vec}(\bm{\mathcal{A}})_i = a_{i_1 i_2 \cdots i_D}$, where $i = i_1 + \sum_{d=2}^{D} (i_d - 1) \prod_{k=1}^{d-1} I_k$, i.e.\ the first index varies fastest.  The reshape operator $\bm{\mathcal{R}}\{\cdot\}_{I_1 \times I_2 \times \cdots \times I_D}$ maps a tensor to the prescribed dimensions $I_1 \times I_2 \times \cdots \times I_D$, preserving the total number of entries and incorporating any index permutations required by the adopted ordering convention. The operator $\operatorname{diag}(\cdot)$ returns a diagonal matrix when applied to a vector and extracts the main diagonal as a vector when applied to a matrix.
The Kronecker product of two matrices \( \mathbf{A} \in \mathbb{R}^{I \times J} \) and \( \mathbf{B} \in \mathbb{R}^{K \times L} \) is denoted by \( \mathbf{A} \otimes \mathbf{B} \in \mathbb{R}^{KI \times LJ} \). The Khatri-Rao product \( \mathbf{A} \khatri \mathbf{B} \) of the matrices \( \mathbf{A} \in \mathbb{R}^{I \times J} \) and \( \mathbf{B} \in \mathbb{R}^{K \times J} \) is an \( IK \times J \) matrix obtained by taking the column-wise Kronecker product. The identity matrix is conventionally denoted by \( \mathbf{I} \), with its dimensions either inferred from the context or explicitly stated as a subscript.

\begin{figure}
    \centering
    \scalebox{0.85}{\input{slices}}
    \caption{Horizontal  $\mathbf{W}^{(d)}_{r_d::}$, lateral $\mathbf{W}^{(d)}_{:m_d:}$, and frontal $\mathbf{W}^{(d)}_{::r_{d+1}}$ slices of a TT core $\bm{\mathcal{W}}^{(d)}$.}
    \label{fig:slices}
\end{figure}
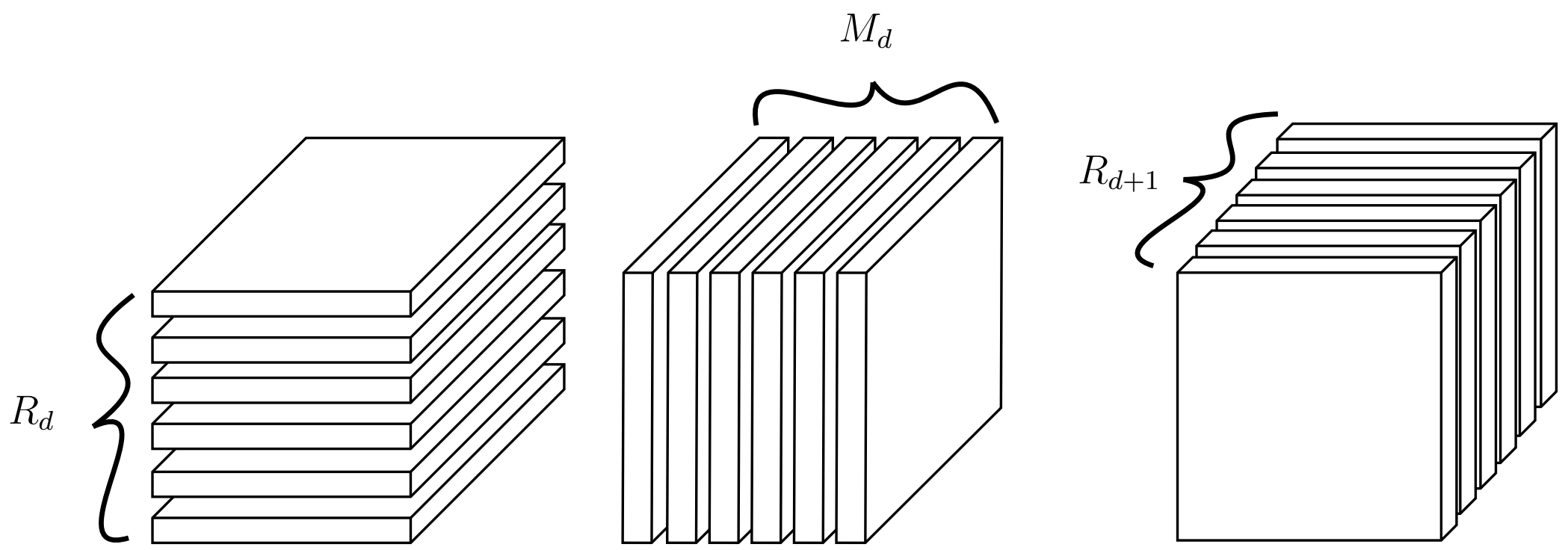

\begin{definition}[Tensor Train Decomposition]
\label{def:tt}
Tensor Train decomposition represents $\mathbf{w} \in \mathbb{R}^{M^D}$ using $D$ third-order tensors $\bm{\mathcal{W}}^{(d)} \in \mathbb{R}^{R_d \times M_d \times R_{d+1}}$, called TT cores
\begin{equation}
w_{m_1 m_2 \dots m_D} =
\sum_{r_1=1}^{R_1} \dots \sum_{r_{D+1}=1}^{R_{D+1}} w^{(1)}_{r_1 m_1 r_2} \dots w^{(D)}_{r_D m_D r_{D+1}}.
    \label{eq:3}
\end{equation}  
The auxiliary dimensions $R_1, R_2, \dots, R_{D+1}$ are called TT ranks. To ensure that the right-hand side of \eqref{eq:3} is a scalar, the boundary condition $R_1 = R_{D+1} = 1$ is required. TT decomposition is non-unique, with a storage complexity of $\mathcal{O}(R^2 M D)$ due to the $D$ tensor cores $\bm{\mathcal{W}}^{(d)}$.  
\end{definition}

\begin{definition}[Slices of a TT core]
Let $\bm{\mathcal{W}}^{(d)} \in \mathbb{R}^{R_d \times M_d \times R_{d+1}}$ be the $d$th
TT core in a Tensor Train decomposition. A slice of $\bm{\mathcal{W}}^{(d)}$ is a
two-dimensional subtensor (i.e., matrix) obtained by fixing one index. The horizontal, lateral, and
frontal slices are denoted by $\mathbf{W}^{(d)}_{r_d::}$, $\mathbf{W}^{(d)}_{:m_d:}$, and
$\mathbf{W}^{(d)}_{::r_{d+1}}$, respectively, as shown in Figure~\ref{fig:slices}.
\end{definition}

\begin{definition}[mode-$n$ unfolding of a TT core]
Let $\bm{\mathcal{W}}^{(d)} \in \mathbb{R}^{R_d \times M_d \times R_{d+1}}$ be a third-order
TT core. The mode-$n$ unfolding rearranges the entries of $\bm{\mathcal{W}}^{(d)}$ into a
matrix by stacking all slices obtained by fixing the $n$th index, while preserving a fixed
ordering of the remaining indices. The resulting matrix is denoted by
\[
\mathbf{W}^{(d)}_{(n)} \in \mathbb{R}^{I_n \times \prod_{m \neq n} I_m},
\qquad n \in \{1,2,3\},
\]
where $(I_1,I_2,I_3) := (R_d,\,M_d,\,R_{d+1})$.
\end{definition}

%% file: slices.tex
\tikzset{every picture/.style={line width=0.75pt}} 
\hspace*{-1.5cm}%
\begin{tikzpicture}[x=0.75pt,y=0.75pt,yscale=-1,xscale=1]

\draw  [fill={rgb, 255:red, 255; green, 255; blue, 255 }  ,fill opacity=1 ] (58.45,298.07) -- (119.1,237.42) -- (221.09,237.42) -- (221.09,247.28) -- (160.44,307.93) -- (58.45,307.93) -- cycle ; \draw   (221.09,237.42) -- (160.44,298.07) -- (58.45,298.07) ; \draw   (160.44,298.07) -- (160.44,307.93) ;
\draw  [fill={rgb, 255:red, 255; green, 255; blue, 255 }  ,fill opacity=1 ] (58.45,279.88) -- (119.1,219.23) -- (221.09,219.23) -- (221.09,229.08) -- (160.44,289.73) -- (58.45,289.73) -- cycle ; \draw   (221.09,219.23) -- (160.44,279.88) -- (58.45,279.88) ; \draw   (160.44,279.88) -- (160.44,289.73) ;
\draw  [fill={rgb, 255:red, 255; green, 255; blue, 255 }  ,fill opacity=1 ] (58.45,260.92) -- (119.1,200.28) -- (221.09,200.28) -- (221.09,210.13) -- (160.44,270.78) -- (58.45,270.78) -- cycle ; \draw   (221.09,200.28) -- (160.44,260.92) -- (58.45,260.92) ; \draw   (160.44,260.92) -- (160.44,270.78) ;
\draw  [fill={rgb, 255:red, 255; green, 255; blue, 255 }  ,fill opacity=1 ] (58.45,242.73) -- (119.1,182.08) -- (221.09,182.08) -- (221.09,191.94) -- (160.44,252.59) -- (58.45,252.59) -- cycle ; \draw   (221.09,182.08) -- (160.44,242.73) -- (58.45,242.73) ; \draw   (160.44,242.73) -- (160.44,252.59) ;
\draw  [fill={rgb, 255:red, 255; green, 255; blue, 255 }  ,fill opacity=1 ] (58.45,226.81) -- (119.1,166.16) -- (221.09,166.16) -- (221.09,176.02) -- (160.44,236.67) -- (58.45,236.67) -- cycle ; \draw   (221.09,166.16) -- (160.44,226.81) -- (58.45,226.81) ; \draw   (160.44,226.81) -- (160.44,236.67) ;
\draw  [fill={rgb, 255:red, 255; green, 255; blue, 255 }  ,fill opacity=1 ] (58.45,208.62) -- (119.1,147.97) -- (221.09,147.97) -- (221.09,157.82) -- (160.44,218.47) -- (58.45,218.47) -- cycle ; \draw   (221.09,147.97) -- (160.44,208.62) -- (58.45,208.62) ; \draw   (160.44,208.62) -- (160.44,218.47) ;
\draw  [fill={rgb, 255:red, 255; green, 255; blue, 255 }  ,fill opacity=1 ] (255.64,307.92) -- (309.22,254.61) -- (309.61,147.94) -- (298.15,147.93) -- (244.57,201.23) -- (244.18,307.9) -- cycle ; \draw   (309.61,147.94) -- (256.03,201.25) -- (255.64,307.92) ; \draw   (256.03,201.25) -- (244.57,201.23) ;
\draw  [fill={rgb, 255:red, 255; green, 255; blue, 255 }  ,fill opacity=1 ] (502.76,148.43) -- (508.83,142.36) -- (613,142.36) -- (613,248.13) -- (606.93,254.2) -- (502.76,254.2) -- cycle ; \draw   (613,142.36) -- (606.93,148.43) -- (502.76,148.43) ; \draw   (606.93,148.43) -- (606.93,254.2) ;
\draw  [fill={rgb, 255:red, 255; green, 255; blue, 255 }  ,fill opacity=1 ] (273.22,307.92) -- (326.8,254.61) -- (327.19,147.94) -- (315.73,147.93) -- (262.16,201.23) -- (261.77,307.9) -- cycle ; \draw   (327.19,147.94) -- (273.62,201.25) -- (273.22,307.92) ; \draw   (273.62,201.25) -- (262.16,201.23) ;
\draw  [fill={rgb, 255:red, 255; green, 255; blue, 255 }  ,fill opacity=1 ] (289.93,307.92) -- (343.51,254.61) -- (343.9,147.94) -- (332.44,147.93) -- (278.86,201.23) -- (278.47,307.9) -- cycle ; \draw   (343.9,147.94) -- (290.32,201.25) -- (289.93,307.92) ; \draw   (290.32,201.25) -- (278.86,201.23) ;
\draw  [fill={rgb, 255:red, 255; green, 255; blue, 255 }  ,fill opacity=1 ] (306.63,307.92) -- (360.21,254.61) -- (360.6,147.94) -- (349.14,147.93) -- (295.56,201.23) -- (295.17,307.9) -- cycle ; \draw   (360.6,147.94) -- (307.02,201.25) -- (306.63,307.92) ; \draw   (307.02,201.25) -- (295.56,201.23) ;
\draw  [fill={rgb, 255:red, 255; green, 255; blue, 255 }  ,fill opacity=1 ] (323.33,307.92) -- (376.91,254.61) -- (377.3,147.94) -- (365.84,147.93) -- (312.27,201.23) -- (311.88,307.9) -- cycle ; \draw   (377.3,147.94) -- (323.73,201.25) -- (323.33,307.92) ; \draw   (323.73,201.25) -- (312.27,201.23) ;
\draw  [fill={rgb, 255:red, 255; green, 255; blue, 255 }  ,fill opacity=1 ] (340.04,307.92) -- (393.62,254.61) -- (394.01,147.94) -- (382.55,147.93) -- (328.97,201.23) -- (328.58,307.9) -- cycle ; \draw   (394.01,147.94) -- (340.43,201.25) -- (340.04,307.92) ; \draw   (340.43,201.25) -- (328.97,201.23) ;
\draw  [fill={rgb, 255:red, 255; green, 255; blue, 255 }  ,fill opacity=1 ] (494.38,159.86) -- (500.45,153.79) -- (604.62,153.79) -- (604.62,259.56) -- (598.55,265.63) -- (494.38,265.63) -- cycle ; \draw   (604.62,153.79) -- (598.55,159.86) -- (494.38,159.86) ; \draw   (598.55,159.86) -- (598.55,265.63) ;
\draw  [fill={rgb, 255:red, 255; green, 255; blue, 255 }  ,fill opacity=1 ] (486.73,170.58) -- (492.79,164.52) -- (596.97,164.52) -- (596.97,270.29) -- (590.9,276.35) -- (486.73,276.35) -- cycle ; \draw   (596.97,164.52) -- (590.9,170.58) -- (486.73,170.58) ; \draw   (590.9,170.58) -- (590.9,276.35) ;
\draw  [fill={rgb, 255:red, 255; green, 255; blue, 255 }  ,fill opacity=1 ] (478.86,180.67) -- (484.92,174.6) -- (589.1,174.6) -- (589.1,280.37) -- (583.03,286.44) -- (478.86,286.44) -- cycle ; \draw   (589.1,174.6) -- (583.03,180.67) -- (478.86,180.67) ; \draw   (583.03,180.67) -- (583.03,286.44) ;
\draw  [fill={rgb, 255:red, 255; green, 255; blue, 255 }  ,fill opacity=1 ] (470.86,190.63) -- (476.93,184.56) -- (581.1,184.56) -- (581.1,290.33) -- (575.03,296.4) -- (470.86,296.4) -- cycle ; \draw   (581.1,184.56) -- (575.03,190.63) -- (470.86,190.63) ; \draw   (575.03,190.63) -- (575.03,296.4) ;
\draw  [fill={rgb, 255:red, 255; green, 255; blue, 255 }  ,fill opacity=1 ] (463.34,201.18) -- (469.41,195.11) -- (573.58,195.11) -- (573.58,300.88) -- (567.51,306.95) -- (463.34,306.95) -- cycle ; \draw   (573.58,195.11) -- (567.51,201.18) -- (463.34,201.18) ; \draw   (567.51,201.18) -- (567.51,306.95) ;
\draw [line width=1.5]    (35,262) .. controls (67,248) and (18,315) .. (49,306) ;
\draw [line width=1.5]    (35,262) .. controls (75,232) and (11,240) .. (51,210) ;
\draw [line width=1.5]    (343,126) .. controls (343.69,151.99) and (291,110) .. (297,143) ;
\draw [line width=1.5]    (343,126) .. controls (361.69,151.99) and (376.99,104.06) .. (392,142) ;
\draw [line width=1.5]    (465.69,164.42) .. controls (495.54,169.66) and (427.18,190.53) .. (453.82,198.5) ;
\draw [line width=1.5]    (465.69,164.42) .. controls (509.21,163.04) and (459.48,139.88) .. (503,138.5) ;

\draw (0,248.4) node [anchor=north west][inner sep=0.75pt]  [font=\large]  {$R_{d}$};
\draw (328,97.4) node [anchor=north west][inner sep=0.75pt]  [font=\large]  {$M_{d}$};
\draw (422.43,153.47) node [anchor=north west][inner sep=0.75pt]  [font=\large]  {$R_{d+1}$};

\node at (100,350) { (a) Horizontal: $\bm{W}^{(d)}_{r_d::} \in \mathbb{R}^{M_d \times R_{d+1}}$};
\node at (320,350) { (b) Lateral: $\bm{W}^{(d)}_{:m_d:} \in \mathbb{R}^{R_d \times R_{d+1}}$};
\node at (540,350) { (c) Frontal: $\bm{W}^{(d)}_{::r_{d+1}} \in \mathbb{R}^{R_d \times M_d}$};

\end{tikzpicture}

%% file: sections/3.model_learning.tex
\section{Probabilistic Tensor Train Kernel Machines}
\label{sec:model}
\subsection{Probabilistic model and priors}
\label{sec:priors}

 When $N$ inputs $\{\mathbf{x}_n\} ^ {N} _ {n=1}$ and outputs $\{y_n\} ^ {N} _ {n=1}$ are available for learning then model \eqref{eq:1} can be expressed as a linear matrix equation 

\begin{equation}
    \mathbf{y} = \mathbf{\Phi}^{\top} \mathbf{w} + \mathbf{e}
\end{equation}

 \noindent where $\mathbf{y} \in \mathbb{R}^{N}$ and $\mathbf{e} \in \mathbb{R}^N$, with $\mathbf{e} \sim \mathcal{N}(\mathbf{0}, \tau^{-1} \mathbf{I}_N)$. The matrix \( \mathbf{\Phi} \in \mathbb{R}^{M^D \times N} \) is constructed such that its \( n \)th column contains \( \bm{\varphi}(\mathbf{x}_n) \). Due to the tensor-product structure of the feature map in \eqref{eq:2}, $\mathbf{\Phi}$ can be expressed as a column-wise Khatri-Rao product of component matrices $\mathbf{\Phi} = \mathbf{\Phi}^{(1)} \khatri  \mathbf{\Phi}^{(2)} \khatri \dots \khatri  \mathbf{\Phi}^{(D)}$, where the columns of $\mathbf{\Phi}^{(d)} \in \mathbb{R}^{M_d \times N}$ are the feature vectors $\bm{\varphi}^{(d)}(x_n^{(d)})$, for all  \( n \in [1, N] \). Consequently, assuming a TT decomposed weight vector $\mathbf{w}$ along with the Gaussian noise model, the likelihood of the observed outputs is given by 
 
 \begin{equation}
    p(\mathbf{y} \mid \{\bm{\mathcal{W}}^{(d)}\}_{d=1}^{D}, \tau) = \prod_{n=1}^{N} \mathcal{N}\left(y_n \mid \bm{\varphi}(\mathbf{x}_n)^\top \mathbf{w}, \tau^{-1} \right),
    \label{likelihood}
\end{equation}
\noindent where the parameter $\tau$ denotes the noise precision. Each TT core $\bm{\mathcal{W}}^{(d)} \in \mathbb{R}^{R_d \times M_d \times R_{d+1}}$
can be considered through its two-dimensional slices obtained by fixing one index, namely
the horizontal, lateral, or frontal slices, as illustrated in
Figure~\ref{fig:slices}. The TT ranks $\{R_d\}_{d=2}^D$ and feature dimensions
$\{M_d\}_{d=1}^D$ control the model complexity, but selecting them a priori is difficult
and computationally expensive.

We therefore adopt a Bayesian model selection approach that infers the effective values of
$\{R_d\}_{d=2}^D$ and $\{M_d\}_{d=1}^D$ directly from the data. To this end, we place an
ARD-type hierarchical prior on the TT cores
$\{\bm{\mathcal{W}}^{(d)}\}_{d=1}^D$, with continuous precision parameters governing the
variances of their horizontal, lateral, and frontal slices. This prior shrinks
irrelevant slices toward zero, yielding simpler TT representations and allowing
automatic inference of model complexity during training. This approach is motivated by automatic relevance determination (ARD)
\citep{neal1996bayesian,mackay1994bayesian} and is related to Bayesian CP tensor
completion methods that infer the CP rank using similar priors
\citep{Zhao_2015_rank_det}.

\input{TT_model}

 More specifically, we introduce two sets of precisions, which act on the edges of Fig.~\ref{fig:tt}. The first set, the rank precisions $\bm{\lambda}_d := [\lambda_{1_d}, \lambda_{2_d}, \ldots, \lambda_{R_d}]$ for $d = 2,\ldots,D$, acts on the $D-1$ rank edges $R_2,\ldots,R_D$. Each parameter $\lambda_{r_d}$ controls the regularization strength of the $r_d$th \emph{horizontal} slice of $\bm{\mathcal{W}}^{(d)}$ and, at the same time, of the $r_d$th \emph{frontal} slice of the preceding core $\bm{\mathcal{W}}^{(d-1)}$ (Fig.~\ref{fig:slices}(a) and (c)). Since both slices belong to the same component of the edge $R_d$, $\bm{\lambda}_d$ couples the two adjacent cores, as also shown by the graphical model in Fig.~\ref{fig:TT_model}. The second set, the feature precisions $\bm{\delta}_d := [\delta_{1_d}, \delta_{2_d}, \ldots, \delta_{M_d}]$ for $d = 1,\ldots,D$, acts on the edge $M_d$ of the $d$th core. Each parameter $\delta_{m_d}$ controls the regularization strength of the $m_d$th \emph{lateral} slice of $\bm{\mathcal{W}}^{(d)}$ (Fig.~\ref{fig:slices}(b)) and therefore acts only within that core. The prior distribution of the vectorized TT cores is then a zero-mean Gaussian
\begin{equation*}
    p(\operatorname{vec}(\bm{\mathcal{W}}^{(d)})\mid \bm{\lambda}_{d}, \bm{\delta}_{d}, \bm{\lambda}_{d+1}) = \mathcal{N} \left(\operatorname{vec}(\bm{\mathcal{W}}^{(d)}) \mid \bm{0},  \bm{\Lambda}_{d+1}^{-1} \otimes \bm{\Delta}_{d}^{-1} \otimes \bm{\Lambda}_{d}^{-1} \right), \quad \forall d \in [1, D], 
\end{equation*}
where $\bm{\Lambda}_{d+1} \otimes \bm{\Delta}_{d} \otimes \bm{\Lambda}_{d} = \operatorname{diag}(\bm{\lambda}_{d+1}) \otimes \operatorname{diag}(\bm{\delta}_{d}) \otimes \operatorname{diag}(\bm{\lambda}_{d})$ represents the inverse covariance matrix. To allow TT ranks and feature dimensions to be inferred from data, we place Gamma hyperpriors over the rank precisions $\{\boldsymbol{\lambda}_{d}\}_{d=2}^{D}$ and the feature precisions $\{\boldsymbol{\delta}_{d}\}_{d=1}^D$.  The hyperprior over $\boldsymbol{\lambda}_{d}$ is factorized across the rank components \begin{equation}
    p(\bm{\lambda}_d) = \prod_{r_d=1}^{R_d} \text{Ga}(\lambda_{r_d} \mid c_0^{r_d}, e_0^{r_d}), 
    \label{lamba_prior}
\end{equation}

\noindent while the hyperprior over $\bm{\delta}_{d}$ is factorized over their individual feature components  
\begin{equation}
    p(\bm{\delta}_{d}) = \prod_{m_d=1}^{M_d} \text{Ga}(\delta_{m_d} \mid g_0^{m_d}, h_0^{m_d}), 
    \label{delta_prior}
\end{equation}

\noindent where $\text{Ga}(x \mid a, b) = b^a x^{a-1} e^{-bx}/\Gamma(a)$ denotes a Gamma distribution and $\Gamma(a)$ is the Gamma function. To complete our model with a fully Bayesian treatment, we also place a hyperprior over the noise precision
$\tau$ , that is,
\begin{equation}
        p(\tau) = \text{Ga}(\tau \mid a_0, b_0).
        \label{tau_prior}
\end{equation}

\noindent Denoting all latent variables by $\Theta
=
\bigl\{
\{\bm{\mathcal{W}}^{(d)}\}_{d=1}^D,
\{\bm{\delta}_d\}_{d=1}^D,
\{\bm{\lambda}_d\}_{d=2}^D,
\tau
\bigr\}$, the joint distribution $p(\mathbf{y}, \Theta)$ can be expressed as

\vspace{-0.5cm}

\begin{equation}
p \left( \mathbf{y} \mid \{\bm{\mathcal{W}}^{(d)}\}_{d=1}^D, \tau \right)
\prod_{d=1}^{D} p\left(\bm{\mathcal{W}}^{(d)} \mid  \bm{\lambda}_d, \bm{\delta}_d, \bm{\lambda}_{d+1}\right)\, p(\bm{\delta}_d)\,\prod_{d=2}^{D} p(\bm{\lambda}_d)\, p(\tau).
\label{the_joint_distribution}
\end{equation}

\noindent Figure~\ref{fig:TT_model} shows the probabilistic graphical model representing the joint distribution for the case where $\mathbf{w}$ is decomposed using the Tensor Train (TT) format. For simplicity, the deterministic design matrices $\{ \bm{\Phi}^{(d)} \}_{d=1}^D$ are not shown. The TT cores $\{\bm{\mathcal{W}}^{(d)}\}_{d=1}^D$, and the shape and rate hyperparameters of the Gamma distributions, represented by unbounded nodes in the figure, must be initialized before model learning.

\subsection{Model Learning}
\label{sec:learning}
The full posterior distribution of all variables in \( \Theta \) given the observed data is

\begin{equation}
    p(\Theta | \mathbf{y}) = \frac{p(\mathbf{y}, \Theta)}{\int p(\mathbf{y}, \Theta) d \Theta}. 
    \label{posterior_formula}
\end{equation}
Based on the posterior distribution of $\Theta$, the predictive distribution of an unseen output $y_i$ can be inferred by

\begin{equation}
     p({y_i} \mid \mathbf{y}) = \int p\left({y_i} \mid \Theta\right) p\left(\Theta \mid \mathbf{y} \right)d\Theta.
     \label{general_predictive ditrsibution} 
\end{equation}
An exact Bayesian inference of \eqref{posterior_formula} and \eqref{general_predictive ditrsibution} requires integrating over all latent variables, making it analytically intractable. Therefore, as presented in \citet{kilic_interpretable_2025}, we perform Bayesian learning using mean-field variational inference \citep{winn_variational_2005}. The key idea is to approximate the true posterior $p(\Theta \mid \mathbf{y})$ with a variational distribution $q(\Theta)$ by minimizing the Kullback--Leibler (KL) divergence. This is equivalent to maximizing a lower bound $\mathcal{L}(q)$ on the log model evidence $\ln p(\mathbf{y})$, defined as $\mathcal{L}(q) = \mathbb{E}_{q(\Theta)}[\ln p(\mathbf{y},\Theta)] + H(q(\Theta))$, where $H(q(\Theta))$ denotes the entropy of the variational distribution.  The maximum of the lower bound occurs when the KL divergence vanishes, implying \( q(\Theta) = p(\Theta \mid \mathbf{y}) \). The mean-field assumption factorizes the variational distribution over each variable $\theta_j \in \bm{\Theta}$ as

\begin{equation}
q(\Theta)
=
\prod_{d=1}^{D}
\left[
q_{\bm{\mathcal{W}}^{(d)}}(\bm{\mathcal{W}}^{(d)})
\, q_{\bm{\delta}_d}(\bm{\delta}_d)
\right]
\prod_{d=2}^{D}
q_{\bm{\lambda}_d}(\bm{\lambda}_d)q_{\tau}(\tau).
\label{variational_distribution}
\end{equation}
In other words, this factorization assumes that all variables $\theta_j \in \bm{\Theta}$ are independent. The functional form of each factor can then be derived analytically in turn. Specifically, the optimal $q_j(\theta_j)$ is obtained by maximizing $\mathcal{L}(q)$, with the maximum occurring when

\begin{equation}
    \ln q_j(\theta_j) = \mathbb{E}_{q(\Theta \setminus \theta_j)}[\ln p(\mathbf{y}, \Theta)] + \text{const},
    \label{update_rule}
\end{equation}
where \( \mathbb{E}_{q(\Theta \setminus \theta_j)} [\cdot] \) denotes the expectation taken with respect to the variational distributions of all variables except \( \theta_j \). Since all distributions in the model belong to the exponential family and each prior is conjugate to the corresponding likelihood term when all other variables are held fixed, \eqref{update_rule} yields  closed-form posterior updates for each factor. Model learning proceeds by initializing each $q_j(\theta_j)$ and iteratively updating them using the update rules until convergence. In the following, we present the posterior update rules for each $q_j(\theta_j)$, for all $\theta_j \in \bm{\Theta}$. 

\subsubsection{Posterior distribution of the TT cores}

To derive the update rule for the $d$th TT core $\boldsymbol{\mathcal{W}}^{(d)}$, we rewrite the model to isolate its dependence on this core.

\begin{theorem} 
Let $\mathbf{w}$ be represented in TT format by the cores $\{\bm{\mathcal{W}}^{(k)}\}_{k=1}^{D}$. For any fixed $d \in [1, D]$, the model output is linear in the $d$th core,
\begin{equation}
\mathbf{\Phi}^{\top}\mathbf{w}
=
\mathbf{G}_{\backslash d}^{\top} \operatorname{vec}(\bm{\mathcal{W}}^{(d)}),
\label{eq:4}
\end{equation}
where
\begin{equation}
\mathbf{G}_{\backslash d}
=
\mathbf{G}_{>d}
\khatri
\mathbf{\Phi}^{(d)}
\khatri
\mathbf{G}_{<d}
\in \mathbb{R}^{R_d M_d R_{d+1} \times N}.
\end{equation}
and \( \operatorname{vec}(\bm{\mathcal{W}}^{(d)}) \in \mathbb{R}^{R_d M_d R_{d+1}}\) is the column-wise vectorization of the \( d \)th TT core. The matrix \(\mathbf{G}_{\backslash d}\) can be interpreted as the design matrix corresponding to \( \operatorname{vec}(\bm{\mathcal{W}}^{(d)})\).
\label{linearity_theorem}
\end{theorem}
\begin{proof}
See Section 1 of the Appendix.
\end{proof}

The matrices $\mathbf{G}_{<d}$ and $\mathbf{G}_{>d}$ in
Theorem~\ref{linearity_theorem} collect the contributions of all modes
preceding and following the $d$th mode, respectively. Specifically,
$\mathbf{G}_{<d}$ aggregates information from modes $1$ through $d-1$,
while $\mathbf{G}_{>d}$ aggregates information from modes $d+1$ through $D$.
Using the mode-$3$ and mode-$1$ unfoldings of the TT cores
$\bm{\mathcal{W}}^{(k)} \in
\mathbb{R}^{R_k \times M_k \times R_{k+1}}$, denoted by
$\mathbf{W}^{(k)}_{(3)}$ and $\mathbf{W}^{(k)}_{(1)}$, respectively,
we introduce forward and backward accumulators that allow efficient
sequential contraction of all TT cores except the $d$th one.

\begin{definition}[Forward and backward accumulators]
Given the boundary condition $R_1 = R_{D+1} = 1$, the forward and
backward accumulators are defined recursively by
\begin{equation}
\begin{aligned}
\mathbf{G}_{<1} &:= \mathbf{1}_N^{\top}, \qquad
\mathbf{G}_{<k}
:=
\mathbf{W}^{(k-1)}_{(3)}
\bigl(
\mathbf{\Phi}^{(k-1)}
\khatri
\mathbf{G}_{<k-1}
\bigr)
\in \mathbb{R}^{R_k \times N},
\quad k=2,\ldots,D+1,
\\
\mathbf{G}_{>D} &:= \mathbf{1}_N^{\top}, \qquad
\mathbf{G}_{>k}
:=
\mathbf{W}^{(k+1)}_{(1)}
\bigl(
\mathbf{G}_{>k+1}
\khatri
\mathbf{\Phi}^{(k+1)}
\bigr)
\in \mathbb{R}^{R_{k+1} \times N},
\quad k=D-1,\ldots,1.
\end{aligned}
\end{equation}
\label{def:accumulators}
\end{definition}

\noindent The variational posterior of
$\operatorname{vec}(\bm{\mathcal{W}}^{(d)})$ is Gaussian,
\begin{equation}
q(\operatorname{vec}(\bm{\mathcal{W}}^{(d)})) =
\mathcal{N}\!\left(
\operatorname{vec}(\bm{\mathcal{W}}^{(d)})
\mid
\operatorname{vec}(\tilde{\bm{\mathcal{W}}}^{(d)}),
\mathbf{\Sigma}^{(d)}
\right),
\end{equation}
with mean and covariance
\begin{equation}
\begin{aligned}
\operatorname{vec}(\tilde{\bm{\mathcal{W}}}^{(d)})
&=
\mathbb{E}_q[\tau]\,
\mathbf{\Sigma}^{(d)}\,
\mathbb{E}_q[\mathbf{G}_{\backslash d}]\,\mathbf{y},
\\
\mathbf{\Sigma}^{(d)}
&=
\Big(
\mathbb{E}_q[\tau]\,
\mathbb{E}_q[
\mathbf{G}_{\backslash d}\mathbf{G}_{\backslash d}^{\top}
]
+
\mathbb{E}_q[\bm{\Lambda}_{d+1}]
\otimes
\mathbb{E}_q[\bm{\Delta}_d]
\otimes
\mathbb{E}_q[\bm{\Lambda}_{d}]
\Big)^{-1}.
\end{aligned}
\label{posterior_updates_cores}
\end{equation}
See Section 3 of the Appendix for a detailed derivation of \eqref{posterior_updates_cores}. The matrix
$\mathbb{E}_q[
\mathbf{G}_{\backslash d}\mathbf{G}_{\backslash d}^{\top}
]$
represents the expected Gram matrix of the model fitting term
$\mathbf{\Phi}^{\top} \mathbf{w}$ in \eqref{eq:4}, excluding the $d$th
core $\bm{\mathcal{W}}^{(d)}$. Equivalently, it corresponds to the
expected second moment of the design matrix
$\mathbf{G}_{\backslash d}$, which combines the features
$\mathbf{\Phi}^{(d)}$ with all TT cores
$\bm{\mathcal{W}}^{(k)}$ for $k \neq d$.
This term cannot be computed straightforwardly, and therefore,
we first need to introduce the following results.
In order to express
$\mathbb{E}_q[
\mathbf{G}_{\backslash d}\mathbf{G}_{\backslash d}^{\top}
]$
in terms of the posterior parameters of the TT cores in
\eqref{posterior_updates_cores}, we reformulate it by isolating
the random variables in the expression as in the following theorem.

\begin{theorem}
For any fixed $d \in \{1,\ldots,D\}$, and assuming that the TT cores $\bm{\mathcal{W}}^{(k)}$ are mutually independent under $q$ for all $k \neq d$, the expected Gram matrix factorizes into left and right contributions as
\begin{equation}
\mathbb{E}_q\!\left[
\mathbf{G}_{\backslash d}\mathbf{G}_{\backslash d}^{\top}
\right]
=
\bm{\mathcal{R}}
\left\{
\left(
\mathbb{E}_q[
\mathbf{G}_{<d} \khatri \mathbf{G}_{<d}
]
\khatri
\mathbf{\Phi}^{(d)}
\right)
\left(
\mathbf{\Phi}^{(d)}
\khatri
\mathbb{E}_q[
\mathbf{G}_{>d} \khatri \mathbf{G}_{>d}
]
\right)^{\!\top}
\right\}_{R_d M_d R_{d+1}\times R_d M_d R_{d+1}}
\end{equation}
\end{theorem}

\begin{proof}
See Section 4 of the Appendix.
\end{proof}

\noindent The matrices
$\mathbb{E}_q[\mathbf{G}_{<d} \khatri \mathbf{G}_{<d}]$ and
$\mathbb{E}_q[\mathbf{G}_{>d} \khatri \mathbf{G}_{>d}]$
represent second-order summaries of the partial contractions
to the left and right of core $d$, respectively.
To evaluate these matrices efficiently, we use the mode-$3$
and mode-$1$ unfoldings of the TT cores and introduce the
\textit{forward} and \textit{backward} Gram matrices.

\begin{definition}[Forward and backward Gram matrices]
We define the forward and backward Gram matrices as
$\mathbf{H}_{<k} :=
\mathbb{E}_q[\mathbf{G}_{<k} \khatri \mathbf{G}_{<k}]$
and
$\mathbf{H}_{>k} :=
\mathbb{E}_q[\mathbf{G}_{>k} \khatri \mathbf{G}_{>k}]$,
respectively, where
$\mathbf{H}_{<k}\in\mathbb{R}^{R_k^2\times N}$ and
$\mathbf{H}_{>k}\in\mathbb{R}^{R_{k+1}^2\times N}$.
These matrices summarize the contributions of all TT cores
located to the left and right of the $k$th core.

Under the boundary conditions $R_1 = R_{D+1} = 1$ and assuming
that the TT cores $\bm{\mathcal{W}}^{(k)}$ are mutually
independent, we initialize the recursions with
$\mathbf{H}_{<1} := \mathbf{1}_N^{\top}$ and
$\mathbf{H}_{>D} := \mathbf{1}_N^{\top}$.
The forward and backward Gram matrices then satisfy the
following recursive relations:
\begin{equation}
\begin{aligned}
\mathbf{H}_{<k}
&=
\mathbb{E}_q\!\left[
\mathbf{W}^{(k-1)}_{(3)}
\otimes
\mathbf{W}^{(k-1)}_{(3)}
\right] \mathbf{P}
\bigl(
(\mathbf{\Phi}^{(k-1)} \khatri \mathbf{\Phi}^{(k-1)})
\khatri
\mathbf{H}_{<k-1}
\bigr),
&& k = 2,\ldots,D+1,
\\
\mathbf{H}_{>k}
&=
\mathbb{E}_q\!\left[
\mathbf{W}^{(k+1)}_{(1)}
\otimes
\mathbf{W}^{(k+1)}_{(1)}
\right]\mathbf{P}
\bigl(
\mathbf{H}_{>k+1}
\khatri
(\mathbf{\Phi}^{(k+1)} \khatri \mathbf{\Phi}^{(k+1)})
\bigr),
&& k = D-1,\ldots,1.
\end{aligned}
\label{gram_matrix_definition}
\end{equation}
\end{definition}
 where $\mathbf{P}$ denotes the fixed permutation matrix that reorders the
Kronecker factors according to the adopted index ordering. A derivation
of the recursive relations in \eqref{gram_matrix_definition} is provided
in Section~5 of the Appendix. The expectation terms are evaluated as
\begin{equation}
\begin{aligned}
\mathbb{E}_q\!\left[
\mathbf{W}^{(k-1)}_{(3)}
\otimes
\mathbf{W}^{(k-1)}_{(3)}
\right]
&=
\mathbb{E}_q\!\left[\mathbf{W}^{(k-1)}_{(3)}\right]
\otimes
\mathbb{E}_q\!\left[\mathbf{W}^{(k-1)}_{(3)}\right]
+
\operatorname{Cov}\!\left[
\mathbf{W}^{(k-1)}_{(3)}
\otimes
\mathbf{W}^{(k-1)}_{(3)}
\right],
\\
\mathbb{E}_q\!\left[
\mathbf{W}^{(k+1)}_{(1)}
\otimes
\mathbf{W}^{(k+1)}_{(1)}
\right]
&=
\mathbb{E}_q\!\left[\mathbf{W}^{(k+1)}_{(1)}\right]
\otimes
\mathbb{E}_q\!\left[\mathbf{W}^{(k+1)}_{(1)}\right]
+
\operatorname{Cov}\!\left[
\mathbf{W}^{(k+1)}_{(1)}
\otimes
\mathbf{W}^{(k+1)}_{(1)}
\right].
\end{aligned}
\label{eq:expected_kronecker}
\end{equation}
The identities in \eqref{eq:expected_kronecker} follow directly from the standard second-moment relation
$\mathbb{E}[x^2]=\mathbb{E}[x]^2+\operatorname{Var}[x]$, where the covariance terms account for the posterior uncertainty of the TT cores required in the forward and backward Gram matrix recursions. When these covariance terms vanish, the other TT cores are treated as deterministic, and the posterior mean update reduces to a regularized ALS update for the $d$th TT core $\bm{\mathcal{W}}^{(d)}$.

To evaluate these covariance terms in terms of the posterior parameters obtained from \eqref{posterior_updates_cores}, we use the following lemma.

\begin{lemma}
\label{lemma1}
Let $\mathbf{W}^{(k)} \in \mathbb{R}^{A \times B}$ and let
$\operatorname{vec}(\mathbf{W}^{(k)}) \in \mathbb{R}^{AB}$ denote its
column-wise vectorization. Then,
\begin{equation}
\mathbf{W}^{(k)} \otimes \mathbf{W}^{(k)}
=
\bm{\mathcal{R}}\!\left\{
\operatorname{vec}(\mathbf{W}^{(k)})
\operatorname{vec}(\mathbf{W}^{(k)})^\top
\right\}_{A^2\,\times B^2}.
\label{eq:25}
\end{equation}
That is, the Kronecker product $\mathbf{W}^{(k)} \otimes \mathbf{W}^{(k)}$ can be obtained
by reshaping the outer product of the vectorized matrix into a matrix of size
$A^2 \times B^2$.
\end{lemma}

The intuitive interpretation of Lemma~\ref{lemma1} is that the Kronecker product of \( \mathbf{W}^{(k)} \) with itself is the same as the outer product of $\operatorname{vec}(\bm{\mathcal{W}}^{(k)})$ with itself, but reshaped into a structured block form. Using this identity, the covariance term in \eqref{eq:expected_kronecker} can be computed by reshaping  the posterior covariance matrices of the corresponding TT cores as

\begin{equation}
\begin{aligned}
\operatorname{Cov}\!\left[
\mathbf{W}^{(k-1)}_{(3)}
\otimes
\mathbf{W}^{(k-1)}_{(3)}
\right]
&=
\mathcal{R}\!\left\{
\mathbf{\Sigma}^{(k-1)}
\right\}_{R_k^2 \times\,
(R_{k-1}M_{k-1})^2}
\\
\operatorname{Cov}\!\left[
\mathbf{W}^{(k+1)}_{(1)}
\otimes
\mathbf{W}^{(k+1)}_{(1)}
\right]
&=
\mathcal{R}\!\left\{
\mathbf{\Sigma}^{(k+1)}
\right\}_{R_{k+1}^2\times\,
(M_{k+1}R_{k+2})^2}.
\end{aligned}
\label{eq:26}
\end{equation}

\subsubsection{\texorpdfstring{Posterior distributions of $\bm{\lambda_d}$ and $\bm{\delta}_d$}
{Posterior of Lambda and Delta}}

As illustrated in Figure \ref{fig:TT_model}, inference over $\bm{\lambda}_d$ is performed by gathering information from two cores that are $\bm{\mathcal{W}}^{(d-1)}$ and  $\bm{\mathcal{W}}^{(d)}$ together with its hyperprior.  Using \eqref{update_rule}, the posterior distributions of $\lambda_{r_d}$, for all $r_d \in [1, R_d]$, are obtained as independent Gamma distributions,

\begin{equation}
    q_{\bm{\lambda}_d}(\bm{\lambda}_d) = \prod_{r_d=1}^{R_d} \text{Ga}(\lambda_{r_d} \mid c_N^{r_d}, e_N^{r_d}), 
\end{equation}
where $c_N^{r_d}, e_N^{r_d}$ are the posterior parameters learned from $N$ training points and updated by

\begin{equation}
\begin{aligned}
\mathbf{c}_N^{d}
&= \mathbf{c}_0^{d}
   + \tfrac{1}{2}\!\left(
      R_{d-1} M_{d-1}
      + M_d R_{d+1} 
   \right), \\[4pt]
\mathbf{e}_N^{d}
&= \mathbf{e}_0^{d}
   + \tfrac{1}{2} \, \text{diag} \Big(
      \mathbb{E}_{q}\!\left[
        \mathbf{W}^{(d-1)}_{(3)}
        (\bm{\Delta}_{d-1}\kronecker\bm{\Lambda}_{d-1})
        \mathbf{W}^{(d-1)\top}_{(3)}
      \right] \\
&\hspace{3.8em}
      + \mathbb{E}_{q}\!\left[
        \mathbf{W}^{(d)}_{(1)}
        (\bm{\Lambda}_{d+1} \kronecker \bm{\Delta}_{d})
        \mathbf{W}^{(d)\top}_{(1)}
      \right]
   \Big).
\end{aligned}
\label{posterior_updates_lambda}
\end{equation}
Section~7 of the Appendix provides a detailed derivation of
\eqref{posterior_updates_lambda}. The two expectations appearing in the update of
$\mathbf{e}_N^{d}$ are evaluated in closed form using the posterior mean and covariance of
the TT cores given in \eqref{posterior_updates_cores}. For an efficient evaluation of these
expectations, it is convenient to explicitly collect the posterior variances of the TT-core
entries. The diagonal of the covariance matrix $\bm{\Sigma}^{(d)}$ contains the marginal
variances of the entries of the $d$-th TT core, ordered consistently with
$\operatorname{vec}(\bm{\mathcal{W}}^{(d)})$. We therefore define a variance tensor
$\bm{\mathcal{V}}^{(d)} \in \mathbb{R}^{R_d \times M_d \times R_{d+1}}$ by reshaping this
vector as
\[
\bm{\mathcal{V}}^{(d)}
:=
\bm{\mathcal{R}}\!\left\{
\operatorname{diag}(\bm{\Sigma}^{(d)})
\right\}_{R_d\times M_d\times R_{d+1}}.
\]
The diagonal of the \emph{first} expected inner product,
$\mathbb{E}_{q}\!\left[
        \mathbf{W}^{(d-1)}_{(3)}
        (\bm{\Delta}_{d-1}\kronecker\bm{\Lambda}_{d-1})
        \mathbf{W}^{(d-1)\top}_{(3)}
      \right]$,
is computed as
\begin{equation}
\label{eq:eval_exp_left}
\operatorname{diag}\!\left(
\tilde{\mathbf{W}}^{(d-1)}_{(3)}
\left(\mathbb{E}_{q}\!\left[\bm{\Delta}_{d-1}\right]\kronecker\mathbb{E}_{q}\!\left[\bm{\Lambda}_{d-1}\right]\right)
\tilde{\mathbf{W}}^{(d-1)\top}_{(3)}
\right)
+
\mathbf{V}^{(d-1)}_{(3)}(\mathbb{E}_{q}\!\left[\bm{\delta}_{d-1}\right]\kronecker\mathbb{E}_{q}\!\left[\bm{\lambda}_{d-1}\right]).
\end{equation}
Similarly, the diagonal of the \emph{second} expected inner product,
$\mathbb{E}_{q}\!\left[
        \mathbf{W}^{(d)}_{(1)}
        (\bm{\Lambda}_{d+1}\kronecker\bm{\Delta}_{d})
        \mathbf{W}^{(d)\top}_{(1)}
      \right]$,
is computed as
\begin{equation}
\label{eq:eval_exp_right}
\operatorname{diag}\!\left(
\tilde{\mathbf{W}}^{(d)}_{(1)}\left(
\mathbb{E}_{q}\!\left[\bm{\Lambda}_{d+1}\right]\kronecker\mathbb{E}_{q}\!\left[\bm{\Delta}_{d}\right]\right)
\tilde{\mathbf{W}}^{(d)\top}_{(1)}
\right)
+
\mathbf{V}^{(d)}_{(1)}(\mathbb{E}_{q}\!\left[\bm{\lambda}_{d+1}\right]\kronecker\mathbb{E}_{q}\!\left[\bm{\delta}_{d}\right]).
\end{equation}
Equations \eqref{eq:eval_exp_left} and \eqref{eq:eval_exp_right} are then substituted into
\eqref{posterior_updates_lambda} to obtain the update of $\mathbf{e}_N^{d}$.
The posterior expectation can be obtained by
$\mathbb{E}_q[\bm{\lambda}_d]
= [c_N^{1_d}/e_N^{1_d},\ldots,c_N^{R_d}/e_N^{R_d}]^{\top}$,
and thus $\mathbb{E}_q[\bm{\Lambda}_d]
= \operatorname{diag}(\mathbb{E}_q[\bm{\lambda}_d])$,
for $d=2,\ldots,D$.
Each $\lambda_{r_d}$ jointly controls the $r_d$th frontal slice
$\mathbf{W}^{(d-1)}_{::r_d}$ of core $\bm{\mathcal{W}}^{(d-1)}$
and the $r_d$th horizontal slice $\mathbf{W}^{(d)}_{r_d::}$
of core $\bm{\mathcal{W}}^{(d)}$. These slices correspond to the
same component of their shared rank dimension $R_d$.
The update of $\lambda_{r_d}$ depends on the weighted squared entries
of both slices, including their posterior variances.
Smaller weighted second moments yield a smaller $e_N^{r_d}$ and hence
a larger $\mathbb{E}_q[\lambda_{r_d}]$, increasing shrinkage of both
slices and promoting a lower effective TT rank $R_d$.

As shown in Figure \ref{fig:TT_model}, the inference of $\bm{\delta}_d$ is performed by gathering information from the $d$th TT core and incorporating information from its hyperprior. By applying \eqref{update_rule}, we can identify the posteriors of $\delta_{m_d}$, $\forall d \in [1, D]$ and $\forall m_d \in [1, M_d]$ as an independent Gamma distribution,

\vspace{-0.5cm}
\begin{equation}
    q_{\bm{\delta_{d}}}(\bm{\delta}_{d}) = \prod_{m_d=1}^{M_d} \text{Ga}(\delta_{m_d} \mid g^{m_d}_N, h^{m_d}_N),
    \label{eq:32}
\end{equation}

\noindent where $g^{m_d}_N, h^{m_d}_N$ are the posterior parameters learned from $N$ training points and updated by 
\begin{equation}
    \begin{aligned}
        \bm{g}^{d}_N &=  \bm{g}^{d}_0 + \frac{1}{2}(R_dR_{d+1}),\\
        \bm{h}^{d}_N &=  \bm{h}^{d}_0 + \frac{1}{2} \, \text{diag} \left(\mathbb{E}_q \left[  \mathbf{W}^{(d)}_{(2)} (\bm{\Lambda}_{d+1} \kronecker \bm{\Lambda}_{d} )\mathbf{W}^{(d)\top}_{(2)}\right]\right).
    \end{aligned}
    \label{posterior_updates_delta}
\end{equation}
See Section 6 of the Appendix for a detailed derivation of \eqref{posterior_updates_delta}. The expectation of the inner product of the $m_d$th lateral slice of the $d$th core w.r.t $q$ distribution can be computed using the posterior parameters of $\bm{\mathcal{W}}^{(d)}$ in \eqref{posterior_updates_cores} as

\begin{equation}
    \operatorname{diag}\left( \tilde{\mathbf{W}}^{(d)}_{(2)} \left(\mathbb{E}_q \left[\bm{\Lambda}_{d+1} \right]\kronecker \mathbb{E}_q \left[\bm{\Lambda}_{d}\right] \right)\tilde{\mathbf{W}}^{(d)\top}_{(2)}\right) + \mathbf{V}^{(d)}_{(2)}(\mathbb{E}_{q}\!\left[\bm{\lambda}_{d+1} \right]\kronecker \mathbb{E}_{q}\!\left[\bm{\lambda}_{d}\right] )
    \label{square_term_h_N}
\end{equation}
which is then substituted into
\eqref{posterior_updates_delta} to obtain the update of $\mathbf{h}_N^{d}$.
 The posterior expectation can be obtained by
$\mathbb{E}_q[\bm{\delta}_d]
= [g_N^{1_d}/h_N^{1_d},\ldots,g_N^{M_d}/h_N^{M_d}]^{\top}$,
and thus $\mathbb{E}_q[\bm{\Delta}_d]
= \operatorname{diag}(\mathbb{E}_q[\bm{\delta}_d])$,
for $d=1,\ldots,D$.
Each $\delta_{m_d}$ controls the relevance of the $m_d$th feature component
through the lateral slice $\mathbf{W}^{(d)}_{:m_d:}$ of core
$\bm{\mathcal{W}}^{(d)}$. Its update depends on the weighted squared
entries of this slice, including their posterior variances.
Smaller weighted second moments yield a smaller $h_N^{m_d}$ and hence
a larger $\mathbb{E}_q[\delta_{m_d}]$, increasing shrinkage of the
slice and suppressing the corresponding feature component.

The updates of $\bm{\lambda}_d$ and $\bm{\delta}_d$ are interdependent.
Each $\lambda_{r_d}$ regularizes a pair of matching slices in two adjacent
cores, while each $\delta_{m_d}$ regularizes a lateral slice within one core.
The rank-precision updates in \eqref{posterior_updates_lambda} depend on
the feature precisions, and the feature-precision updates in
\eqref{posterior_updates_delta} depend on both adjacent rank precisions.
For fixed posterior core moments, larger rank precisions increase
$h_N^{m_d}$ and reduce $\mathbb{E}_q[\delta_{m_d}]$, weakening feature
shrinkage. Conversely, larger feature precisions increase the corresponding
$e_N^{r_d}$ and reduce $\mathbb{E}_q[\lambda_{r_d}]$, weakening rank
shrinkage. This compensating interaction balances shrinkage across
rank and feature components.

\subsubsection{\texorpdfstring{Posterior distribution of noise precision $\tau$}{Posterior distribution of noise precision tau}}
The noise precision $\tau$ can be inferred by receiving information from observed data and its co-parents, including all the TT cores, and incorporating the information from its hyperprior. Applying \eqref{update_rule}, the variational posterior is a Gamma distribution, given by 

\begin{equation}
    q_{\tau}(\tau) = \text{Ga}(\tau \mid a_N, b_N),
\end{equation}

\noindent where the posterior parameters are updated by

\begin{equation}
    \begin{aligned}
        a_N &= a_0 + \frac{N}{2}, \\
        b_N &= b_0 + \frac{1}{2} \mathbb{E}_q \left[\| \mathbf{y} - \mathbf{\Phi}^{\top} \mathbf{w}\|_F^{2}\right].
 \end{aligned}
 \label{eq:20}
 \end{equation}   
See Section 8 of the Appendix for a detailed derivation of \eqref{eq:20}. The posterior expectation of model residuals $\mathbb{E}_q \left[\| \mathbf{y} - \mathbf{\Phi}^\top \mathbf{w}\|_F^{2}\right]$ in Equation (\ref{eq:20}) cannot be computed straightforwardly and, therefore, we need to introduce the following results. When no core is excluded, the full model is obtained by applying the forward accumulator recursion through all $D$ modes. Starting from $\mathbf{G}_{<1}= \bm{1}_N^{\top}$, the forward contraction propagates sequentially across the entire tensor train, yielding
\begin{equation}
   \mathbf{w}^{\top}\mathbf{\Phi} = \mathbf{G}_{<D+1} \in \mathbb{R}^{1 \times N},
\end{equation}
since $R_{D+1}=1$. Expanding the squared norm gives
\begin{equation}
    \mathbb{E}_q\!\left[\|\mathbf{y} - \mathbf{\Phi}^{\top}\mathbf{w}\|_F^{2}\right]
    = \|\mathbf{y}\|_F^{2}
    - 2\,\mathbb{E}_q\!\left[\mathbf{G}_{<D+1}\right]\mathbf{y}
    + \mathbb{E}_q\!\left[\mathbf{G}_{<D+1}\mathbf{G}_{<D+1}^{\top}\right].
    \label{eq:residual_expansion}
\end{equation}
To evaluate the last term, note that for a row vector $\mathbf{G}_{<D+1}\in\mathbb{R}^{1\times N}$ the Khatri--Rao product $\mathbf{G}_{<D+1}\khatri\mathbf{G}_{<D+1}\in\mathbb{R}^{1\times N}$ contains the squared entries of $\mathbf{G}_{<D+1}$. Hence
\begin{equation}
    \mathbb{E}_q\!\left[\mathbf{G}_{<D+1}\mathbf{G}_{<D+1}^{\top}\right]
    = \mathbb{E}_q\!\left[\mathbf{G}_{<D+1}\khatri\mathbf{G}_{<D+1}\right]\mathbf{1}_N
    = \mathbf{H}_{<D+1}\,\mathbf{1}_N. 
    \label{eq:khatri_identity}
\end{equation}
Here $\mathbf{H}_{<D+1}\in\mathbb{R}^{1\times N}$ is the forward Gram matrix propagated through all modes without excluding any core. It follows the same recursion as \eqref{gram_matrix_definition}, initialized with $\mathbf{H}_{<1}=\mathbf{1}_N^{\top}$ and run for $k=2,\ldots,D+1$. Substituting \eqref{eq:khatri_identity} into \eqref{eq:residual_expansion} yields
\begin{equation}
    b_N = b_0 + \frac{1}{2}\Big(
        \|\mathbf{y}\|_F^{2}
        - 2\,\mathbb{E}_q[\mathbf{G}_{<D+1}]\,\mathbf{y}
        + \mathbf{H}_{<D+1}\mathbf{1}_N
    \Big).
\end{equation}
Finally, the posterior expectation of the noise precision is $\mathbb{E}_q[\tau] = a_N / b_N$.

\subsubsection{Lower bound model evidence}
\label{sec:elbo}
The mean-field variational inference framework maximizes the evidence lower
bound (ELBO). For fixed ranks and feature dimensions, exact coordinate updates
ensure that the ELBO is non-decreasing across iterations, allowing it to be
used as a convergence criterion. The lower bound of the log marginal
likelihood is defined by
\begin{equation}
    \mathcal{L}(q)
    =
    \mathbb{E}_{q(\bm{\Theta})}
    [\ln p(\mathbf{y},\bm{\Theta})]
    + H(q(\bm{\Theta})),
    \label{eq:42}
\end{equation}
where the first term denotes the posterior expectation of the log-joint
distribution, and the second term denotes the entropy of the variational
posterior. Evaluating these terms using the Gaussian core posteriors and
Gamma precision posteriors yields
{\fontsize{10pt}{11pt}\selectfont
\begin{equation}
\begin{aligned}
\mathcal{L}(q)
&=
-\frac{a_N}{2b_N}
\mathbb{E}_q\!\left[
\|\mathbf{y}-\mathbf{\Phi}^{\top}\mathbf{w}\|_F^2
\right]
\\
&\quad
-\frac{1}{2}\sum_{d=1}^{D}
\operatorname{Tr}\!\Biggl\{
\bigl(
\tilde{\bm{\Lambda}}_{d+1}
\otimes
\tilde{\bm{\Delta}}_d
\otimes
\tilde{\bm{\Lambda}}_d
\bigr)
\Bigl(
\operatorname{vec}(\tilde{\bm{\mathcal{W}}}^{(d)})
\operatorname{vec}(\tilde{\bm{\mathcal{W}}}^{(d)})^{\top}
+\mathbf{\Sigma}^{(d)}
\Bigr)
\Biggr\} 
\\
&\quad
+\frac{1}{2}\sum_{d=1}^{D}\ln|\mathbf{\Sigma}^{(d)}|
+\sum_{d=2}^{D}\sum_{r_d=1}^{R_d}
\left\{
\ln\Gamma(c_N^{r_d})
+c_N^{r_d}
\left(
1-\ln e_N^{r_d}-\frac{e_0^{r_d}}{e_N^{r_d}}
\right)
\right\}
\\
&\quad
+\sum_{d=1}^{D}\sum_{m_d=1}^{M_d}
\left\{
\ln\Gamma(g_N^{m_d})
+g_N^{m_d}
\left(
1-\ln h_N^{m_d}-\frac{h_0^{m_d}}{h_N^{m_d}}
\right)
\right\}
\\
&\quad
+\ln\Gamma(a_N)
+a_N\left(1-\ln b_N-\frac{b_0}{b_N}\right)
+\text{const}
\end{aligned}
\label{eq:43}
\end{equation}
}
\noindent where $\tilde{\bm{\Lambda}}_d := \operatorname{diag}(\tilde{\bm{\lambda}}_d)$ and $\tilde{\bm{\Delta}}_d := \operatorname{diag}(\tilde{\bm{\delta}}_d)$, with $\tilde{\bm{\lambda}}_d := \mathbb{E}_q[\bm{\lambda}_d]$ and $\tilde{\bm{\delta}}_d := \mathbb{E}_q[\bm{\delta}_d]$ the posterior means of the rank and feature precisions, respectively. See Section~9 of the Appendix for a detailed derivation of \eqref{eq:43}.
The posterior expectation of the squared model residual,
$\mathbb{E}_q[\|\mathbf{y}-\mathbf{\Phi}^{\top}\mathbf{w}\|_F^2]$,
can be computed using \eqref{eq:residual_expansion} and
\eqref{eq:khatri_identity}. The lower
bound can be interpreted as follows. The first term captures the model residual.
The second term represents a weighted sum of the squared entries of the
TT cores, incorporating their posterior uncertainty. The remaining terms correspond to the negative KL divergence between the posterior and prior distributions of the hyperparameters.

\subsection{Predictive distribution}
\label{sec:predictive}
The predictive distribution over unseen data points, given the training
data, can be approximated using the variational posterior distribution
as
\begin{equation}
\begin{aligned}
p({y}_i\mid\mathbf{y})
&=
\int
p({y}_i\mid\bm{\Theta})
p(\bm{\Theta}\mid\mathbf{y})\,d\bm{\Theta}
\\
&\simeq
\int\!\int
p\!\left(
{y}_i\mid\{\bm{\mathcal{W}}^{(d)}\}_{d=1}^{D},\tau
\right)
\prod_{d=1}^{D}q(\bm{\mathcal{W}}^{(d)})
q(\tau)\,
d\{\bm{\mathcal{W}}^{(d)}\}_{d=1}^{D}\,d\tau.
\end{aligned}
\label{eq:predictive-distribution}
\end{equation}
\noindent Approximating these integrals yields a Student's t-distribution  $y_i \mid \mathbf{y} \sim \mathcal{T}(\tilde{y_i}, \mathcal{S}_i, \nu_y)$  with its parameters given by 

\begin{equation}
\begin{aligned}
\tilde{y_i}&=\mathbb{E}_q[\mathbf{g}_{<D+1}(\mathbf{x}_i)],\qquad\nu_y=2a_N,\\[2mm]
\mathcal{S}_i&=\left\{\frac{b_N}{a_N}+\sum_{d=1}^{D}\mathbb{E}_q[\mathbf{g}_{\backslash d}(\mathbf{x}_i)]^{\top}\mathbf{\Sigma}^{(d)}\mathbb{E}_q[\mathbf{g}_{\backslash d}(\mathbf{x}_i)]\right\}^{-1}.
\end{aligned}
\label{eq:t-distribution}
\end{equation}
Here, $\mathbb{E}_q[\mathbf{g}_{\backslash d}(\mathbf{x}_i)] = \mathbb{E}_q[\mathbf{g}_{>d}(\mathbf{x}_i)] \otimes \bm{\varphi}^{(d)}(x_i^{(d)}) \otimes \mathbb{E}_q[\mathbf{g}_{<d}(\mathbf{x}_i)] \in \mathbb{R}^{R_d M_d R_{d+1}}$, where $\mathbf{g}_{<d}(\mathbf{x}_i)$ and $\mathbf{g}_{>d}(\mathbf{x}_i)$ are the forward and backward accumulators of Definition~\ref{def:accumulators} evaluated at the test input $\mathbf{x}_i$, with their expectations obtained by replacing each core with its posterior mean. See Section 10 of the Appendix for a detailed derivation. Thus, the predictive variance can be obtained by $\operatorname{Var}(y_i \mid \mathbf{y}) = \frac{\nu_y}{\nu_y - 2} \mathcal{S}_i ^{-1}$, for $\nu_y>2$. 

\subsection{Implementation details}
\label{sec:implementation}

This section describes the practical implementation of the inference procedure in Algorithm~\ref{alg:bttkm}. Most design choices follow the CPD case \citep{kilic_interpretable_2025}, so we restate them only where the TT structure changes them.

\subsubsection{Initialization of hyperparameters}
\label{sec:initialization}

Variational Bayesian inference converges to a local optimum of the lower bound. Appropriate initialization is therefore crucial to avoid poor local solutions. As in the CPD case \citep{kilic_interpretable_2025}, we adopt \textit{uninformative priors} by setting the hyperparameters $\mathbf{c}_0^d$, $\mathbf{e}_0^d$, $\mathbf{g}_0^d$ and $\mathbf{h}_0^d$ to $10^{-6}$ by default. The precisions are initialized at their prior means, $\bm{\Lambda}_d = \mathrm{diag}(\mathbf{c}_0^d/\mathbf{e}_0^d)$ for $d \in [2, D]$ and $\bm{\Delta}_d = \mathrm{diag}(\mathbf{g}_0^d/\mathbf{h}_0^d)$ for $d \in [1, D]$, which gives $\bm{\Lambda}_d = \bm{\Delta}_d = \mathbf{I}$ for the default values. Since $R_1 = R_{D+1} = 1$, the boundary precisions are scalars; they have no hyperprior, are not inferred, and are fixed at $\bm{\Lambda}_1 = \bm{\Lambda}_{D+1} = [1]$.
For the noise precision, the default choice $a_0 = b_0 = 10^{-3}$ initializes $\tau$ at $1$. Note, however, that each TT core is regularized by the product of three precision matrices, $\bm{\Lambda}_{d+1} \otimes \bm{\Delta}_d \otimes \bm{\Lambda}_d$, rather than two as in the CPD case, which strengthens the prior relative to the likelihood. A larger initial noise precision, obtained with a larger $a_0$ for a fixed $b_0 = 10^{-3}$, therefore increases the influence of the likelihood in~\eqref{posterior_updates_cores} and compensates for the stronger regularization. In the UCI experiments, we use $a_0 \in \{10^{-1}, 1\}$, so that $\tau$ starts at $10^{2}$ or $10^{3}$.

To stabilize the first iterations, BTN-KM uses an annealing phase in which the covariances of the factor matrices are introduced gradually, since for large $D$ their product can dominate the first updates and cause numerical instability~\citep{kilic_interpretable_2025}. BTT-KM does not need this phase. Instead, only the cores are updated during the first five iterations (warm-up), so $\tau$, $\bm{\lambda}_d$ and $\bm{\delta}_d$ stay at their initial values and no ranks are truncated. The cores $\bm{\mathcal{W}}^{(d)}$ are initialized from $\mathcal{N}(0, \mathbf{I})$, and the covariance matrices $\bm{\Sigma}^{(d)}$ are set to $\sigma^2 \mathbf{I}$ with $\sigma^2 = 10^{-5}$. The initial ranks $\{R_d\}_{d=2}^{D}$ and feature dimensions $\{M_d\}_{d=1}^{D}$ are chosen manually, sufficiently large to provide enough model capacity and within the available computational resources. The hyperparameter values used in our experiments are given in Sections~\ref{sec:synthetic} and~\ref{sec:uci}.

\input{algorithm}
\subsubsection{Efficient variational updates and computational complexity}
\label{sec:efficient_updates}

The inference sweeps over the TT cores. In each iteration, we update for $d = 1,\ldots,D$ the posterior of the core $q(\operatorname{vec}(\bm{\mathcal{W}}^{(d)}))$ using \eqref{posterior_updates_cores}, then $q(\bm{\lambda}_d)$ using \eqref{posterior_updates_lambda} (for $d \geq 2$) and $q(\bm{\delta}_d)$ using \eqref{posterior_updates_delta}. After the sweep, we update $q(\tau)$ using \eqref{eq:20}, optionally truncate redundant ranks, and evaluate the lower bound using \eqref{eq:43}.

Each core update needs the expectations $\mathbb{E}_q[\mathbf{G}_{\backslash d}]$ and $\mathbb{E}_q[\mathbf{G}_{\backslash d}\mathbf{G}_{\backslash d}^{\top}]$ of the contraction of all other cores. Recomputing them from scratch would add a factor $D$ to the cost, so we propagate them recursively. The backward quantities $\mathbf{G}_{>d}$ and $\mathbf{H}_{>d}$ are computed once per iteration, and the forward quantities $\mathbf{G}_{<d+1}$ and $\mathbf{H}_{<d+1}$ are updated right after core $d$ is updated, so each core needs only one extra contraction. The mean $\mathbb{E}_q[\mathbf{G}_{\backslash d}]$ uses the posterior means of the cores, whereas the Gram matrices also include the posterior covariances through \eqref{eq:expected_kronecker}. The updates are sequential as in ALS: cores $k<d$ use their values from the current iteration and cores $k>d$ from the previous one.

When rank truncation is enabled, it is applied at every iteration after the initial five. For each rank $d \in [2,D]$, we define the variance contribution of rank component $r_d$ as $\|\tilde{\mathbf{W}}^{(d-1)}_{::r_d}\|_F^2 + \|\tilde{\mathbf{W}}^{(d)}_{r_d::}\|_F^2$, the squared norm of its associated frontal and horizontal slices, normalized over all components of $R_d$. The component $r_d$ is removed if its variance contribution is below a threshold $\epsilon$, set to $10^{-3}\,\%$ by default, while at least the largest component is always retained. This removes the corresponding slices of both neighboring cores, the matching rows and columns of their covariances, and the associated $\lambda_{r_d}$ together with its posterior parameters. 
Convergence is defined as the point at which the relative change in the lower bound between consecutive iterations falls below a user-defined threshold of $10^{-4}$; otherwise, training is stopped when the maximum number of iterations is reached. Since truncation changes the ranks, the lower bound is not guaranteed to be non-decreasing at iterations where components are removed.

The cost per iteration is dominated by the core updates. With $R=\max_d R_d$ and $M=\max_d M_d$, forming $\mathbb{E}_q[\mathbf{G}_{\backslash d}\mathbf{G}_{\backslash d}^{\top}]$ for each core costs $\mathcal{O}(N(MR^2)^2)$ and inverting the $R_dM_dR_{d+1}$-dimensional precision matrix costs $\mathcal{O}((MR^2)^3)$. The total is therefore $\mathcal{O}(ND(MR^2)^2 + D(MR^2)^3)$. The hyperparameter updates are of lower order. This matches the cost of deterministic TT-ALS \citep{holtz_alternating_2012, batselier_tensor_2017}, so uncertainty quantification comes at no extra asymptotic computational cost. The main additional requirement is storing the covariances $\bm{\Sigma}^{(d)}$, which takes $\mathcal{O}(D(MR^2)^2)$ memory.

%% file: TT_model.tex
\begin{figure}[t]
    \centering
    \resizebox{0.55\textwidth}{!}{ 
        \begin{tikzpicture}[x=0.75pt,y=0.75pt,yscale=-1,xscale=1]

\draw   (64.67,240.57) .. controls (64.67,222.07) and (79.67,207.07) .. (98.17,207.07) .. controls (116.67,207.07) and (131.67,222.07) .. (131.67,240.57) .. controls (131.67,259.08) and (116.67,274.07) .. (98.17,274.07) .. controls (79.67,274.07) and (64.67,259.08) .. (64.67,240.57) -- cycle ;
\draw   (299.67,239.57) .. controls (299.67,221.07) and (314.67,206.07) .. (333.17,206.07) .. controls (351.67,206.07) and (366.67,221.07) .. (366.67,239.57) .. controls (366.67,258.08) and (351.67,273.07) .. (333.17,273.07) .. controls (314.67,273.07) and (299.67,258.08) .. (299.67,239.57) -- cycle ;
\draw   (537.33,242.24) .. controls (537.33,223.74) and (552.33,208.74) .. (570.83,208.74) .. controls (589.33,208.74) and (604.33,223.74) .. (604.33,242.24) .. controls (604.33,260.74) and (589.33,275.74) .. (570.83,275.74) .. controls (552.33,275.74) and (537.33,260.74) .. (537.33,242.24) -- cycle ;
\draw [dashed]  (314.99,341.18) .. controls (314.94,331.23) and (322.96,323.13) .. (332.9,323.09) .. controls (342.84,323.04) and (350.94,331.06) .. (350.99,341) .. controls (351.04,350.94) and (343.02,359.04) .. (333.08,359.09) .. controls (323.14,359.14) and (315.04,351.12) .. (314.99,341.18) -- cycle ;
\draw   (312.17,95.57) .. controls (312.17,84.8) and (320.9,76.07) .. (331.67,76.07) .. controls (342.44,76.07) and (351.17,84.8) .. (351.17,95.57) .. controls (351.17,106.34) and (342.44,115.07) .. (331.67,115.07) .. controls (320.9,115.07) and (312.17,106.34) .. (312.17,95.57) -- cycle ;
\draw   (75,96.24) .. controls (75,85.47) and (83.73,76.74) .. (94.5,76.74) .. controls (105.27,76.74) and (114,85.47) .. (114,96.24) .. controls (114,107.01) and (105.27,115.74) .. (94.5,115.74) .. controls (83.73,115.74) and (75,107.01) .. (75,96.24) -- cycle ;
\draw   (550.33,95.91) .. controls (550.33,85.14) and (559.06,76.41) .. (569.83,76.41) .. controls (580.6,76.41) and (589.33,85.14) .. (589.33,95.91) .. controls (589.33,106.68) and (580.6,115.41) .. (569.83,115.41) .. controls (559.06,115.41) and (550.33,106.68) .. (550.33,95.91) -- cycle ;
\draw   (544.99,341.18) .. controls (544.94,331.23) and (552.96,323.13) .. (562.9,323.09) .. controls (572.84,323.04) and (580.94,331.06) .. (580.99,341) .. controls (581.04,350.94) and (573.02,359.04) .. (563.08,359.09) .. controls (553.14,359.14) and (545.04,351.12) .. (544.99,341.18) -- cycle ;
\draw    (119.67,266.07) -- (304.13,336.29) ;
\draw [shift={(306,337)}, rotate = 200.84] [color={rgb, 255:red, 0; green, 0; blue, 0 }  ][line width=0.75]    (10.93,-3.29) .. controls (6.95,-1.4) and (3.31,-0.3) .. (0,0) .. controls (3.31,0.3) and (6.95,1.4) .. (10.93,3.29)   ;
\draw    (543,261) -- (358.85,336.24) ;
\draw [shift={(357,337)}, rotate = 337.77] [color={rgb, 255:red, 0; green, 0; blue, 0 }  ][line width=0.75]    (10.93,-3.29) .. controls (6.95,-1.4) and (3.31,-0.3) .. (0,0) .. controls (3.31,0.3) and (6.95,1.4) .. (10.93,3.29)   ;
\draw    (544.99,341.18) -- (363.99,342.16) ;
\draw [shift={(361.99,342.18)}, rotate = 359.69] [color={rgb, 255:red, 0; green, 0; blue, 0 }  ][line width=0.75]    (10.93,-3.29) .. controls (6.95,-1.4) and (3.31,-0.3) .. (0,0) .. controls (3.31,0.3) and (6.95,1.4) .. (10.93,3.29)   ;
\draw    (333.17,273.07) -- (332.69,313.07) ;
\draw [shift={(332.67,315.07)}, rotate = 270.68] [color={rgb, 255:red, 0; green, 0; blue, 0 }  ][line width=0.75]    (10.93,-3.29) .. controls (6.95,-1.4) and (3.31,-0.3) .. (0,0) .. controls (3.31,0.3) and (6.95,1.4) .. (10.93,3.29)   ;
\draw    (312.5,62.41) -- (319.39,72.74) ;
\draw [shift={(320.5,74.41)}, rotate = 236.31] [color={rgb, 255:red, 0; green, 0; blue, 0 }  ][line width=0.75]    (10.93,-3.29) .. controls (6.95,-1.4) and (3.31,-0.3) .. (0,0) .. controls (3.31,0.3) and (6.95,1.4) .. (10.93,3.29)   ;
\draw    (349.83,61.41) -- (342.88,72.7) ;
\draw [shift={(341.83,74.41)}, rotate = 301.61] [color={rgb, 255:red, 0; green, 0; blue, 0 }  ][line width=0.75]    (10.93,-3.29) .. controls (6.95,-1.4) and (3.31,-0.3) .. (0,0) .. controls (3.31,0.3) and (6.95,1.4) .. (10.93,3.29)   ;
\draw    (331.67,115.07) -- (330.04,192.57) ;
\draw [shift={(330,194.57)}, rotate = 271.2] [color={rgb, 255:red, 0; green, 0; blue, 0 }  ][line width=0.75]    (10.93,-3.29) .. controls (6.95,-1.4) and (3.31,-0.3) .. (0,0) .. controls (3.31,0.3) and (6.95,1.4) .. (10.93,3.29)   ;
\draw    (542,306.07) -- (548.89,316.41) ;
\draw [shift={(550,318.07)}, rotate = 236.31] [color={rgb, 255:red, 0; green, 0; blue, 0 }  ][line width=0.75]    (10.93,-3.29) .. controls (6.95,-1.4) and (3.31,-0.3) .. (0,0) .. controls (3.31,0.3) and (6.95,1.4) .. (10.93,3.29)   ;
\draw    (582,304.07) -- (575.95,315.31) ;
\draw [shift={(575,317.07)}, rotate = 298.3] [color={rgb, 255:red, 0; green, 0; blue, 0 }  ][line width=0.75]    (10.93,-3.29) .. controls (6.95,-1.4) and (3.31,-0.3) .. (0,0) .. controls (3.31,0.3) and (6.95,1.4) .. (10.93,3.29)   ;
\draw    (94.5,115.74) -- (93.04,195.57) ;
\draw [shift={(93,197.57)}, rotate = 271.05] [color={rgb, 255:red, 0; green, 0; blue, 0 }  ][line width=0.75]    (10.93,-3.29) .. controls (6.95,-1.4) and (3.31,-0.3) .. (0,0) .. controls (3.31,0.3) and (6.95,1.4) .. (10.93,3.29)   ;
\draw    (569.83,115.41) -- (569.02,195.57) ;
\draw [shift={(569,197.57)}, rotate = 270.58] [color={rgb, 255:red, 0; green, 0; blue, 0 }  ][line width=0.75]    (10.93,-3.29) .. controls (6.95,-1.4) and (3.31,-0.3) .. (0,0) .. controls (3.31,0.3) and (6.95,1.4) .. (10.93,3.29)   ;
\draw    (257.64,147.73) -- (263.07,155.81) ;
\draw [shift={(264.18,157.47)}, rotate = 236.08] [color={rgb, 255:red, 0; green, 0; blue, 0 }  ][line width=0.75]    (10.93,-3.29) .. controls (6.95,-1.4) and (3.31,-0.3) .. (0,0) .. controls (3.31,0.3) and (6.95,1.4) .. (10.93,3.29)   ;
\draw    (293.36,146.11) -- (288.59,154.9) ;
\draw [shift={(287.64,156.66)}, rotate = 298.5] [color={rgb, 255:red, 0; green, 0; blue, 0 }  ][line width=0.75]    (10.93,-3.29) .. controls (6.95,-1.4) and (3.31,-0.3) .. (0,0) .. controls (3.31,0.3) and (6.95,1.4) .. (10.93,3.29)   ;
\draw   (253,182.91) .. controls (253,170.16) and (263.67,159.82) .. (276.83,159.82) .. controls (290,159.82) and (300.67,170.16) .. (300.67,182.91) .. controls (300.67,195.66) and (290,206) .. (276.83,206) .. controls (263.67,206) and (253,195.66) .. (253,182.91) -- cycle ;
\draw    (370.67,199.74) -- (363.84,209.12) ;
\draw [shift={(362.67,210.74)}, rotate = 306.03] [color={rgb, 255:red, 0; green, 0; blue, 0 }  ][line width=0.75]    (10.93,-3.29) .. controls (6.95,-1.4) and (3.31,-0.3) .. (0,0) .. controls (3.31,0.3) and (6.95,1.4) .. (10.93,3.29)   ;
\draw    (292.33,200.41) -- (300.4,210.2) ;
\draw [shift={(301.67,211.74)}, rotate = 230.53] [color={rgb, 255:red, 0; green, 0; blue, 0 }  ][line width=0.75]    (10.93,-3.29) .. controls (6.95,-1.4) and (3.31,-0.3) .. (0,0) .. controls (3.31,0.3) and (6.95,1.4) .. (10.93,3.29)   ;
\draw    (75.83,62.07) -- (82.72,72.41) ;
\draw [shift={(83.83,74.07)}, rotate = 236.31] [color={rgb, 255:red, 0; green, 0; blue, 0 }  ][line width=0.75]    (10.93,-3.29) .. controls (6.95,-1.4) and (3.31,-0.3) .. (0,0) .. controls (3.31,0.3) and (6.95,1.4) .. (10.93,3.29)   ;
\draw    (113.17,61.07) -- (106.21,72.37) ;
\draw [shift={(105.17,74.07)}, rotate = 301.61] [color={rgb, 255:red, 0; green, 0; blue, 0 }  ][line width=0.75]    (10.93,-3.29) .. controls (6.95,-1.4) and (3.31,-0.3) .. (0,0) .. controls (3.31,0.3) and (6.95,1.4) .. (10.93,3.29)   ;
\draw    (552.17,63.07) -- (559.06,73.41) ;
\draw [shift={(560.17,75.07)}, rotate = 236.31] [color={rgb, 255:red, 0; green, 0; blue, 0 }  ][line width=0.75]    (10.93,-3.29) .. controls (6.95,-1.4) and (3.31,-0.3) .. (0,0) .. controls (3.31,0.3) and (6.95,1.4) .. (10.93,3.29)   ;
\draw    (589.5,62.07) -- (582.55,73.37) ;
\draw [shift={(581.5,75.07)}, rotate = 301.61] [color={rgb, 255:red, 0; green, 0; blue, 0 }  ][line width=0.75]    (10.93,-3.29) .. controls (6.95,-1.4) and (3.31,-0.3) .. (0,0) .. controls (3.31,0.3) and (6.95,1.4) .. (10.93,3.29)   ;
\draw    (366.64,146.73) -- (372.07,154.81) ;
\draw [shift={(373.18,156.47)}, rotate = 236.08] [color={rgb, 255:red, 0; green, 0; blue, 0 }  ][line width=0.75]    (10.93,-3.29) .. controls (6.95,-1.4) and (3.31,-0.3) .. (0,0) .. controls (3.31,0.3) and (6.95,1.4) .. (10.93,3.29)   ;
\draw    (402.36,145.11) -- (397.59,153.9) ;
\draw [shift={(396.64,155.66)}, rotate = 298.5] [color={rgb, 255:red, 0; green, 0; blue, 0 }  ][line width=0.75]    (10.93,-3.29) .. controls (6.95,-1.4) and (3.31,-0.3) .. (0,0) .. controls (3.31,0.3) and (6.95,1.4) .. (10.93,3.29)   ;
\draw   (362,181.91) .. controls (362,169.16) and (372.67,158.82) .. (385.83,158.82) .. controls (399,158.82) and (409.67,169.16) .. (409.67,181.91) .. controls (409.67,194.66) and (399,205) .. (385.83,205) .. controls (372.67,205) and (362,194.66) .. (362,181.91) -- cycle ;
\draw    (134.67,200.74) -- (127.84,210.12) ;
\draw [shift={(126.67,211.74)}, rotate = 306.03] [color={rgb, 255:red, 0; green, 0; blue, 0 }  ][line width=0.75]    (10.93,-3.29) .. controls (6.95,-1.4) and (3.31,-0.3) .. (0,0) .. controls (3.31,0.3) and (6.95,1.4) .. (10.93,3.29)   ;
\draw    (130.64,147.73) -- (136.07,155.81) ;
\draw [shift={(137.18,157.47)}, rotate = 236.08] [color={rgb, 255:red, 0; green, 0; blue, 0 }  ][line width=0.75]    (10.93,-3.29) .. controls (6.95,-1.4) and (3.31,-0.3) .. (0,0) .. controls (3.31,0.3) and (6.95,1.4) .. (10.93,3.29)   ;
\draw    (166.36,146.11) -- (161.59,154.9) ;
\draw [shift={(160.64,156.66)}, rotate = 298.5] [color={rgb, 255:red, 0; green, 0; blue, 0 }  ][line width=0.75]    (10.93,-3.29) .. controls (6.95,-1.4) and (3.31,-0.3) .. (0,0) .. controls (3.31,0.3) and (6.95,1.4) .. (10.93,3.29)   ;
\draw   (126,182.91) .. controls (126,170.16) and (136.67,159.82) .. (149.83,159.82) .. controls (163,159.82) and (173.67,170.16) .. (173.67,182.91) .. controls (173.67,195.66) and (163,206) .. (149.83,206) .. controls (136.67,206) and (126,195.66) .. (126,182.91) -- cycle ;
\draw    (495.64,146.73) -- (501.07,154.81) ;
\draw [shift={(502.18,156.47)}, rotate = 236.08] [color={rgb, 255:red, 0; green, 0; blue, 0 }  ][line width=0.75]    (10.93,-3.29) .. controls (6.95,-1.4) and (3.31,-0.3) .. (0,0) .. controls (3.31,0.3) and (6.95,1.4) .. (10.93,3.29)   ;
\draw    (531.36,145.11) -- (526.59,153.9) ;
\draw [shift={(525.64,155.66)}, rotate = 298.5] [color={rgb, 255:red, 0; green, 0; blue, 0 }  ][line width=0.75]    (10.93,-3.29) .. controls (6.95,-1.4) and (3.31,-0.3) .. (0,0) .. controls (3.31,0.3) and (6.95,1.4) .. (10.93,3.29)   ;
\draw   (491,181.91) .. controls (491,169.16) and (501.67,158.82) .. (514.83,158.82) .. controls (528,158.82) and (538.67,169.16) .. (538.67,181.91) .. controls (538.67,194.66) and (528,205) .. (514.83,205) .. controls (501.67,205) and (491,194.66) .. (491,181.91) -- cycle ;
\draw    (530.33,199.41) -- (538.4,209.2) ;
\draw [shift={(539.67,210.74)}, rotate = 230.53] [color={rgb, 255:red, 0; green, 0; blue, 0 }  ][line width=0.75]    (10.93,-3.29) .. controls (6.95,-1.4) and (3.31,-0.3) .. (0,0) .. controls (3.31,0.3) and (6.95,1.4) .. (10.93,3.29)   ;

\draw (290.17,40.07) node [anchor=north west][inner sep=0.75pt]   [align=left] {\Large $\bm{g^{d}}$};
\draw (340.83,40.41) node [anchor=north west][inner sep=0.75pt]   [align=left] {\Large $\bm{h^{d}}$};
\draw (80.67,85.81) node [anchor=north west][inner sep=0.75pt]     {\Large $\bm{\delta}_{1}$};
\draw (317.17,85.47) node [anchor=north west][inner sep=0.75pt]     {\Large $\bm{\delta}_{d}$};
\draw (554.33,84.81) node [anchor=north west][inner sep=0.75pt]     {\Large $\bm{\delta}_{D}$};
\draw (555,335.47) node [anchor=north west][inner sep=0.75pt]    {\Large$\tau$};
\draw (328,335.47) node [anchor=north west][inner sep=0.75pt]    {\Large$\bm{y}$};
\draw (310.67,230.48) node [anchor=north west][inner sep=0.75pt]  [rotate=-359.99] {\Large$\bm{\mathcal{W}}^{(d)}$};
\draw (80.67,230.48) node [anchor=north west][inner sep=0.75pt]  [rotate=-359.99]  {\Large$\bm{\mathcal{W}}^{(1)}$};
\draw (550.33,230.48) node [anchor=north west][inner sep=0.75pt]  [rotate=-359.99]  {\Large$\bm{\mathcal{W}}^{(D)}$};
\draw (530,288.07) node [anchor=north west][inner sep=0.75pt]   [align=left] {\Large$a$};
\draw (583,285.07) node [anchor=north west][inner sep=0.75pt]   [align=left] {\Large$b$};
\draw (237.32,128.09) node [anchor=north west][inner sep=0.75pt]   [align=left] {\Large $\bm{c}^{d}$};
\draw (286.18,125.09) node [anchor=north west][inner sep=0.75pt]   [align=left] {\Large $\bm{e}^{d}$};
\draw (260.27,171.34) node [anchor=north west][inner sep=0.75pt]    {\Large $\bm{\lambda} _{{d}}$};
\draw (53.5,39.74) node [anchor=north west][inner sep=0.75pt]   [align=left] {\Large $\bm{g^{1}}$};
\draw (104.17,40.07) node [anchor=north west][inner sep=0.75pt]   [align=left] {\Large $\bm{h^{1}}$};
\draw (529.83,40.74) node [anchor=north west][inner sep=0.75pt]   [align=left] {\Large $\bm{g^{D}}$};
\draw (580.5,41.07) node [anchor=north west][inner sep=0.75pt]   [align=left] {\Large $\bm{h^{D}}$};
\draw (349.32,127.09) node [anchor=north west][inner sep=0.75pt]   [align=left] {\Large $\bm{c}^{d+1}$};
\draw (395.18,124.09) node [anchor=north west][inner sep=0.75pt]   [align=left] {\Large $\bm{e}^{d+1}$};
\draw (370.27,170.34) node [anchor=north west][inner sep=0.75pt]    {\Large $\bm{\lambda}_{{d+1}}$};
\draw (110.32,128.09) node [anchor=north west][inner sep=0.75pt]   [align=left] {\Large $\bm{c}^2$};
\draw (159.18,125.09) node [anchor=north west][inner sep=0.75pt]   [align=left] {\Large $\bm{e}^2$};
\draw (134.27,171.34) node [anchor=north west][inner sep=0.75pt]    {\Large $\bm{\lambda} _{2}$};
\draw (475.32,127.09) node [anchor=north west][inner sep=0.75pt]   [align=left] {\Large $\bm{c}^{D}$};
\draw (524.18,124.09) node [anchor=north west][inner sep=0.75pt]   [align=left] {\Large $\bm{e}^{D}$};
\draw (496.27,170.34) node [anchor=north west][inner sep=0.75pt]    {\Large $\bm{\lambda}_{D}$};

\node at (215,240) {\Huge $\cdots$};  
\node at (460,240) {\Huge $\cdots$};  

        \end{tikzpicture}
    }
    \caption{\footnotesize Representation of BTT-KM with the TT-decomposed weight vector $\mathbf{w}$ as a probabilistic graphical model showing the ARD-type hierarchical  priors over the TT cores $\{\bm{\mathcal{W}}^{(d)}\}_{d=1}^D$ by the rank precisions $\{\boldsymbol{\lambda}_{d}\}_{d=2}^{D}$ and the feature precisions $\{\boldsymbol{\delta}_{d}\}_{d=1}^D$. The dashed node denotes the observed data $\mathbf{y}$, while the solid nodes represent random variables. Shape and rate hyperparameters of the Gamma priors placed on $\{\boldsymbol{\lambda}_{d}\}_{d=2}^{D}$, $\{\boldsymbol{\delta}_{d}\}_{d=1}^D$ and $\tau$ are shown as unbounded nodes.}
    \label{fig:TT_model}
\end{figure}
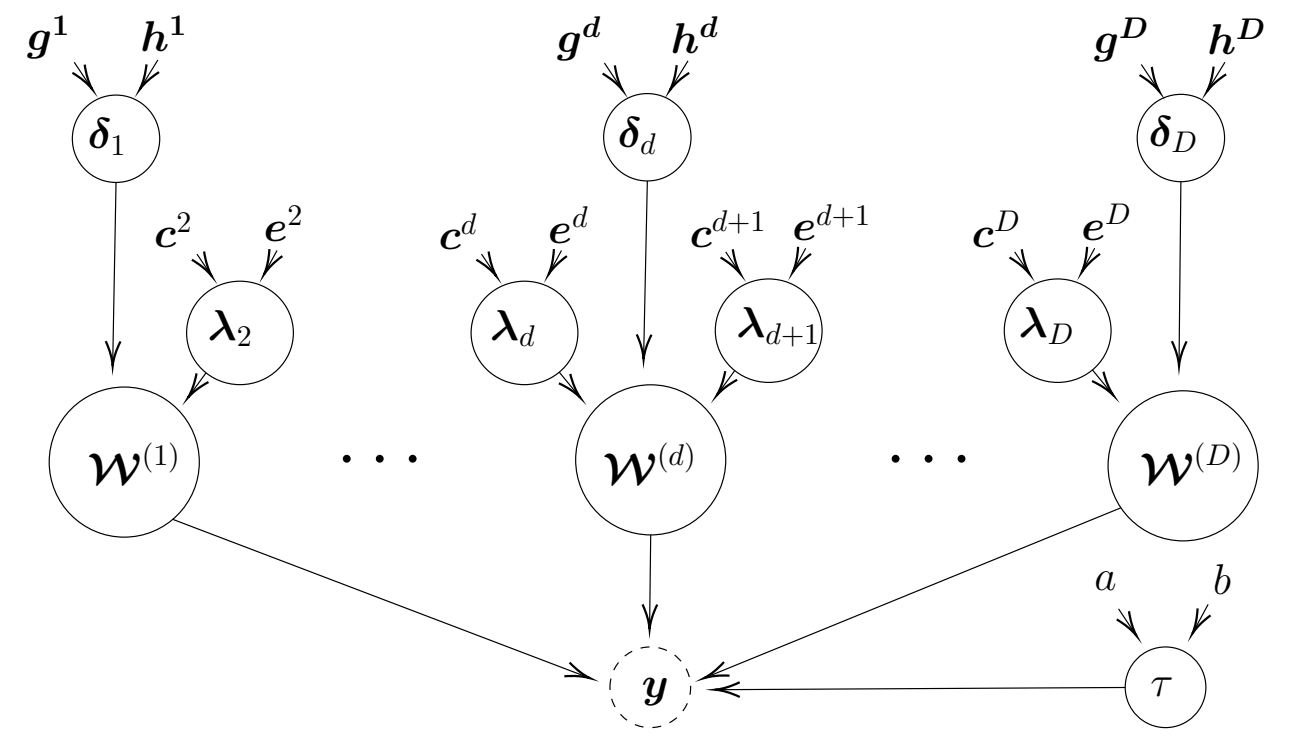

%% file: algorithm.tex
\begin{algorithm}[t]
\footnotesize
\caption{Learning Bayesian TT Kernel Machines}
\label{alg:bttkm}
\begin{algorithmic}[1]

\REQUIRE Inputs $\{\mathbf{x}_n\}_{n=1}^N$ and outputs $\mathbf{y} = \{y_n\}_{n=1}^N$

\STATE \textbf{Initialization:} choose TT ranks $\{R_d\}_{d=2}^D$ and feature dimensions $\{M_d\}_{d=1}^D$
\STATE Set $R_1 = R_{D+1} = 1$ and $\bm{\Lambda}_1 = \bm{\Lambda}_{D+1} = [1]$
\STATE Initialize $\bm{\mathcal{W}}^{(d)}$ and $\mathbf{\Sigma}^{(d)}$ for all $d \in \{1,\dots,D\}$
\STATE Initialize hyperparameters $a_0$, $b_0$, $\{\mathbf{c}_0^{d}, \mathbf{e}_0^{d}\}_{d=2}^D$, $\{\mathbf{g}_0^{d}, \mathbf{h}_0^{d}\}_{d=1}^D$
\STATE Set $\tau = a_0 / b_0$
\STATE Set $\lambda_{r_d} = c_0^{r_d} / e_0^{r_d}$ for all $r_d \in \{1,\dots,R_d\}$ and $d \in \{2,\dots,D\}$
\STATE Set $\delta_{m_d} = g_0^{m_d} / h_0^{m_d}$ for all $m_d \in \{1,\dots,M_d\}$ and $d \in \{1,\dots,D\}$
\STATE Set iteration counter $t = 0$

\REPEAT
    \STATE $t \leftarrow t + 1$
    \FOR{$d = 1$ \TO $D$}
        \STATE Update the posterior $q(\operatorname{vec}(\bm{\mathcal{W}}^{(d)}))$ using Eq.~\eqref{posterior_updates_cores}
        \IF{$t > 5$}
            \IF{$d \geq 2$}
                \STATE Update the posterior $q(\bm{\lambda}_d)$ using Eq.~\eqref{posterior_updates_lambda}
            \ENDIF
            \STATE Update the posterior $q(\bm{\delta}_d)$ using Eq.~\eqref{posterior_updates_delta}
        \ENDIF
    \ENDFOR

    \IF{$t > 5$}
        \STATE Update the posterior $q(\tau)$ using Eq.~\eqref{eq:20}
        \IF{rank truncation is enabled}
            \STATE For all $d \in \{2,\dots,D\}$, remove the rank components $r_d$ whose variance contribution is below the threshold $\epsilon$, i.e., the corresponding frontal slices of $\bm{\mathcal{W}}^{(d-1)}$ and horizontal slices of $\bm{\mathcal{W}}^{(d)}$
        \ENDIF
    \ENDIF

    \STATE Evaluate the lower bound using Eq.~\eqref{eq:43}

\UNTIL{convergence}

\STATE Compute the predictive distribution using Eq.~\eqref{eq:t-distribution}

\end{algorithmic}
\end{algorithm}
\vspace{-0.3cm}

%% file: sections/4.experiments.tex
\section{Experiments}
\label{sec:experiments}
In this section, we evaluate BTT-KM in two steps. We first use synthetic data with a known ground truth to show how the rank precisions $\bm{\lambda}_d$ and the feature precisions $\bm{\delta}_d$ regularize the slices of the TT cores, and to check whether the true TT ranks and feature dimensions are recovered from an over-parameterized initialization. We then compare BTT-KM with its CPD-based counterpart BTN-KM \citep{kilic_interpretable_2025} and a Gaussian process (GP) reference on six UCI data sets \citep{dua_uci_2019}, in terms of accuracy, uncertainty quantification, inferred model complexity and run time. Finally, we validate the analytic predictive distribution~\eqref{eq:t-distribution} against a Monte Carlo (MC) sampling of the same variational posteriors. The Python code used in all experiments is available at \url{https://github.com/afrakilic/BTTKM}

\subsection{Ground Truth Recovery with Synthetic Data}
\label{sec:synthetic}

\begin{figure}[t]
\centering
\includegraphics[width=0.8\linewidth]{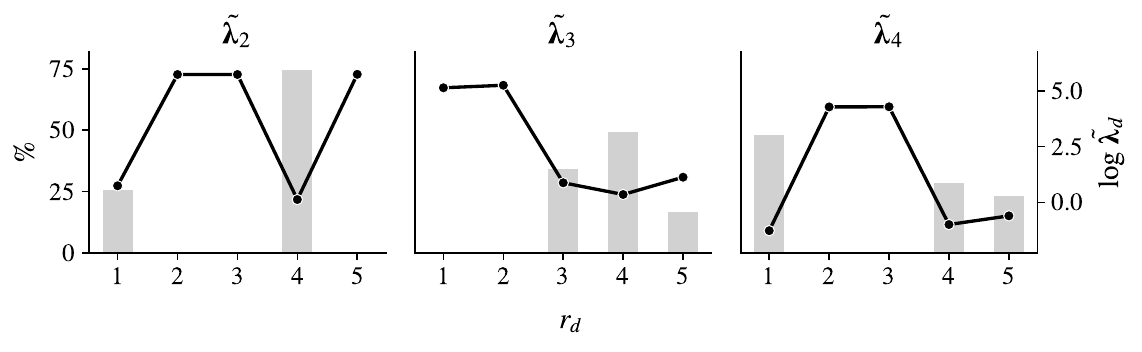}
\caption{Variance contribution (\%) per rank component of each TT rank $R_d$, together with the posterior mean log-precisions $\log \tilde{\bm{\lambda}}_d$.}
\label{fig:synthetic_lambda}
\end{figure}

\begin{figure}[t]
\centering
\includegraphics[width=\linewidth]{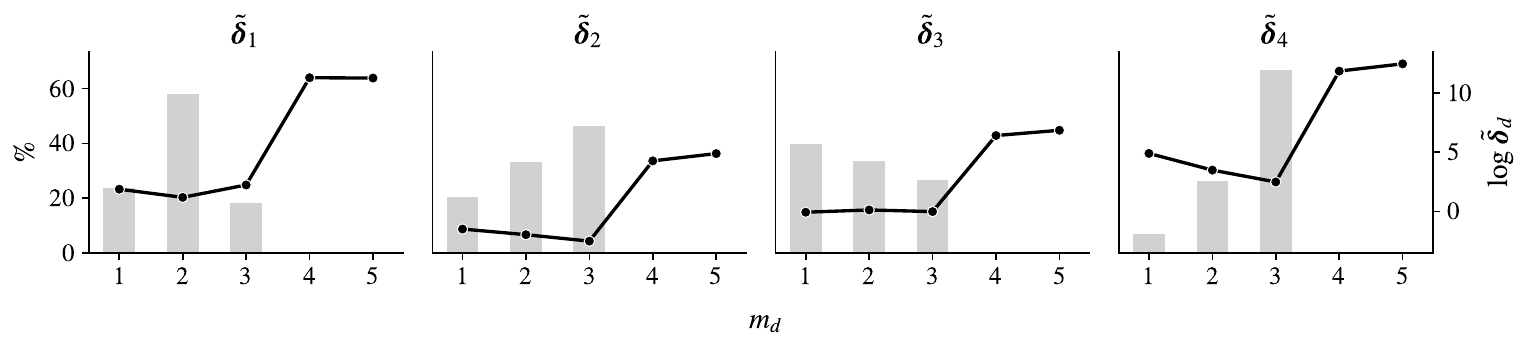}
\caption{Variance contribution (\%) per feature component of each core, together with the posterior mean log-precisions $\log \tilde{\bm{\delta}}_d$.}
\label{fig:synthetic_delta}
\end{figure}

To illustrate how the rank precisions $\bm{\lambda}_d$ and the feature precisions $\bm{\delta}_d$ regularize the TT cores, we consider a synthetic experiment constructed analogously to the CPD case~\citep{kilic_interpretable_2025}. The entries of the input matrix $\mathbf{X} \in \mathbb{R}^{N \times D}$ are sampled independently from a standard normal distribution, with $N = 100$ and $D = 4$, so that the tensor train has three TT ranks $R_2$, $R_3$ and $R_4$. The inputs are mapped to the same polynomial features as in~\citet{kilic_interpretable_2025}, with $M_d = 5$ for every core. The ground-truth TT cores have ranks $R_d = 3$, are drawn from $\mathcal{N}(0, \mathbf{I})$, and have nonzero entries only for the first three feature components of every core, so that the true feature dimension is $3$. The outputs are generated as $\mathbf{y} = \bm{\Phi}^{\top}\mathbf{w} + \mathbf{e}$, with $\mathbf{e} \sim \mathcal{N}(\mathbf{0}, \sigma_e^2 \mathbf{I})$ and $\sigma_e = 10^{-3}$.
The model is initialized with $R_d = M_d = 5$ for every core, which is larger than the true values, and with hyperparameters $a_0 = b_0 = 10^{-3}$ and $c_0 = e_0 = g_0 = h_0 = 10^{-6}$. Rank truncation is disabled, so that the shrinkage of all components remains visible, and the model is trained for the full $100$ iterations without the convergence criterion of Section~\ref{sec:efficient_updates}. As in the CPD case~\citep{kilic_interpretable_2025}, the lower bound increases monotonically after the warm-up, quickly at first and slowly afterwards. We report the variance contribution by each rank component, as defined in Section~\ref{sec:efficient_updates}, and similarly the variance contribution by feature component $m_d$ of core $d$ as $\|\tilde{\mathbf{W}}^{(d)}_{:m_d:}\|_F^2$, normalized over all feature components of that core.

Figure~\ref{fig:synthetic_lambda} shows the variance contribution per rank component together with $\log \tilde{\bm{\lambda}}_d$, the posterior mean rank precisions defined below~\eqref{eq:43}. For $R_3$ and $R_4$, two of the five components shrink to zero, recovering the true rank of $3$. For $R_2$, three components shrink to zero, so the inferred rank is $2$, one below the true rank. The inactive components contribute at most $5.7 \times 10^{-8}\%$ of the variance, whereas the smallest active component contributes $16.5\%$, so the inferred ranks are insensitive to the truncation threshold over a wide range. The inactive components also receive the largest precisions of their rank; for $R_3$, for example, they have rank precisions of approximately $170$ and $190$, compared with $1.4$ to $3.1$ for the active ones. Since the overall magnitude of $\tilde{\bm{\lambda}}_d$ differs between ranks, we compare rank precisions only within each rank. Finally, the active components differ between ranks, e.g., components $3$, $4$ and $5$ for $R_3$ and components $1$, $4$ and $5$ for $R_4$. This is expected, since both the TT representation and the prior in~\eqref{lamba_prior} are invariant to permutations of the rank index, so the prior determines how many components are relevant but not which ones.

Figure~\ref{fig:synthetic_delta} shows the corresponding results for the feature components, together with $\log \tilde{\bm{\delta}}_d$, the posterior mean feature precisions defined below~\eqref{eq:43}. Unlike the rank components, each feature component corresponds to a fixed basis function, so the relevant ones are identifiable. In every core, the three true components carry the variance, with the smallest of them still contributing $7.0\%$, while each of the two irrelevant components contributes at most $0.0045\%$. Likewise, the irrelevant components receive the largest entries of $\tilde{\bm{\delta}}_d$, two to four orders of magnitude above those of the true components of the same core. For the fourth core, for instance, the irrelevant components have feature precisions above $10^{5}$, compared with values between $12$ and $133$ for the true ones. As for the ranks, we compare feature precisions only within each core.
\input{predictive_performance}

\subsection{Comparison on UCI Benchmarks}
\label{sec:uci}

We compare BTT-KM with BTN-KM~\citep{kilic_interpretable_2025} and a Gaussian process (GP) reference on the six UCI data sets~\citep{dua_uci_2019} of Table~\ref{tab:acc}. To attribute any difference to the TN format, we follow the protocol of~\citet{kilic_interpretable_2025} and use the same feature map, feature dimensions $M$, train-test splits, normalization, metrics and Gamma hyperparameters, except for $a_0$. All BTN-KM and GP results are taken from~\citet{kilic_interpretable_2025}, apart from the BTN-KM run times, which we measured on the same laptop as BTT-KM. There are three differences. First, we initialize all TT ranks at $R_{\max}=5$, except for Adult, for which $R_{\max}=3$, so that an inner TT core ($R_{\max}^2M$ entries) is as close as possible in size to a BTN-KM factor matrix of rank $25$ and $10$, respectively. Second, each TT core is regularized by the product of three precision matrices rather than two, which shrinks its entries more strongly. We therefore choose $a_0$, and thus the initial noise precision, 10-100 times larger than in BTN-KM. Third, BTT-KM is trained without the annealing used for BTN-KM, and only its precisions are kept fixed during the first five iterations. We report the effective rank $R_{\mathrm{eff}}$ after truncation, which is the CP rank for BTN-KM and the mean inner TT rank for BTT-KM.

We evaluate the uncertainty by the test negative log-likelihood (NLL) and, for regression, by the coverage of the nominal 95\% predictive interval (PICP-95). We summarize the coverage by the coverage error $\dpp$, the mean absolute deviation of PICP-95 from 95\%. The closed-form Student's t distribution~\eqref{eq:t-distribution} is an approximation of the predictive integral~\eqref{eq:predictive-distribution}. To assess this approximation separately from the variational posterior itself, we compare it with a Monte Carlo (MC) estimate of the same integral. For this, we draw 3000 samples of the TT cores and $\tau$ from $q(\Theta)$ and sample $y_i$ for each of them. The GP serves as a reference for the calibration of the predictive uncertainty. Its hyperparameters are point estimates obtained by maximizing the log marginal likelihood, but given these estimates its predictive covariance is exact. The uncertainty of BTT-KM, in contrast, also depends on the mean-field approximation of the posterior.

\input{UQ_results}

\paragraph{Accuracy, model complexity and run time.}
Table~\ref{tab:acc} shows that BTT-KM is competitive with BTN-KM in predictive accuracy. It is the most accurate model on Airfoil, while BTN-KM is slightly more accurate on Yacht, Energy and Concrete, although on Concrete the difference is within one standard deviation. On the classification data sets, both models reach the same error on Adult, and BTN-KM is more accurate on Spambase. The run times of the two models are comparable on the low-dimensional data sets, while BTT-KM is slower on the high-dimensional ones, taking about 2.3 times as long on Spambase ($D=57$) and 1.25 times as long on Adult ($D=96$).

Unlike BTN-KM, BTT-KM does not need annealing. When trained without annealing but with the same $a_0$ and initialization, BTN-KM either converges to a rank-one solution with an error of 0.23-0.28 or truncates all rank components on 7 of the 10 Spambase splits, whereas BTT-KM reaches an error between 0.065 and 0.102 on 9 splits and 0.195 on the remaining one. This agrees with the expectation in Section~\ref{sec:introduction} that, with randomly initialized cores, the TT output is better behaved than the CPD output for large $D$.

 Since each TT rank is inferred separately, BTT-KM can adapt its model complexity along the train. On Concrete, for example, the inferred ranks decrease from their initial value of 5 at the first edges to 2-3 at the last ones, with average ranks over the 10 splits of $(R_2,\ldots,R_8) = (5.0, 4.3, 4.8, 4.0, 2.9, 3.6, 2.7)$. Note that TT and CP ranks are not directly comparable, since the size of a TT core grows quadratically with its ranks.

\paragraph{Uncertainty quantification.}
The uncertainty quantification results in Table~\ref{tab:uq} are consistent with the accuracy results in Table~\ref{tab:acc}. BTN-KM has a lower NLL than BTT-KM on Yacht, Energy and Concrete, and BTT-KM has the lower NLL on Airfoil, while the GP has the lowest NLL on Yacht. Since the NLL reflects both the accuracy and the calibration of the predictive distribution, these differences largely follow the RMSE. In terms of calibration alone, the analytic intervals of BTT-KM are closest to the nominal 95\% coverage, with an average coverage error of $\dpp=0.9$ percentage points, compared with 1.3 for BTN-KM and 1.4 for the GP. For both TN models, the analytic predictive distribution closely matches its MC estimate, which confirms that the Student's-t approximation is accurate. For classification, BTT-KM and BTN-KM reach a similar NLL on Adult ($0.329$ and $0.332$), BTN-KM has a lower NLL on Spambase ($0.164$ versus $0.257$), and both are well below the GP on Spambase ($0.720$).

%% file: predictive_performance.tex
\begin{table}[t]
\centering
\small
\setlength{\tabcolsep}{1.8pt}
\resizebox{\linewidth}{!}{%
\begin{tabular}{lrrr r@{}lr@{}l r@{}lr@{}lr@{}l r@{}lr@{}l}
 & & & & \multicolumn{4}{c}{$R_{\mathrm{eff}}$}
 & \multicolumn{6}{c}{RMSE / error} & \multicolumn{4}{c}{Time (s)}\\
\cmidrule(lr){5-8}\cmidrule(lr){9-14}\cmidrule(lr){15-18}
Dataset & $N$ & $D$ & $M$ & \multicolumn{2}{c}{BTT} & \multicolumn{2}{c}{BTN} & \multicolumn{2}{c}{GP}
 & \multicolumn{2}{c}{BTT} & \multicolumn{2}{c}{BTN} & \multicolumn{2}{c}{BTT} & \multicolumn{2}{c}{BTN}\\
\midrule
Yacht        & 308   & 6  & 20 & 4.3 & \sd{0.2} & 4.6  & \sd{0.7} & 0.402 & \sd{0.124} & 0.587 & \sd{0.144} & \textbf{0.401} & \sd{0.154} & 5.3   & \sd{0.4}    & 3.4   & \sd{0.6}\\
Energy       & 768   & 8  & 20 & 4.8 & \sd{0.1} & 6.5  & \sd{0.7} & 0.448 & \sd{0.037} & 0.480 & \sd{0.054} & \textbf{0.412} & \sd{0.060} & 10.6  & \sd{0.2}    & 6.3   & \sd{0.7}\\
Concrete     & 1030  & 8  & 20 & 3.9 & \sd{0.1} & 8.3  & \sd{0.5} & 5.565 & \sd{0.494} & 5.083 & \sd{0.611} & \textbf{4.969} & \sd{0.487} & 9.4   & \sd{0.4}    & 12.6  & \sd{2.8}\\
Airfoil      & 1503  & 5  & 20 & 4.9 & \sd{0.1} & 9.4  & \sd{0.8} & 2.293 & \sd{0.189} & \textbf{1.526} & \sd{0.161} & 1.723 & \sd{0.180} & 6.9   & \sd{0.2}    & 7.5   & \sd{1.4}\\
Spambase$^*$ & 4601  & 57 & 30 & 2.8 & \sd{0.2} & 3.3  & \sd{0.6} & 0.095 & \sd{0.016} & 0.094 & \sd{0.036} & \textbf{0.054} & \sd{0.008} & 266.6 & \sd{12.6}   & 118.3 & \sd{22.4}\\
Adult$^*$    & 45222 & 96 & 40 & 3.0 & \sd{0.0} & 4.1  & \sd{0.8} & \multicolumn{2}{c}{N/A} & \textbf{0.145} & \sd{0.005} & \textbf{0.145} & \sd{0.005} & 2446.2 & \sd{69.1} & 1955.0 &\sd{386.0}\\
\bottomrule
\end{tabular}%
}
\caption{\footnotesize{Data sets, model complexity and accuracy (mean $\pm$ standard deviation over
the 10 splits). $N$, $D$: number of samples and inputs; $M$: feature dimension;
$R_{\mathrm{eff}}$: effective rank; RMSE, or misclassification
error for the classification data sets ($^*$); time: per split, on an Apple M2 Pro
with 16\,GB RAM. Bold: lowest RMSE/error.}}
\label{tab:acc}
\end{table}

%% file: UQ_results.tex
\begin{table}[t]
\centering
\setlength{\tabcolsep}{4pt}
\resizebox{\linewidth}{!}{%
\begin{tabular}{@{}l l ccccc@{}}
 & & & \multicolumn{2}{c}{BTT-KM} & \multicolumn{2}{c}{BTN-KM}\\
\cmidrule(lr){4-5}\cmidrule(lr){6-7}
 & Dataset  & GP & Analytic & MC & Analytic & MC\\
\midrule
\multirow{4}{*}{\textbf{NLL}}
 & Yacht    & \textbf{0.105\,$\pm$\,0.319} & 0.814\,$\pm$\,0.254 & 0.818\,$\pm$\,0.256 & 0.347\,$\pm$\,0.350 & 0.378\,$\pm$\,0.388\\
 & Energy   & 0.608\,$\pm$\,0.091 & 0.708\,$\pm$\,0.091 & 0.722\,$\pm$\,0.081 & \textbf{0.537\,$\pm$\,0.137} & 0.542\,$\pm$\,0.137\\
 & Concrete & 3.078\,$\pm$\,0.076 & 3.052\,$\pm$\,0.132 & 3.050\,$\pm$\,0.126 & \textbf{3.030\,$\pm$\,0.140} & 3.020\,$\pm$\,0.126\\
 & Airfoil  & 2.281\,$\pm$\,0.080 & \textbf{1.810\,$\pm$\,0.095} & 1.812\,$\pm$\,0.093 & 1.940\,$\pm$\,0.101 & 1.942\,$\pm$\,0.103\\
\midrule
\multirow{5}{*}{\shortstack[l]{\textbf{PICP-95}\\\textbf{(\%)}}}
 & Yacht    & 92.6\,$\pm$\,6.6 & 95.8\,$\pm$\,2.9 & 96.1\,$\pm$\,3.2 & \textbf{94.8\,$\pm$\,4.1} & 94.8\,$\pm$\,4.1\\
 & Energy   & \textbf{93.9\,$\pm$\,2.0} & \textbf{96.1\,$\pm$\,2.7} & 97.3\,$\pm$\,1.5 & 96.4\,$\pm$\,2.0 & 96.4\,$\pm$\,2.2\\
 & Concrete & 93.3\,$\pm$\,2.1 & \textbf{93.7\,$\pm$\,2.5} & 93.8\,$\pm$\,2.2 & 91.3\,$\pm$\,2.9 & 91.4\,$\pm$\,2.9\\
 & Airfoil  & 95.2\,$\pm$\,1.0 & 95.2\,$\pm$\,1.5 & 95.4\,$\pm$\,1.6 & \textbf{95.0\,$\pm$\,1.2} & 95.0\,$\pm$\,1.1\\
    & $\dpp$   & 1.4 & \textbf{0.9} & 1.3 & 1.3 & 1.3\\
\bottomrule
\end{tabular}%
}
\caption{\footnotesize{Uncertainty quantification on the regression data sets: test NLL (lower is better)
and PICP-95 (closer to 95\% is better) of the GP and of the analytic and MC predictive
distributions of BTT-KM and BTN-KM. $\dpp$: mean absolute deviation of PICP-95 from 95\%
(percentage points). Bold: best among GP and the analytic distributions.}}
\label{tab:uq}
\end{table}

%% file: sections/5.conclusion.tex
\section{Conclusion}
\label{sec:conclusion}

We have presented BTT-KM, a probabilistic treatment of TT-based kernel machines. By placing ARD-type hierarchical priors on the TT cores, the model automatically infers all TT ranks and feature dimensions, allowing its model complexity to vary along the train. To make Bayesian learning tractable, we employed mean-field variational inference, resulting in a Bayesian ALS algorithm with the same computational complexity as deterministic TT-ALS, so uncertainty quantification comes at no additional asymptotic cost. A synthetic experiment confirms that the regularization shrinks an over-parameterized model to a low-rank solution close to the ground truth. Empirical evaluations on six UCI data sets demonstrate that BTT-KM achieves competitive performance in both predictive accuracy and uncertainty quantification, with predictive intervals closest to the nominal coverage. Unlike BTN-KM, it also does not require an annealing phase.

%% file: sections/6.appendix.tex
\vspace*{3cm}
\begin{center}
    {\large \textbf{Appendix}}
\end{center}

\vspace{0.8cm}

\begin{center}
    \begin{minipage}{0.85\textwidth}
        This appendix collects the proofs and derivations supporting the main text.
        We refer the reader to Bishop (2006)\footnote{C.~M.~Bishop, \textit{Pattern
        Recognition and Machine Learning}, Springer, New York, 2006.} for background
        on variational inference and exponential families.

        \vspace{0.4cm}

        In variational Bayesian (VB) inference, the intractable posterior
        $p(\bm{\Theta} \mid \mathbf{y})$ is approximated by a factorized distribution
        $q(\bm{\Theta}) = \prod_j q_j(\theta_j)$, where each factor is updated via
        \begin{equation*}
            \ln q_j(\theta_j)
            = \mathbb{E}_{q(\bm{\Theta} \setminus \theta_j)}
            [\ln p(\mathbf{y}, \bm{\Theta})] + \text{const}.
        \end{equation*}
        When the resulting expression matches the log of a known distribution,
        the update is available in closed form. Section~1 proves Theorem~4, which isolates the $d$th core in the model output, and Section~2 expands the log-joint distribution. Section~3 derives the TT core update, whose expected Gram matrix is evaluated by Theorem~6 (proved in Section~4) and the forward and backward Gram recursions (Section~5). Sections~6-8 derive the posteriors of the feature precisions, rank precisions and noise precision. Section~9 derives the ELBO used as the convergence criterion, and Section~10 the approximate predictive distribution.

        \vspace{0.4cm}

        \textit{Throughout the posterior derivations, exponential family
        posteriors are identified by completing the square or by matching
        coefficients with the log of a known distribution.}
    \end{minipage}
\end{center}

\newpage

\textbf{1. Proof of Theorem 4}
\begin{proof}[\textit{Theorem 4}]
We rewrite the data fitting term $\mathbf{\Phi}^{\top}\mathbf{w}$
in two equivalent forms.
The first isolates $\bm{\mathcal{W}}^{(d)}$, giving the
design matrix $\mathbf{G}_{\backslash d}$ needed for the
core updates. The second is the full model expression
used in the residual computations.

\begin{align*}
\mathbf{\Phi}^{\top}\mathbf{w}
&=
\left(
\mathbf{\Phi}^{(1)}\khatri\cdots\khatri\mathbf{\Phi}^{(D)}
\right)^{\top}
\left[
\sum_{r_2=1}^{R_2}
\sum_{r_3=1}^{R_3}
\cdots
\sum_{r_D=1}^{R_D}
\mathbf{w}_{1:r_2}^{(1)}
\otimes
\mathbf{w}_{r_2:r_3}^{(2)}
\otimes\cdots\otimes
\mathbf{w}_{r_D:1}^{(D)}
\right]
\\
&=
\sum_{r_2=1}^{R_2}
\sum_{r_3=1}^{R_3}
\cdots
\sum_{r_D=1}^{R_D}
\left(
\mathbf{\Phi}^{(1)}\khatri\cdots\khatri\mathbf{\Phi}^{(D)}
\right)^{\top}
\left(
\mathbf{w}_{1:r_2}^{(1)}
\otimes
\mathbf{w}_{r_2:r_3}^{(2)}
\otimes\cdots\otimes
\mathbf{w}_{r_D:1}^{(D)}
\right)
\\
&=
\sum_{r_2=1}^{R_2}
\cdots
\sum_{r_D=1}^{R_D}
\sum_{m_1=1}^{M_1}
\cdots
\sum_{m_D=1}^{M_D}
\left(
\mathbf{\Phi}^{(1)}_{m_1,:}\,w_{1m_1r_2}^{(1)}
\right)
\hadamard\cdots\hadamard
\left(
\mathbf{\Phi}^{(d)}_{m_d,:}\,w_{r_dm_dr_{d+1}}^{(d)}
\right)
\hadamard\cdots\hadamard
\left(
\mathbf{\Phi}^{(D)}_{m_D,:}\,w_{r_Dm_D1}^{(D)}
\right)
\\
&=
\sum_{r_2=1}^{R_2}
\cdots
\sum_{r_D=1}^{R_D}
\left(
\sum_{m_1=1}^{M_1}
\mathbf{\Phi}^{(1)}_{m_1,:}\,w_{1m_1r_2}^{(1)}
\right)
\hadamard\cdots\hadamard
\left(
\sum_{m_d=1}^{M_d}
\mathbf{\Phi}^{(d)}_{m_d,:}\,w_{r_dm_dr_{d+1}}^{(d)}
\right)
\hadamard\cdots\hadamard
\left(
\sum_{m_D=1}^{M_D}
\mathbf{\Phi}^{(D)}_{m_D,:}\,w_{r_Dm_D1}^{(D)}
\right)
\\
\intertext{We now isolate the $d$th core by grouping the
contributions of the preceding and following modes:}
&=
\sum_{r_d=1}^{R_d}
\sum_{r_{d+1}=1}^{R_{d+1}}
\sum_{m_d=1}^{M_d}
\left[
\sum_{r_2=1}^{R_2}
\cdots
\sum_{r_{d-1}=1}^{R_{d-1}}
\mathop{\hadamard}_{k=1}^{d-1}
\left(
\sum_{m_k=1}^{M_k}
\mathbf{\Phi}^{(k)}_{m_k,:} \,
w_{r_km_kr_{k+1}}^{(k)}
\right)
\right]
\hadamard
\left(
\mathbf{\Phi}^{(d)}_{m_d,:}\,
w_{r_dm_dr_{d+1}}^{(d)}
\right)
\\
&\qquad{}\hadamard
\left[
\sum_{r_{d+2}=1}^{R_{d+2}}
\cdots
\sum_{r_D=1}^{R_D}
\mathop{\hadamard}_{k=d+1}^{D}
\left(
\sum_{m_k=1}^{M_k}
\mathbf{\Phi}^{(k)}_{m_k,:}\,
w_{r_km_kr_{k+1}}^{(k)}
\right)
\right]
\\
&=
\sum_{r_d=1}^{R_d}
\sum_{r_{d+1}=1}^{R_{d+1}}
\left[
\sum_{r_2=1}^{R_2}
\cdots
\sum_{r_{d-1}=1}^{R_{d-1}}
\mathop{\hadamard}_{k=1}^{d-1}
\left(
\mathbf{\Phi}^{(k)^\top}
\mathbf{w}_{r_k:r_{k+1}}^{(k)}
\right)
\right]
\hadamard
\left(
\mathbf{\Phi}^{(d)^\top}
\mathbf{w}_{r_d:r_{d+1}}^{(d)}
\right)
\hadamard
\left[
\sum_{r_{d+2}=1}^{R_{d+2}}
\cdots
\sum_{r_D=1}^{R_D}
\mathop{\hadamard}_{k=d+1}^{D}
\left(
\mathbf{\Phi}^{(k)^\top}
\mathbf{w}_{r_k:r_{k+1}}^{(k)}
\right)
\right]
\\
&=
\sum_{r_d=1}^{R_d}
\sum_{r_{d+1}=1}^{R_{d+1}}
(\mathbf{G}_{<d})_{r_d,:}
\hadamard
\left(
\mathbf{\Phi}^{(d)^\top}
\mathbf{w}_{r_d:r_{d+1}}^{(d)}
\right)
\hadamard
(\mathbf{G}_{>d})_{r_{d+1},:}
\\
&=
\left(
\mathbf{G}_{>d}
\khatri
\mathbf{\Phi}^{(d)}
\khatri
\mathbf{G}_{<d}
\right)^\top
\operatorname{vec}(\bm{\mathcal{W}}^{(d)})
\\
&=
\mathbf{G}_{\backslash d}^\top
\operatorname{vec}(\bm{\mathcal{W}}^{(d)}).
\end{align*}
\end{proof}
Notice that if we do not isolate the $d$th core,
the data fitting term can also be written as
\begin{equation*}
\mathbf{\Phi}^{\top}\mathbf{w}
=
\sum_{r_2=1}^{R_2}
\sum_{r_3=1}^{R_3}
\cdots
\sum_{r_D=1}^{R_D}
\left(
\mathop{\hadamard}_{d=1}^{D}
\left(
\mathbf{\Phi}^{(d)^{\top}}
\mathbf{w}_{r_d:r_{d+1}}^{(d)}
\right)
\right)
=
\mathbf{G}_{<D+1}^{\top},
\end{equation*}
where $\hadamard$ denotes the element-wise product
and $\mathbf{G}_{<D+1}$ is obtained by continuing
the forward recursion through all $D$ cores.
This expression is used to evaluate the posterior
expectation of the squared residual.

\newpage

\vspace{1cm}
\textbf{2. The log of the joint distribution}
We expand the log of the joint distribution $p(\mathbf{y}, \bm{\Theta})$ by substituting 
the likelihood and all priors, and collect terms by parameter. The resulting expression 
is the starting point for all variational updates, since each posterior $q_j(\theta_j)$ 
is obtained by taking the expectation of this log-joint over all parameters except $\theta_j$.

\begin{equation*}
\begin{aligned}
l(\bm{\Theta})
&=
\sum_{n=1}^{N}
\left(
\frac{1}{2}\ln\tau
-
\frac{\tau}{2}
\left(
y_n-\bm{\varphi}(\mathbf{x}_n)^\top\mathbf{w}
\right)^2
\right)
\\
&\quad
+
\sum_{d=1}^{D}
\sum_{r_d=1}^{R_d}
\sum_{m_d=1}^{M_d}
\sum_{r_{d+1}=1}^{R_{d+1}}
\left(
\frac{1}{2}
\ln
\left|
\lambda_{r_d}\delta_{m_d}\lambda_{r_{d+1}}
\right|
-
\frac{1}{2}
w^{(d)}_{r_d m_d r_{d+1}}
\lambda_{r_d}\delta_{m_d}\lambda_{r_{d+1}}
w^{(d)}_{r_d m_d r_{d+1}}
\right)
\\
&\quad
+
\sum_{d=2}^{D}
\sum_{r_d=1}^{R_d}
\left(
(c_0^{r_d}-1)\ln\lambda_{r_d}
-
e_0^{r_d}\lambda_{r_d}
\right)
+
\sum_{d=1}^{D}
\sum_{m_d=1}^{M_d}
\left(
(g_0^{m_d}-1)\ln\delta_{m_d}
-
h_0^{m_d}\delta_{m_d}
\right)
+
(a_0-1)\ln\tau
-
b_0\tau
+
\text{const}
\\[2mm]
&=
\frac{N}{2}\ln\tau
-
\frac{\tau}{2}
\sum_{n=1}^{N}
\left(
y_n-\bm{\varphi}(\mathbf{x}_n)^\top\mathbf{w}
\right)^2
+
\frac{1}{2}
\sum_{d=1}^{D}
M_dR_{d+1}\ln|\bm{\Lambda}_d|
+
\frac{1}{2}
\sum_{d=1}^{D}
R_dM_d\ln|\bm{\Lambda}_{d+1}|
+
\frac{1}{2}
\sum_{d=1}^{D}
R_dR_{d+1}\ln|\bm{\Delta}_d|
\\
&\quad
-
\frac{1}{2}
\sum_{d=1}^{D}
\sum_{r_d=1}^{R_d}
\sum_{m_d=1}^{M_d}
\sum_{r_{d+1}=1}^{R_{d+1}}
w^{(d)}_{r_d m_d r_{d+1}}
\lambda_{r_d}\delta_{m_d}\lambda_{r_{d+1}}
w^{(d)}_{r_d m_d r_{d+1}}
+
\sum_{d=2}^{D}
\sum_{r_d=1}^{R_d}
\left(
(c_0^{r_d}-1)\ln\lambda_{r_d}
-
e_0^{r_d}\lambda_{r_d}
\right)
\\
&\quad
+
\sum_{d=1}^{D}
\sum_{m_d=1}^{M_d}
\left(
(g_0^{m_d}-1)\ln\delta_{m_d}
-
h_0^{m_d}\delta_{m_d}
\right)
+
(a_0-1)\ln\tau
-
b_0\tau
+
\text{const}
\\[2mm]
&=
-
\frac{\tau}{2}
\left\|
\mathbf{y}
-
\mathbf{\Phi}^{\top}\mathbf{w}
\right\|_F^2
-
\frac{1}{2}
\sum_{d=1}^{D}
\sum_{r_d=1}^{R_d}
\sum_{m_d=1}^{M_d}
\sum_{r_{d+1}=1}^{R_{d+1}}
w^{(d)}_{r_d m_d r_{d+1}}
\lambda_{r_d}\delta_{m_d}\lambda_{r_{d+1}}
w^{(d)}_{r_d m_d r_{d+1}}
+
\left(
\frac{N}{2}
+
a_0
-
1
\right)
\ln\tau
\\
&\quad
+
\sum_{d=2}^{D}
\sum_{r_d=1}^{R_d}
\left(
\frac{R_{d-1}M_{d-1}+M_dR_{d+1}}{2}
+
c_0^{r_d}
-
1
\right)
\ln\lambda_{r_d}
+
\sum_{d=1}^{D}
\sum_{m_d=1}^{M_d}
\left(
\frac{R_dR_{d+1}}{2}
+
g_0^{m_d}
-
1
\right)
\ln\delta_{m_d}
\\
&\quad
-
\sum_{d=2}^{D}
\sum_{r_d=1}^{R_d}
e_0^{r_d}\lambda_{r_d}
-
\sum_{d=1}^{D}
\sum_{m_d=1}^{M_d}
h_0^{m_d}\delta_{m_d}
-
b_0\tau
+
\text{const}.
\end{aligned}
\end{equation*}

\noindent Here, the boundary rank precisions associated with $R_1=R_{D+1}=1$ are fixed and therefore do not contribute to the corresponding logarithmic terms.

This form of the log of the joint distribution will be used frequently in variational Bayesian inference, because the inference of each $\Theta_j$ can be done by
\begin{equation*}
\ln q_j(\Theta_j)
=
\mathbb{E}_{q(\bm{\Theta}\backslash\Theta_j)}
\left[
\ln p(\mathbf{y},\bm{\Theta})
\right]
+
\text{const}.
\end{equation*}

\newpage
\textbf{3. TT core update}
We derive the variational posterior of
$\operatorname{vec}(\bm{\mathcal{W}}^{(d)})$ by collecting only the terms in
the log-joint that depend on $\bm{\mathcal{W}}^{(d)}$. The residual term is
rewritten using the design matrix $\mathbf{G}_{\backslash d}$, and expanded
into a quadratic form in $\operatorname{vec}(\bm{\mathcal{W}}^{(d)})$.

\begin{equation*}
\begin{aligned}
\ln q\left(
\operatorname{vec}(\bm{\mathcal{W}}^{(d)})
\right)
&=
\mathbb{E}_{q(\bm{\Theta}\backslash\bm{\mathcal{W}}^{(d)})}
\left[
\ln p(\mathbf{y},\bm{\Theta})
\right]
+\text{const}
\\
&=
\mathbb{E}\left[
-
\frac{\tau}{2}
\left\|
\mathbf{y}
-
\mathbf{\Phi}^{\top}\mathbf{w}
\right\|_F^2
-
\frac{1}{2}
\sum_{d=1}^{D}
\sum_{r_d=1}^{R_d}
\sum_{m_d=1}^{M_d}
\sum_{r_{d+1}=1}^{R_{d+1}}
w^{(d)}_{r_d m_d r_{d+1}}
\lambda_{r_d}\delta_{m_d}\lambda_{r_{d+1}}
w^{(d)}_{r_d m_d r_{d+1}} \right] +\text{const}
\\
&=
\mathbb{E}\left[
-\frac{\tau}{2}
\left\|
\mathbf{y}-\mathbf{\Phi}^{\top}\mathbf{w}
\right\|_F^2
-
\frac{1}{2}
\operatorname{vec}(\bm{\mathcal{W}}^{(d)})^\top
\left(
\bm{\Lambda}_{d+1}
\otimes
\bm{\Delta}_{d}
\otimes
\bm{\Lambda}_{d}
\right)
\operatorname{vec}(\bm{\mathcal{W}}^{(d)})
\right]
+\text{const}
\\
&=
\mathbb{E}\left[
-\frac{\tau}{2}
\left\|
\mathbf{y}
-
\mathbf{G}_{\backslash d}^{\top}
\operatorname{vec}(\bm{\mathcal{W}}^{(d)})
\right\|_F^2
-
\frac{1}{2}
\operatorname{vec}(\bm{\mathcal{W}}^{(d)})^\top
\left(
\bm{\Lambda}_{d+1}
\otimes
\bm{\Delta}_{d}
\otimes
\bm{\Lambda}_{d}
\right)
\operatorname{vec}(\bm{\mathcal{W}}^{(d)})
\right]
+\text{const}
\\
&=
\mathbb{E}\left[
-\frac{\tau}{2}
\operatorname{vec}(\bm{\mathcal{W}}^{(d)})^\top
\mathbf{G}_{\backslash d}
\mathbf{G}_{\backslash d}^{\top}
\operatorname{vec}(\bm{\mathcal{W}}^{(d)})
-\frac{1}{2}
\operatorname{vec}(\bm{\mathcal{W}}^{(d)})^\top
\left(
\bm{\Lambda}_{d+1}
\otimes
\bm{\Delta}_{d}
\otimes
\bm{\Lambda}_{d}
\right)
\operatorname{vec}(\bm{\mathcal{W}}^{(d)})
\right.
\\
&\qquad\left.
+
\tau
\operatorname{vec}(\bm{\mathcal{W}}^{(d)})^\top
\mathbf{G}_{\backslash d}\mathbf{y}
\right]
+\text{const}
\\
&=
-\frac{1}{2}
\operatorname{vec}(\bm{\mathcal{W}}^{(d)})^\top
\Big(
\mathbb{E}_q[\tau]\,
\mathbb{E}_q[
\mathbf{G}_{\backslash d}
\mathbf{G}_{\backslash d}^{\top}]
+
\mathbb{E}_q[\bm{\Lambda}_{d+1}]
\otimes
\mathbb{E}_q[\bm{\Delta}_{d}]
\otimes
\mathbb{E}_q[\bm{\Lambda}_{d}]
\Big)
\operatorname{vec}(\bm{\mathcal{W}}^{(d)})
\\
&\quad
+
\operatorname{vec}(\bm{\mathcal{W}}^{(d)})^\top
\mathbb{E}_q[\tau]\,
\mathbb{E}_q[\mathbf{G}_{\backslash d}]
\mathbf{y}
+\text{const}.
\end{aligned}
\end{equation*}

\noindent Identifying this as a log-Gaussian in
$\operatorname{vec}(\bm{\mathcal{W}}^{(d)})$ by completing the square, the
posterior mean and covariance are given by

\begin{equation*}
\begin{aligned}
\operatorname{vec}(\tilde{\bm{\mathcal{W}}}^{(d)})
&=
\mathbb{E}_q[\tau]\,
\mathbf{\Sigma}^{(d)}
\mathbb{E}_q[\mathbf{G}_{\backslash d}]
\mathbf{y},
\\
\mathbf{\Sigma}^{(d)}
&=
\left[
\mathbb{E}_q[\tau]\,
\mathbb{E}_q[
\mathbf{G}_{\backslash d}
\mathbf{G}_{\backslash d}^{\top}]
+
\mathbb{E}_q[\bm{\Lambda}_{d+1}]
\otimes
\mathbb{E}_q[\bm{\Delta}_{d}]
\otimes
\mathbb{E}_q[\bm{\Lambda}_{d}]
\right]^{-1}.
\end{aligned}
\end{equation*}

\newpage

\textbf{4. Proof of Theorem 6}
\begin{proof}[\textit{Theorem 6}]
We rewrite the expected Gram matrix $\mathbb{E}_q[\mathbf{G}_{\backslash d}\mathbf{G}_{\backslash d}^{\top}]$ in terms of the left and right TT contractions, which can be evaluated recursively. This exploits the column-wise Kronecker structure of the Khatri--Rao product: by the transpose and mixed-product properties of the Kronecker product,
\[
(\mathbf{A}\otimes\mathbf{B})^{\top}
=
\mathbf{A}^{\top}\otimes\mathbf{B}^{\top},
\qquad
(\mathbf{A}\otimes\mathbf{B})
(\mathbf{C}\otimes\mathbf{D})
=
(\mathbf{A}\mathbf{C})\otimes(\mathbf{B}\mathbf{D}),
\]
the outer product of each column of $\mathbf{G}_{\backslash d}$ splits into left, feature, and right factors. Since $\mathbf{\Phi}^{(d)}$ is deterministic and the cores $\{\bm{\mathcal{W}}^{(k)}\}_{k\neq d}$ are mutually independent under $q$, the expectation then factorizes over the left and right factors.
\begin{align*}
\mathbb{E}_q\!\left[
\mathbf{G}_{\backslash d}
\mathbf{G}_{\backslash d}^{\top}
\right]
&=
\mathbb{E}_q\!\left[
\left(
\mathbf{G}_{>d}
\khatri\mathbf{\Phi}^{(d)}
\khatri\mathbf{G}_{<d}
\right)
\left(
\mathbf{G}_{>d}
\khatri\mathbf{\Phi}^{(d)}
\khatri\mathbf{G}_{<d}
\right)^{\top}
\right]
\\
&=
\sum_{n=1}^{N}
\mathbb{E}_q\!\left[
\left(
\mathbf{G}_{>d}
\khatri\mathbf{\Phi}^{(d)}
\khatri\mathbf{G}_{<d}
\right)_{:,n}
\left[
\left(
\mathbf{G}_{>d}
\khatri\mathbf{\Phi}^{(d)}
\khatri\mathbf{G}_{<d}
\right)_{:,n}
\right]^{\top}
\right]
\\
&=
\sum_{n=1}^{N}
\mathbb{E}_q\!\left[
\left(
(\mathbf{G}_{>d})_{:,n}
\otimes\mathbf{\Phi}^{(d)}_{:,n}
\otimes(\mathbf{G}_{<d})_{:,n}
\right)
\left(
(\mathbf{G}_{>d})_{:,n}^{\top}
\otimes(\mathbf{\Phi}^{(d)}_{:,n})^{\top}
\otimes(\mathbf{G}_{<d})_{:,n}^{\top}
\right)
\right]
\\
&=
\sum_{n=1}^{N}
\mathbb{E}_q\!\left[
\left(
(\mathbf{G}_{>d})_{:,n}
(\mathbf{G}_{>d})_{:,n}^{\top}
\right)
\otimes
\left(
\mathbf{\Phi}^{(d)}_{:,n}
(\mathbf{\Phi}^{(d)}_{:,n})^{\top}
\right)
\otimes
\left(
(\mathbf{G}_{<d})_{:,n}
(\mathbf{G}_{<d})_{:,n}^{\top}
\right)
\right]
\\
&=
\sum_{n=1}^{N}
\mathbb{E}_q\!\left[
(\mathbf{G}_{>d})_{:,n}
(\mathbf{G}_{>d})_{:,n}^{\top}
\right]
\otimes
\left(
\mathbf{\Phi}^{(d)}_{:,n}
(\mathbf{\Phi}^{(d)}_{:,n})^{\top}
\right)
\otimes
\mathbb{E}_q\!\left[
(\mathbf{G}_{<d})_{:,n}
(\mathbf{G}_{<d})_{:,n}^{\top}
\right]
\\
&=
\bm{\mathcal{R}}\!\left\{
\sum_{n=1}^{N}
\left(
\mathbb{E}_q\!\left[
(\mathbf{G}_{<d})_{:,n}
\otimes(\mathbf{G}_{<d})_{:,n}
\right]
\otimes\mathbf{\Phi}^{(d)}_{:,n}
\right)
\left(
\mathbf{\Phi}^{(d)}_{:,n}
\otimes
\mathbb{E}_q\!\left[
(\mathbf{G}_{>d})_{:,n}
\otimes(\mathbf{G}_{>d})_{:,n}
\right]
\right)^{\top}
\right\}_{R_dM_dR_{d+1}, R_dM_dR_{d+1}}
\\
&=
\bm{\mathcal{R}}\!\left\{
\sum_{n=1}^{N}
\left(
\mathbb{E}_q[
\mathbf{G}_{<d}\khatri\mathbf{G}_{<d}]
\khatri\mathbf{\Phi}^{(d)}
\right)_{:,n}
\left[
\left(
\mathbf{\Phi}^{(d)}
\khatri
\mathbb{E}_q[
\mathbf{G}_{>d}\khatri\mathbf{G}_{>d}]
\right)_{:,n}
\right]^{\top}\right\}_{R_dM_dR_{d+1}, R_dM_dR_{d+1}}
\\
&=
\bm{\mathcal{R}}\!\left\{
\left(
\mathbb{E}_q[
\mathbf{G}_{<d}\khatri\mathbf{G}_{<d}]
\khatri\mathbf{\Phi}^{(d)}
\right)
\left(
\mathbf{\Phi}^{(d)}
\khatri
\mathbb{E}_q[
\mathbf{G}_{>d}\khatri\mathbf{G}_{>d}]
\right)^{\top}
\right\}_{R_dM_dR_{d+1}, R_dM_dR_{d+1}}.
\end{align*}
\end{proof}

\newpage

\textbf{5. Derivation of the Gram recursions}
Let $\mathbf{P}$ denote the fixed permutation matrix that reorders the
Kronecker factors according to the adopted index ordering. Using the
mixed-product property of the Khatri--Rao product,
\[
(\mathbf{A}\mathbf{X})\khatri(\mathbf{A}\mathbf{X})
=
(\mathbf{A}\otimes\mathbf{A})(\mathbf{X}\khatri\mathbf{X}),
\]
with $\mathbf{X}=\mathbf{B}\khatri\mathbf{C}$ gives
\[
\bigl(\mathbf{A}(\mathbf{B}\khatri\mathbf{C})\bigr)
\khatri
\bigl(\mathbf{A}(\mathbf{B}\khatri\mathbf{C})\bigr)
=
(\mathbf{A}\otimes\mathbf{A})
\bigl[
(\mathbf{B}\khatri\mathbf{C})
\khatri
(\mathbf{B}\khatri\mathbf{C})
\bigr].
\]
For each column $n$,
\[
(\mathbf{b}_n\otimes\mathbf{c}_n)
\otimes
(\mathbf{b}_n\otimes\mathbf{c}_n)
=
\mathbf{P}
\bigl[
(\mathbf{b}_n\otimes\mathbf{b}_n)
\otimes
(\mathbf{c}_n\otimes\mathbf{c}_n)
\bigr].
\]
Hence,
\[
(\mathbf{B}\khatri\mathbf{C})
\khatri
(\mathbf{B}\khatri\mathbf{C})
=
\mathbf{P}
\bigl[
(\mathbf{B}\khatri\mathbf{B})
\khatri
(\mathbf{C}\khatri\mathbf{C})
\bigr],
\]
and therefore
\begin{equation}
\label{eq:key}
\bigl(\mathbf{A}(\mathbf{B}\khatri\mathbf{C})\bigr)
\khatri
\bigl(\mathbf{A}(\mathbf{B}\khatri\mathbf{C})\bigr)
=
(\mathbf{A}\otimes\mathbf{A})
\mathbf{P}
\bigl[
(\mathbf{B}\khatri\mathbf{B})
\khatri
(\mathbf{C}\khatri\mathbf{C})
\bigr].
\end{equation}

\vspace{1cm}
\noindent \textit{Forward accumulator.} Given the definition $\mathbf{G}_{<k}: = \mathbf{W}^{(k-1)}_{(3)}
\bigl(
\mathbf{\Phi}^{(k-1)}
\khatri
\mathbf{G}_{<k-1}
\bigr)$, take
$\mathbf{A}=\mathbf{W}^{(k-1)}_{(3)}$,
$\mathbf{B}=\mathbf{\Phi}^{(k-1)}$, and
$\mathbf{C}=\mathbf{G}_{<k-1}$ in \eqref{eq:key} 
\[
\begin{aligned}
\mathbb{E}_q[\mathbf{G}_{<k}\khatri\mathbf{G}_{<k}]
&=
\mathbb{E}_q\!\left[
\mathbf{W}^{(k-1)}_{(3)}
\otimes
\mathbf{W}^{(k-1)}_{(3)}
\right]
\mathbf{P}
\left[
(\mathbf{\Phi}^{(k-1)}\khatri\mathbf{\Phi}^{(k-1)})
\khatri
\mathbb{E}_q[
\mathbf{G}_{<k-1}\khatri\mathbf{G}_{<k-1}]
\right].
\end{aligned}
\]
Since $\mathbf{W}^{(k-1)}$ is independent of the preceding TT cores, the
expectation factorizes as above. Together with the boundary condition
$\mathbf{H}_{<1}=\mathbf{1}_N^\top$, this is precisely the forward recursion,
and hence
$\mathbf{H}_{<k}=\mathbb{E}_q[\mathbf{G}_{<k}\khatri\mathbf{G}_{<k}]$.

\vspace{1cm}
\noindent \textit{Backward accumulator.} Given the definition $\mathbf{G}_{>k}:=
\mathbf{W}^{(k+1)}_{(1)}
\bigl(
\mathbf{G}_{>k+1}
\khatri
\mathbf{\Phi}^{(k+1)}
\bigr)$, take
$\mathbf{A}=\mathbf{W}^{(k+1)}_{(1)}$,
$\mathbf{B}=\mathbf{G}_{>k+1}$, and
$\mathbf{C}=\mathbf{\Phi}^{(k+1)}$ in \eqref{eq:key}
\[
\begin{aligned}
\mathbb{E}_q[\mathbf{G}_{>k}\khatri\mathbf{G}_{>k}]
&=
\mathbb{E}_q\!\left[
\mathbf{W}^{(k+1)}_{(1)}
\otimes
\mathbf{W}^{(k+1)}_{(1)}
\right]
\mathbf{P}
\left[
\mathbb{E}_q[\mathbf{G}_{>k+1}\khatri\mathbf{G}_{>k+1}]
\khatri
(\mathbf{\Phi}^{(k+1)}\khatri\mathbf{\Phi}^{(k+1)})
\right].
\end{aligned}
\]
Since $\mathbf{W}^{(k+1)}$ is independent of the subsequent TT cores, the
expectation factorizes as above. Together with the boundary condition
$\mathbf{H}_{>D}=\mathbf{1}_N^\top$, this is precisely the backward recursion,
and hence
$\mathbf{H}_{>k}=\mathbb{E}_q[\mathbf{G}_{>k}\khatri\mathbf{G}_{>k}]$.

\newpage

\newpage
\textbf{6. The variational posterior distribution of feature precisions $\bm{\delta}_{d}$}
We derive the variational posterior of $\bm{\delta}_{d}$ by collecting only the terms
in the log-joint that depend on $\bm{\delta}_{d}$.

\begin{equation*}
\begin{aligned}
\ln q(\bm{\delta}_{d})
&=
\mathbb{E}_{q(\bm{\Theta}\backslash\bm{\delta}_{d})}
\left[
\ln p(\mathbf{y},\bm{\Theta})
\right]
+\text{const}
\\
&=
\mathbb{E}\Bigg[
-
\frac{1}{2}
\sum_{d=1}^{D}
\sum_{r_d=1}^{R_d}
\sum_{m_d=1}^{M_d}
\sum_{r_{d+1}=1}^{R_{d+1}}
w^{(d)}_{r_d m_d r_{d+1}}
\lambda_{r_d}\delta_{m_d}\lambda_{r_{d+1}}
w^{(d)}_{r_d m_d r_{d+1}}
\\
&\hspace{1cm}
+
\sum_{m_d=1}^{M_d}
\left(
\frac{R_dR_{d+1}}{2}
+
g_0^{m_d}-1
\right)
\ln\delta_{m_d}
-
\sum_{m_d=1}^{M_d}
h_0^{m_d}\delta_{m_d}
\Bigg]
+\text{const}
\\
&=
\mathbb{E}\Bigg[
\sum_{m_d=1}^{M_d}
\Bigg\{
-\frac{1}{2}
\sum_{r_d=1}^{R_d}
\sum_{r_{d+1}=1}^{R_{d+1}}
w^{(d)}_{r_dm_dr_{d+1}}
\lambda_{r_d}
\lambda_{r_{d+1}}
w^{(d)}_{r_dm_dr_{d+1}}
\,\delta_{m_d}
+
\left(
g_0^{m_d}+\frac{R_dR_{d+1}}{2}-1
\right)
\ln\delta_{m_d}
-
h_0^{m_d}\delta_{m_d}
\Bigg\}
\Bigg]\\
 &\hspace{1cm}+\text{const}
\\
&=
\sum_{m_d=1}^{M_d}
\Bigg\{
\left(
g_0^{m_d}+\frac{R_dR_{d+1}}{2}-1
\right)
\ln\delta_{m_d}
-
\left(
h_0^{m_d}
+
\frac{1}{2}
\mathbb{E}_q\!\left[
\operatorname{vec}\!\left(\mathbf{W}^{(d)}_{:m_d:}\right)^{\top}
\left(
\bm{\Lambda}_{d+1}\otimes\bm{\Lambda}_{d}
\right)
\operatorname{vec}\!\left(\mathbf{W}^{(d)}_{:m_d:}\right)
\right]
\right)
\delta_{m_d}
\Bigg\}\\
 &\hspace{1cm}
+\text{const}. 
\end{aligned}
\end{equation*}

\noindent Identifying this as a log-Gamma in $\delta_{m_d}$, the posterior is a Gamma
distribution $\operatorname{Ga}(\delta_{m_d}\mid g_N^{m_d},h_N^{m_d})$
with updated parameters

\begin{equation*}
\begin{aligned}
g_N^{m_d}
&=
g_0^{m_d}+\frac{R_dR_{d+1}}{2},
\\
h_N^{m_d}
&=
h_0^{m_d}
+
\frac{1}{2}
\mathbb{E}_q\!\left[
\operatorname{vec}\!\left(\mathbf{W}^{(d)}_{:m_d:}\right)^{\top}
\left(
\bm{\Lambda}_{d+1}\otimes\bm{\Lambda}_{d}
\right)
\operatorname{vec}\!\left(\mathbf{W}^{(d)}_{:m_d:}\right)
\right].
\end{aligned}
\end{equation*}

\noindent The posterior expectation can be computed as

\begin{equation*}
\begin{aligned}
&
\mathbb{E}_q\!\left[
\operatorname{vec}\!\left(\mathbf{W}^{(d)}_{:m_d:}\right)^{\top}
\left(
\bm{\Lambda}_{d+1}\otimes\bm{\Lambda}_{d}
\right)
\operatorname{vec}\!\left(\mathbf{W}^{(d)}_{:m_d:}\right)
\right]
\\
&=
\sum_{r_d=1}^{R_d}
\sum_{r_{d+1}=1}^{R_{d+1}}
\mathbb{E}_q[\lambda_{r_d}]
\mathbb{E}_q[\lambda_{r_{d+1}}]
\mathbb{E}_q\!\left[
\left(
w^{(d)}_{r_dm_dr_{d+1}}
\right)^2
\right]
\\
&=
\sum_{r_d=1}^{R_d}
\sum_{r_{d+1}=1}^{R_{d+1}}
\mathbb{E}_q[\lambda_{r_d}]
\mathbb{E}_q[\lambda_{r_{d+1}}]
\left\{
\left(
\mathbb{E}_q[
w^{(d)}_{r_dm_dr_{d+1}}]
\right)^2
+
\operatorname{Var}\!\left(
w^{(d)}_{r_dm_dr_{d+1}}
\right)
\right\}
\\
&=
\operatorname{vec}\!\left(
\tilde{\mathbf{W}}^{(d)}_{:m_d:}
\right)^{\top}
\left(
\mathbb{E}_q[\bm{\Lambda}_{d+1}]
\otimes
\mathbb{E}_q[\bm{\Lambda}_{d}]
\right)
\operatorname{vec}\!\left(
\tilde{\mathbf{W}}^{(d)}_{:m_d:}
\right)
+
\operatorname{Var}\!\left(
\operatorname{vec}\!\left(
\mathbf{W}^{(d)}_{:m_d:}
\right)
\right)^{\top}
\left(
\mathbb{E}_q[\bm{\lambda}_{d+1}]
\otimes
\mathbb{E}_q[\bm{\lambda}_{d}]
\right).
\end{aligned}
\end{equation*}

\noindent Here, $\operatorname{Var}$ applied to a vector denotes the vector
of its marginal variances.

\newpage

\textbf{7. The variational posterior distribution of rank precisions $\bm{\lambda}_{d}$}
We derive the variational posterior of $\bm{\lambda}_{d}$ by collecting only the terms
in the log-joint that depend on $\bm{\lambda}_{d}$. Since $\bm{\lambda}_{d}$ is shared
between the $(d-1)$th and $d$th TT cores, both cores contribute to its posterior.

\begin{equation*}
\begin{aligned}
\ln q(\bm{\lambda}_{d})
&=
\mathbb{E}_{q(\bm{\Theta}\backslash\bm{\lambda}_{d})}
\left[
\ln p(\mathbf{y},\bm{\Theta})
\right]
+\text{const}
\\
&=
\mathbb{E}\Bigg[
-
\frac{1}{2}
\sum_{d=1}^{D}
\sum_{r_d=1}^{R_d}
\sum_{m_d=1}^{M_d}
\sum_{r_{d+1}=1}^{R_{d+1}}
w^{(d)}_{r_d m_d r_{d+1}}
\lambda_{r_d}\delta_{m_d}\lambda_{r_{d+1}}
w^{(d)}_{r_d m_d r_{d+1}}
\\
&\hspace{1cm}
+
\sum_{d=2}^{D}
\sum_{r_d=1}^{R_d}
\left(
\frac{R_{d-1}M_{d-1}+M_dR_{d+1}}{2}
+
c_0^{r_d}
-
1
\right)
\ln\lambda_{r_d}
-
\sum_{r_d=1}^{R_d}
e_0^{r_d}\lambda_{r_d}
\Bigg]
+\text{const}
\\
&=
\mathbb{E}\Bigg[
\sum_{r_d=1}^{R_d}
\Bigg\{
\left(
c_0^{r_d}+
\frac{R_{d-1}M_{d-1}+M_dR_{d+1}}{2}
-1
\right)
\ln\lambda_{r_d}
\\
&\hspace{1.6cm}
-
\Bigg(
e_0^{r_d}
+
\frac{1}{2}
\sum_{r_{d-1}=1}^{R_{d-1}}
\sum_{m_{d-1}=1}^{M_{d-1}}
w^{(d-1)}_{r_{d-1}m_{d-1}r_d}
\lambda_{r_{d-1}}
\delta_{m_{d-1}}
w^{(d-1)}_{r_{d-1}m_{d-1}r_d}
\\
&\hspace{3.0cm}
+
\frac{1}{2}
\sum_{m_d=1}^{M_d}
\sum_{r_{d+1}=1}^{R_{d+1}}
w^{(d)}_{r_dm_dr_{d+1}}
\delta_{m_d}
\lambda_{r_{d+1}}
w^{(d)}_{r_dm_dr_{d+1}}
\Bigg)
\lambda_{r_d}
\Bigg\}
\Bigg]
+\text{const}
\\
&=
\sum_{r_d=1}^{R_d}
\Bigg\{
\left(
c_0^{r_d}+
\frac{R_{d-1}M_{d-1}+M_dR_{d+1}}{2}
-1
\right)
\ln\lambda_{r_d}
\\
&\hspace{1cm}
-
\Bigg(
e_0^{r_d}
+
\frac{1}{2}
\mathbb{E}_q\!\left[
\operatorname{vec}\!\left(\mathbf{W}^{(d-1)}_{::r_d}\right)^{\top}
\left(
\bm{\Delta}_{d-1}\otimes\bm{\Lambda}_{d-1}
\right)
\operatorname{vec}\!\left(\mathbf{W}^{(d-1)}_{::r_d}\right)
\right]
\\
&\qquad\qquad
+
\frac{1}{2}
\mathbb{E}_q\!\left[
\operatorname{vec}\!\left(\mathbf{W}^{(d)}_{r_d::}\right)^{\top}
\left(
\bm{\Lambda}_{d+1}\otimes\bm{\Delta}_{d}
\right)
\operatorname{vec}\!\left(\mathbf{W}^{(d)}_{r_d::}\right)
\right]
\Bigg)
\lambda_{r_d}
\Bigg\}
+\text{const}.
\end{aligned}
\end{equation*}

\noindent Identifying this as a log-Gamma distribution in $\lambda_{r_d}$, the posterior is
$\operatorname{Ga}(\lambda_{r_d}\mid c_N^{r_d},e_N^{r_d})$, with updated parameters

\begin{equation*}
\begin{aligned}
c_N^{r_d}
&=
c_0^{r_d}+
\frac{1}{2}
\left(
R_{d-1}M_{d-1}
+
M_dR_{d+1}
\right),
\\
e_N^{r_d}
&=
e_0^{r_d}
+
\frac{1}{2}
\mathbb{E}_q\!\left[
\operatorname{vec}\!\left(\mathbf{W}^{(d-1)}_{::r_d}\right)^{\top}
\left(
\bm{\Delta}_{d-1}\otimes\bm{\Lambda}_{d-1}
\right)
\operatorname{vec}\!\left(\mathbf{W}^{(d-1)}_{::r_d}\right)
\right]
\\
&\quad
+
\frac{1}{2}
\mathbb{E}_q\!\left[
\operatorname{vec}\!\left(\mathbf{W}^{(d)}_{r_d::}\right)^{\top}
\left(
\bm{\Lambda}_{d+1}\otimes\bm{\Delta}_{d}
\right)
\operatorname{vec}\!\left(\mathbf{W}^{(d)}_{r_d::}\right)
\right].
\end{aligned}
\end{equation*}

\newpage

\noindent The first posterior expectation can be computed as

\begin{equation*}
\begin{aligned}
&
\mathbb{E}_q\!\left[
\operatorname{vec}\!\left(\mathbf{W}^{(d-1)}_{::r_d}\right)^{\top}
\left(
\bm{\Delta}_{d-1}\otimes\bm{\Lambda}_{d-1}
\right)
\operatorname{vec}\!\left(\mathbf{W}^{(d-1)}_{::r_d}\right)
\right]
\\
&=
\sum_{r_{d-1}=1}^{R_{d-1}}
\sum_{m_{d-1}=1}^{M_{d-1}}
\mathbb{E}_q[\delta_{m_{d-1}}]
\mathbb{E}_q[\lambda_{r_{d-1}}]
\mathbb{E}_q\!\left[
\left(
w^{(d-1)}_{r_{d-1}m_{d-1}r_d}
\right)^2
\right]
\\
&=
\sum_{r_{d-1}=1}^{R_{d-1}}
\sum_{m_{d-1}=1}^{M_{d-1}}
\mathbb{E}_q[\delta_{m_{d-1}}]
\mathbb{E}_q[\lambda_{r_{d-1}}]
\left\{
\left(
\mathbb{E}_q\!\left[
w^{(d-1)}_{r_{d-1}m_{d-1}r_d}
\right]
\right)^2
+
\operatorname{Var}\!\left(
w^{(d-1)}_{r_{d-1}m_{d-1}r_d}
\right)
\right\}
\\
&=
\operatorname{vec}\!\left(
\tilde{\mathbf{W}}^{(d-1)}_{::r_d}
\right)^{\top}
\left(
\mathbb{E}_q[\bm{\Delta}_{d-1}]
\otimes
\mathbb{E}_q[\bm{\Lambda}_{d-1}]
\right)
\operatorname{vec}\!\left(
\tilde{\mathbf{W}}^{(d-1)}_{::r_d}
\right)
+
\operatorname{Var}\!\left(
\operatorname{vec}\!\left(
\mathbf{W}^{(d-1)}_{::r_d}
\right)
\right)^{\top}
\left(
\mathbb{E}_q[\bm{\delta}_{d-1}]
\otimes
\mathbb{E}_q[\bm{\lambda}_{d-1}]
\right).
\end{aligned}
\end{equation*}

 \noindent Similarly, the second posterior expectation can be computed as

\begin{equation*}
\begin{aligned}
&
\mathbb{E}_q\!\left[
\operatorname{vec}\!\left(\mathbf{W}^{(d)}_{r_d::}\right)^{\top}
\left(
\bm{\Lambda}_{d+1}\otimes\bm{\Delta}_{d}
\right)
\operatorname{vec}\!\left(\mathbf{W}^{(d)}_{r_d::}\right)
\right]
\\
&=
\sum_{m_d=1}^{M_d}
\sum_{r_{d+1}=1}^{R_{d+1}}
\mathbb{E}_q[\delta_{m_d}]
\mathbb{E}_q[\lambda_{r_{d+1}}]
\mathbb{E}_q\!\left[
\left(
w^{(d)}_{r_dm_dr_{d+1}}
\right)^2
\right]
\\
&=
\sum_{m_d=1}^{M_d}
\sum_{r_{d+1}=1}^{R_{d+1}}
\mathbb{E}_q[\delta_{m_d}]
\mathbb{E}_q[\lambda_{r_{d+1}}]
\left\{
\left(
\mathbb{E}_q\!\left[
w^{(d)}_{r_dm_dr_{d+1}}
\right]
\right)^2
+
\operatorname{Var}\!\left(
w^{(d)}_{r_dm_dr_{d+1}}
\right)
\right\}
\\
&=
\operatorname{vec}\!\left(
\tilde{\mathbf{W}}^{(d)}_{r_d::}
\right)^{\top}
\left(
\mathbb{E}_q[\bm{\Lambda}_{d+1}]
\otimes
\mathbb{E}_q[\bm{\Delta}_{d}]
\right)
\operatorname{vec}\!\left(
\tilde{\mathbf{W}}^{(d)}_{r_d::}
\right)
+
\operatorname{Var}\!\left(
\operatorname{vec}\!\left(
\mathbf{W}^{(d)}_{r_d::}
\right)
\right)^{\top}
\left(
\mathbb{E}_q[\bm{\lambda}_{d+1}]
\otimes
\mathbb{E}_q[\bm{\delta}_{d}]
\right).
\end{aligned}
\end{equation*}

\vspace{1cm }
\textbf{8. The variational posterior distribution of noise precision $\tau$}
We derive the variational posterior of $\tau$ by collecting only the terms in the 
log-joint that depend on $\tau$.
\begin{equation*}
    \begin{aligned}
        \operatorname{ln} \hspace{1.5mm} q(\tau) &= \mathbb{E}_{q(\bm{\Theta} \backslash \tau)} [\operatorname{ln} \hspace{1.5mm}  p(\mathbf{y},\bm{\Theta})] + \text{const} \\  
        &= \mathbb{E} \left[ - \frac{\tau}{2} \| \mathbf{y} -\mathbf{\Phi}^ {\top} \mathbf{w} \|_F^{2}  + \left( \frac{N}{2} + a_0 -1\right)  \operatorname{ln} \tau  - b_0 \tau \right] + \text{const} \\
        &=  \left( \frac{N}{2} + a_0 -1\right)  \operatorname{ln} \tau - \left( b_0 + \frac{1}{2} \mathbb{E}_q\left[ \| \mathbf{y} -\mathbf{\Phi}^ {\top} \mathbf{w}\|_F^{2}\right] \right) \tau + \text{const}
    \end{aligned}
\end{equation*}

\vspace{-0.3cm}
\noindent Thus we observe that the posterior is a Gamma distribution with updated parameters: 
\begin{equation*}
    \begin{aligned}
        a_N &= a_0 + \frac{N}{2}, \qquad
        b_N &= b_0 + \frac{1}{2} \mathbb{E}_q \left[\| \mathbf{y} - \mathbf{\Phi}^ {\top} \mathbf{w}\|_F^{2}\right].
 \end{aligned}
 \end{equation*}  

\newpage

\begingroup
\allowdisplaybreaks[4]
\textbf{9. Lower bound model evidence}
We expand the ELBO into expected log-likelihood, expected log-prior, and posterior
entropy terms for each group of parameters, evaluate each using the posterior
parameters, and collect them into the compact form used as the convergence criterion.

\begin{align*}
\mathcal{L}(q)
&=\mathbb{E}_{q(\Theta)}[\ln p(\mathbf{y},\Theta)]+H(q(\Theta))\\
&=\mathbb{E}_q\!\left[
\ln p\left(\mathbf{y}\mid\{\bm{\mathcal{W}}^{(d)}\}_{d=1}^{D},\tau\right)
\right]
+\mathbb{E}_q\!\left[
\sum_{d=1}^{D}\ln p\left(\bm{\mathcal{W}}^{(d)}\mid
\bm{\lambda}_d,\bm{\delta}_d,\bm{\lambda}_{d+1}\right)\right] +\mathbb{E}_q\!\left[\sum_{d=1}^{D}\ln p(\bm{\delta}_d)\right] \\
\quad
 &+\mathbb{E}_q\!\left[\sum_{d=2}^{D}\ln p(\bm{\lambda}_d)\right]
+\mathbb{E}_q[\ln p(\tau)]
-\mathbb{E}_q\!\left[\sum_{d=1}^{D}\ln q(\bm{\mathcal{W}}^{(d)})\right]
-\mathbb{E}_q\!\left[\sum_{d=1}^{D}\ln q(\bm{\delta}_d)\right]\\
&\quad-\mathbb{E}_q\!\left[\sum_{d=2}^{D}\ln q(\bm{\lambda}_d)\right]
-\mathbb{E}_q[\ln q(\tau)].
\end{align*}

\noindent All expectations are with respect to the posterior $q$. The first term is the
expected log-likelihood, followed by expected log-priors on
$\{\bm{\mathcal{W}}^{(d)}\}_{d=1}^{D}$,
$\{\bm{\delta}_d\}_{d=1}^{D}$,
$\{\bm{\lambda}_d\}_{d=2}^{D}$, and $\tau$;
the final four terms are the corresponding posterior entropies.
In the following expressions, $\psi(\cdot)$ denotes the digamma function and
$\Gamma(\cdot)$ denotes the Gamma function. The boundary rank precisions are
fixed at $\bm{\Lambda}_1=\bm{\Lambda}_{D+1}=[1]$, with
$R_1=R_{D+1}=1$, and have no hyperprior or posterior entropy terms.
Each term can be computed by

\begin{align*}
\mathbb{E}_q\!\left[
\ln p\left(\mathbf{y}\mid\{\bm{\mathcal{W}}^{(d)}\}_{d=1}^{D},\tau\right)
\right] &=-\frac{N}{2}\ln(2\pi)
+\frac{N}{2}\mathbb{E}_q[\ln\tau]
-\frac{1}{2}\mathbb{E}_q[\tau]\,
\mathbb{E}_q\!\left[\|\mathbf{y}-\mathbf{\Phi}^{\top}\mathbf{w}\|_F^2\right]\\
&=-\frac{N}{2}\ln(2\pi)
+\frac{N}{2}\bigl(\psi(a_N)-\ln b_N\bigr)
-\frac{a_N}{2b_N}\mathbb{E}_q\!\left[\|\mathbf{y}-\mathbf{\Phi}^{\top}\mathbf{w}\|_F^2\right].
\end{align*}

\noindent Here $N$ denotes the number of observations. The expected squared
residual is computed using the forward accumulators in the TT model as
\begin{equation*}
\mathbb{E}_q\!\left[\|\mathbf{y}-\mathbf{\Phi}^{\top}\mathbf{w}\|_F^2\right]
=\|\mathbf{y}\|_F^2
-2\mathbb{E}_q[\mathbf{G}_{<D+1}]\mathbf{y}
+\mathbf{H}_{<D+1}\mathbf{1}_N.
\end{equation*}

\vspace{-1cm}
\begin{align*}
\mathbb{E}_q\!\left[\sum_{d=2}^{D}\ln p(\bm{\lambda}_d)\right]
&=\sum_{d=2}^{D}\sum_{r_d=1}^{R_d}\Bigl\{
-\ln\Gamma(c_0^{r_d})+c_0^{r_d}\ln e_0^{r_d}
+(c_0^{r_d}-1)\mathbb{E}_q[\ln\lambda_{r_d}]
-e_0^{r_d}\mathbb{E}_q[\lambda_{r_d}]
\Bigr\}\\
&=\sum_{d=2}^{D}\sum_{r_d=1}^{R_d}\Bigl\{
-\ln\Gamma(c_0^{r_d})+c_0^{r_d}\ln e_0^{r_d}
+(c_0^{r_d}-1)\bigl(\psi(c_N^{r_d})-\ln e_N^{r_d}\bigr)
-e_0^{r_d}\frac{c_N^{r_d}}{e_N^{r_d}}
\Bigr\},\\[2mm]
\mathbb{E}_q\!\left[\sum_{d=1}^{D}\ln p(\bm{\delta}_d)\right]
&=\sum_{d=1}^{D}\sum_{m_d=1}^{M_d}\Bigl\{
-\ln\Gamma(g_0^{m_d})+g_0^{m_d}\ln h_0^{m_d}
+(g_0^{m_d}-1)\mathbb{E}_q[\ln\delta_{m_d}]
-h_0^{m_d}\mathbb{E}_q[\delta_{m_d}]
\Bigr\}\\
&=\sum_{d=1}^{D}\sum_{m_d=1}^{M_d}\Bigl\{
-\ln\Gamma(g_0^{m_d})+g_0^{m_d}\ln h_0^{m_d}
+(g_0^{m_d}-1)\bigl(\psi(g_N^{m_d})-\ln h_N^{m_d}\bigr)
-h_0^{m_d}\frac{g_N^{m_d}}{h_N^{m_d}}
\Bigr\},\\[2mm]
\mathbb{E}_q[\ln p(\tau)]&=-\ln\Gamma(a_0)+a_0\ln b_0
+(a_0-1)\mathbb{E}_q[\ln\tau]-b_0\mathbb{E}_q[\tau]\\
&=-\ln\Gamma(a_0)+a_0\ln b_0
+(a_0-1)\bigl(\psi(a_N)-\ln b_N\bigr)-b_0\frac{a_N}{b_N}.
\end{align*}

\begin{align*}
&\mathbb{E}_q\!\left[\sum_{d=1}^{D}
\ln p\left(\bm{\mathcal{W}}^{(d)}\mid
\bm{\lambda}_d,\bm{\delta}_d,\bm{\lambda}_{d+1}\right)\right]\\
\\
&=\mathbb{E}_q\!\Biggl[
\sum_{d=1}^{D}\sum_{r_d=1}^{R_d}\sum_{m_d=1}^{M_d}
\sum_{r_{d+1}=1}^{R_{d+1}}\Biggl\{
-\frac{1}{2}\ln(2\pi)
+\frac{1}{2}\ln\bigl(\lambda_{r_d}\delta_{m_d}\lambda_{r_{d+1}}\bigr)
-\frac{1}{2}\bigl(w^{(d)}_{r_dm_dr_{d+1}}\bigr)^2
\lambda_{r_d}\delta_{m_d}\lambda_{r_{d+1}}
\Biggr\}\Biggr]\\
\\
&=\mathbb{E}_q\!\Biggl[
\sum_{d=1}^{D}\sum_{r_d=1}^{R_d}\sum_{m_d=1}^{M_d}
\sum_{r_{d+1}=1}^{R_{d+1}}\Biggl\{
-\frac{1}{2}\ln(2\pi)
+\frac{1}{2}\bigl(\ln\lambda_{r_d}+\ln\delta_{m_d}
+\ln\lambda_{r_{d+1}}\bigr)
-\frac{1}{2}\bigl(w^{(d)}_{r_dm_dr_{d+1}}\bigr)^2
\lambda_{r_d}\delta_{m_d}\lambda_{r_{d+1}}
\Biggr\}\Biggr]\\
\\
&=\sum_{d=1}^{D}\Biggl\{
-\frac{R_dM_dR_{d+1}}{2}\ln(2\pi)
+\frac{M_dR_{d+1}}{2}\sum_{r_d=1}^{R_d}
\mathbb{E}_q[\ln\lambda_{r_d}]+\frac{R_dR_{d+1}}{2}\sum_{m_d=1}^{M_d}
\mathbb{E}_q[\ln\delta_{m_d}]
+\frac{R_dM_d}{2}\sum_{r_{d+1}=1}^{R_{d+1}}
\mathbb{E}_q[\ln\lambda_{r_{d+1}}]\\
&\qquad-\frac{1}{2}\mathbb{E}_q\!\left[
\operatorname{vec}(\bm{\mathcal{W}}^{(d)})^{\top}
\bigl(\bm{\Lambda}_{d+1}\otimes\bm{\Delta}_d\otimes\bm{\Lambda}_d\bigr)
\operatorname{vec}(\bm{\mathcal{W}}^{(d)})\right]\Biggr\}\\
\\
&=-\frac{\sum_{d=1}^{D}R_dM_dR_{d+1}}{2}\ln(2\pi)
+\sum_{d=2}^{D}\sum_{r_d=1}^{R_d}\frac{R_{d-1}M_{d-1}+M_dR_{d+1}}{2}\mathbb{E}_q[\ln\lambda_{r_d}]+\sum_{d=1}^{D}\sum_{m_d=1}^{M_d}\frac{R_dR_{d+1}}{2}\mathbb{E}_q[\ln\delta_{m_d}]\\
&\quad-\frac{1}{2}\sum_{d=1}^{D}\Biggl\{
\operatorname{Tr}\!\left(\bigl(\mathbb{E}_q[\bm{\Lambda}_{d+1}]\otimes\mathbb{E}_q[\bm{\Delta}_d]\otimes\mathbb{E}_q[\bm{\Lambda}_d]\bigr)\mathbf{\Sigma}^{(d)}\right)
+\operatorname{vec}(\tilde{\bm{\mathcal{W}}}^{(d)})^{\top}\bigl(\mathbb{E}_q[\bm{\Lambda}_{d+1}]\otimes\mathbb{E}_q[\bm{\Delta}_d]\otimes\mathbb{E}_q[\bm{\Lambda}_d]\bigr)\operatorname{vec}(\tilde{\bm{\mathcal{W}}}^{(d)})\Biggr\}\\
\\
&=-\frac{\sum_{d=1}^{D}R_dM_dR_{d+1}}{2}\ln(2\pi)
+\sum_{d=2}^{D}\sum_{r_d=1}^{R_d}\frac{R_{d-1}M_{d-1}+M_dR_{d+1}}{2}\bigl(\psi(c_N^{r_d})-\ln e_N^{r_d}\bigr)\\
&\quad+\sum_{d=1}^{D}\sum_{m_d=1}^{M_d}\frac{R_dR_{d+1}}{2}\bigl(\psi(g_N^{m_d})-\ln h_N^{m_d}\bigr)-\frac{1}{2}\sum_{d=1}^{D}\operatorname{Tr}\!\Bigl\{\bigl(\mathbb{E}_q[\bm{\Lambda}_{d+1}]\otimes\mathbb{E}_q[\bm{\Delta}_d]\otimes\mathbb{E}_q[\bm{\Lambda}_d]\bigr)\bigl(\operatorname{vec}(\tilde{\bm{\mathcal{W}}}^{(d)})\operatorname{vec}(\tilde{\bm{\mathcal{W}}}^{(d)})^{\top}+\mathbf{\Sigma}^{(d)}\bigr)\Bigr\}.
\end{align*}

The posterior entropy terms are given by
\begin{align*}
-\mathbb{E}_q\!\left[\sum_{d=1}^{D}\ln q(\bm{\mathcal{W}}^{(d)})\right]
&=\frac{1}{2}\sum_{d=1}^{D}\ln|\mathbf{\Sigma}^{(d)}|
+\frac{\sum_{d=1}^{D}R_dM_dR_{d+1}}{2}\bigl[1+\ln(2\pi)\bigr],\\
-\mathbb{E}_q\!\left[\sum_{d=2}^{D}\ln q(\bm{\lambda}_d)\right]
&=\sum_{d=2}^{D}\sum_{r_d=1}^{R_d}\Bigl\{
\ln\Gamma(c_N^{r_d})-(c_N^{r_d}-1)\psi(c_N^{r_d})
-\ln e_N^{r_d}+c_N^{r_d}\Bigr\},\\[2mm]
-\mathbb{E}_q\!\left[\sum_{d=1}^{D}\ln q(\bm{\delta}_d)\right]
&=\sum_{d=1}^{D}\sum_{m_d=1}^{M_d}\Bigl\{
\ln\Gamma(g_N^{m_d})-(g_N^{m_d}-1)\psi(g_N^{m_d})
-\ln h_N^{m_d}+g_N^{m_d}\Bigr\},\\[2mm]
-\mathbb{E}_q[\ln q(\tau)]
&=\ln\Gamma(a_N)-(a_N-1)\psi(a_N)-\ln b_N+a_N.
\end{align*}

\noindent Combining all these terms together and collecting the terms independent of the
variational posterior into $\text{const}$ gives
\begin{align*}
\mathcal{L}(q)
&=\frac{N}{2}\bigl(\psi(a_N)-\ln b_N\bigr)-\frac{a_N}{2b_N}\mathbb{E}_q\!\left[\|\mathbf{y}-\mathbf{\Phi}^{\top}\mathbf{w}\|_F^2\right]\\
&\quad+\sum_{d=2}^{D}\sum_{r_d=1}^{R_d}\frac{R_{d-1}M_{d-1}+M_dR_{d+1}}{2}\bigl(\psi(c_N^{r_d})-\ln e_N^{r_d}\bigr)
+\sum_{d=1}^{D}\sum_{m_d=1}^{M_d}\frac{R_dR_{d+1}}{2}\bigl(\psi(g_N^{m_d})-\ln h_N^{m_d}\bigr)\\
&\quad-\frac{1}{2}\sum_{d=1}^{D}\operatorname{Tr}\!\Bigl\{\bigl(\mathbb{E}_q[\bm{\Lambda}_{d+1}]\otimes\mathbb{E}_q[\bm{\Delta}_d]\otimes\mathbb{E}_q[\bm{\Lambda}_d]\bigr)\bigl(\operatorname{vec}(\tilde{\bm{\mathcal{W}}}^{(d)})\operatorname{vec}(\tilde{\bm{\mathcal{W}}}^{(d)})^{\top}+\mathbf{\Sigma}^{(d)}\bigr)\Bigr\}\\
&\quad+\sum_{d=2}^{D}\sum_{r_d=1}^{R_d}\Bigl\{(c_0^{r_d}-1)\bigl(\psi(c_N^{r_d})-\ln e_N^{r_d}\bigr)
-e_0^{r_d}\frac{c_N^{r_d}}{e_N^{r_d}}\Bigr\}+\sum_{d=1}^{D}\sum_{m_d=1}^{M_d}\Bigl\{(g_0^{m_d}-1)\bigl(\psi(g_N^{m_d})-\ln h_N^{m_d}\bigr)
-h_0^{m_d}\frac{g_N^{m_d}}{h_N^{m_d}}\Bigr\}\\
&\quad+(a_0-1)\bigl(\psi(a_N)-\ln b_N\bigr)-b_0\frac{a_N}{b_N}
+\frac{1}{2}\sum_{d=1}^{D}\ln|\mathbf{\Sigma}^{(d)}|+\sum_{d=2}^{D}\sum_{r_d=1}^{R_d}\Bigl\{\ln\Gamma(c_N^{r_d})
-(c_N^{r_d}-1)\psi(c_N^{r_d})-\ln e_N^{r_d}+c_N^{r_d}\Bigr\}\\
&\quad+\sum_{d=1}^{D}\sum_{m_d=1}^{M_d}\Bigl\{\ln\Gamma(g_N^{m_d})
-(g_N^{m_d}-1)\psi(g_N^{m_d})-\ln h_N^{m_d}+g_N^{m_d}\Bigr\}+\ln\Gamma(a_N)-(a_N-1)\psi(a_N)-\ln b_N+a_N
\\&\quad+\text{const}\\
&=\left(\frac{N}{2}+a_0-a_N\right)\psi(a_N)
-\left(\frac{N}{2}+a_0\right)\ln b_N
-\frac{a_N}{2b_N}\mathbb{E}_q\!\left[\|\mathbf{y}-\mathbf{\Phi}^{\top}\mathbf{w}\|_F^2\right]\\
&\quad-\frac{1}{2}\sum_{d=1}^{D}\operatorname{Tr}\!\Bigl\{\bigl(\mathbb{E}_q[\bm{\Lambda}_{d+1}]\otimes\mathbb{E}_q[\bm{\Delta}_d]\otimes\mathbb{E}_q[\bm{\Lambda}_d]\bigr)\bigl(\operatorname{vec}(\tilde{\bm{\mathcal{W}}}^{(d)})\operatorname{vec}(\tilde{\bm{\mathcal{W}}}^{(d)})^{\top}+\mathbf{\Sigma}^{(d)}\bigr)\Bigr\}+\frac{1}{2}\sum_{d=1}^{D}\ln|\mathbf{\Sigma}^{(d)}|\\
&\quad+\sum_{d=2}^{D}\sum_{r_d=1}^{R_d}\Biggl\{
\left(\frac{R_{d-1}M_{d-1}+M_dR_{d+1}}{2}+c_0^{r_d}-c_N^{r_d}\right)\psi(c_N^{r_d})-\left(\frac{R_{d-1}M_{d-1}+M_dR_{d+1}}{2}+c_0^{r_d}\right)\ln e_N^{r_d}
\Biggr\}\\
&\quad+\sum_{d=1}^{D}\sum_{m_d=1}^{M_d}\Biggl\{
\left(\frac{R_dR_{d+1}}{2}+g_0^{m_d}-g_N^{m_d}\right)\psi(g_N^{m_d})
-\left(\frac{R_dR_{d+1}}{2}+g_0^{m_d}\right)\ln h_N^{m_d}\Biggr\}\\
&\quad+\sum_{d=2}^{D}\sum_{r_d=1}^{R_d}\Bigl\{\ln\Gamma(c_N^{r_d})+c_N^{r_d}
-e_0^{r_d}\frac{c_N^{r_d}}{e_N^{r_d}}\Bigr\}+\sum_{d=1}^{D}\sum_{m_d=1}^{M_d}\Bigl\{\ln\Gamma(g_N^{m_d})+g_N^{m_d}
-h_0^{m_d}\frac{g_N^{m_d}}{h_N^{m_d}}\Bigr\}+\ln\Gamma(a_N)+a_N-b_0\frac{a_N}{b_N}\\&\quad+\text{const}.\\
&=-\frac{a_N}{2b_N}\mathbb{E}_q\!\left[\|\mathbf{y}-\mathbf{\Phi}^{\top}\mathbf{w}\|_F^2\right]-\frac{1}{2}\sum_{d=1}^{D}\operatorname{Tr}\!\Bigl\{\bigl(\mathbb{E}_q[\bm{\Lambda}_{d+1}]\otimes\mathbb{E}_q[\bm{\Delta}_d]\otimes\mathbb{E}_q[\bm{\Lambda}_d]\bigr)\bigl(\operatorname{vec}(\tilde{\bm{\mathcal{W}}}^{(d)})\operatorname{vec}(\tilde{\bm{\mathcal{W}}}^{(d)})^{\top}+\mathbf{\Sigma}^{(d)}\bigr)\Bigr\}\\
&\quad+\sum_{d=2}^{D}\sum_{r_d=1}^{R_d}\Biggl\{
\ln\Gamma(c_N^{r_d})
+c_N^{r_d}\left(1-\ln e_N^{r_d}-\frac{e_0^{r_d}}{e_N^{r_d}}\right)
\Biggr\}+\sum_{d=1}^{D}\sum_{m_d=1}^{M_d}\Biggl\{
\ln\Gamma(g_N^{m_d})
+g_N^{m_d}\left(1-\ln h_N^{m_d}-\frac{h_0^{m_d}}{h_N^{m_d}}\right)
\Biggr\}\\
&\quad+\frac{1}{2}\sum_{d=1}^{D}\ln|\mathbf{\Sigma}^{(d)}|
+\ln\Gamma(a_N)
+a_N\left(1-\ln b_N-\frac{b_0}{b_N}\right) + \text{const}.
\end{align*}

\noindent Finally, the lower bound can be computed using the posterior parameters of all
unknowns in $\Theta$, where the posterior parameters are updated at each iteration.
\endgroup

\newpage
\textbf{10. Predictive Distribution}
We marginalize sequentially over each TT core $\bm{\mathcal{W}}^{(d)}$ via a
Gaussian-Gaussian approximation, adding one variance term per mode, and then over
$\tau$ with its Gamma posterior, yielding an approximate Student's t-distribution.
Note that the true posterior $p(\bm{\Theta}\mid\mathbf{y})$ is approximated by
$q(\bm{\Theta})$, whose factors were derived in Sections 3, 6, 7, and 8.

\begin{align*}
p(y_i\mid\mathbf{y})
&=\int p(y_i\mid\bm{\Theta})\,
p(\bm{\Theta}\mid\mathbf{y})\,d\bm{\Theta}\\[2mm]
&\simeq\int\!\int
p\!\left(y_i\mid\{\bm{\mathcal{W}}^{(d)}\},\tau\right)
q\!\left(\{\bm{\mathcal{W}}^{(d)}\}\right)q(\tau)
\,d\{\bm{\mathcal{W}}^{(d)}\}\,d\tau\\[2mm]
&=\int\!\int\mathcal{N}\!\left(y_i\mid \mathbf{g}_{<D+1}(\mathbf{x}_i),\tau^{-1}\right)\prod_{d=1}^{D}\mathcal{N}\!\left(
\operatorname{vec}(\bm{\mathcal{W}}^{(d)})\mid
\operatorname{vec}(\tilde{\bm{\mathcal{W}}}^{(d)}),\mathbf{\Sigma}^{(d)}\right)
q(\tau)\,d\{\operatorname{vec}(\bm{\mathcal{W}}^{(d)})\}\,d\tau\\[2mm]
&=\int\!\int\mathcal{N}\!\left(y_i\mid
\operatorname{vec}(\bm{\mathcal{W}}^{(1)})^{\top}\mathbf{g}_{\backslash 1}(\mathbf{x}_i),\tau^{-1}\right)\mathcal{N}\!\left(
\operatorname{vec}(\bm{\mathcal{W}}^{(1)})\mid
\operatorname{vec}(\tilde{\bm{\mathcal{W}}}^{(1)}),\mathbf{\Sigma}^{(1)}\right)
\,d\operatorname{vec}(\bm{\mathcal{W}}^{(1)})\,\cdots\\*
&\qquad\cdots\prod_{d\ne1}\mathcal{N}\!\left(
\operatorname{vec}(\bm{\mathcal{W}}^{(d)})\mid
\operatorname{vec}(\tilde{\bm{\mathcal{W}}}^{(d)}),\mathbf{\Sigma}^{(d)}\right) q(\tau)
\,d\{\operatorname{vec}(\bm{\mathcal{W}}^{(d)})\}_{d\ne1}\,d\tau\\[2mm]
&=\int\!\int\mathcal{N}\!\Bigl(y_i\mid
\operatorname{vec}(\tilde{\bm{\mathcal{W}}}^{(1)})^{\top}\mathbf{g}_{\backslash 1}(\mathbf{x}_i),\tau^{-1}+\mathbf{g}_{\backslash 1}(\mathbf{x}_i)^{\top}\mathbf{\Sigma}^{(1)}\mathbf{g}_{\backslash 1}(\mathbf{x}_i)\Bigr)
\,\cdots\\*
&\qquad\cdots\prod_{d\ne1}\mathcal{N}\!\left(
\operatorname{vec}(\bm{\mathcal{W}}^{(d)})\mid
\operatorname{vec}(\tilde{\bm{\mathcal{W}}}^{(d)}),\mathbf{\Sigma}^{(d)}\right) q(\tau)
\,d\{\operatorname{vec}(\bm{\mathcal{W}}^{(d)})\}_{d\ne1}\,d\tau\\[2mm]
&\hspace{1cm}\vdots\\[2mm]
&\simeq\int\mathcal{N}\!\Bigl(y_i\mid\mathbb{E}_q[\mathbf{g}_{<D+1}(\mathbf{x}_i)],\tau^{-1}+\sum_{d=1}^{D}\mathbb{E}_q[\mathbf{g}_{\backslash d}(\mathbf{x}_i)]^{\top}\mathbf{\Sigma}^{(d)}\mathbb{E}_q[\mathbf{g}_{\backslash d}(\mathbf{x}_i)]\Bigr)q(\tau)\,d\tau\\[2mm]
&=\int\mathcal{N}\!\Bigl(y_i\mid\mathbb{E}_q[\mathbf{g}_{<D+1}(\mathbf{x}_i)],\tau^{-1}+\sum_{d=1}^{D}\mathbb{E}_q[\mathbf{g}_{\backslash d}(\mathbf{x}_i)]^{\top}\mathbf{\Sigma}^{(d)}\mathbb{E}_q[\mathbf{g}_{\backslash d}(\mathbf{x}_i)]\Bigr)
\operatorname{Ga}(\tau\mid a_N,b_N)\,d\tau\\[2mm]
&\simeq\mathcal{T}\!\Biggl(y_i\mid\mathbb{E}_q[\mathbf{g}_{<D+1}(\mathbf{x}_i)],\left\{\frac{b_N}{a_N}+\sum_{d=1}^{D}\mathbb{E}_q[\mathbf{g}_{\backslash d}(\mathbf{x}_i)]^{\top}\mathbf{\Sigma}^{(d)}\mathbb{E}_q[\mathbf{g}_{\backslash d}(\mathbf{x}_i)]\right\}^{-1},\,2a_N\Biggr).
\end{align*}
\noindent where $\mathcal{T}$ is a Student's t-distribution
${y_i}\mid\mathbf{y}\sim\mathcal{T}(\tilde{y_i},\mathcal{S}_i,\nu_y)$
with its parameters given by

\begin{align*}
\tilde{y_i}&=\mathbb{E}_q[\mathbf{g}_{<D+1}(\mathbf{x}_i)],\qquad\nu_y=2a_N,\\[2mm]
\mathcal{S}_i&=\left\{\frac{b_N}{a_N}+\sum_{d=1}^{D}\mathbb{E}_q[\mathbf{g}_{\backslash d}(\mathbf{x}_i)]^{\top}\mathbf{\Sigma}^{(d)}\mathbb{E}_q[\mathbf{g}_{\backslash d}(\mathbf{x}_i)]\right\}^{-1}.
\end{align*}

\noindent Thus, the predictive variance can be obtained by
$\operatorname{Var}(y_i\mid\mathbf{y})=\frac{\nu_y}{\nu_y-2}\mathcal{S}_i^{-1}$,
for $\nu_y>2$.